# USING MULTIPLE SOURCES OF INFORMATION FOR CONSTRAINT-BASED MORPHOLOGICAL DISAMBIGUATION

A THESIS
SUBMITTED TO THE DEPARTMENT OF COMPUTER ENGINEERING
AND INFORMATION SCIENCE
AND THE INSTITUTE OF ENGINEERING AND SCIENCE
OF BILKENT UNIVERSITY
IN PARTIAL FULFILLMENT OF THE REQUIREMENTS
FOR THE DEGREE OF
MASTER OF SCIENCE

By
Gökhan Tür

July, 1996

I certify that I have read this thesis and that in my opinion it is fully adequate, in scope and in quality, as a thesis for the degree of Master of Science.

Asst. Prof. Kemal Oflazer(Advisor)

I certify that I have read this thesis and that in my opinion it is fully adequate, in scope and in quality, as a thesis for the degree of Master of Science.

Assoc. Prof. Halil Altay Güvenir

I certify that I have read this thesis and that in my opinion it is fully adequate, in scope and in quality, as a thesis for the degree of Master of Science.

Asst. Prof. İlyas Çiçekli

Approved for the Institute of Engineering and Science:

Prof. Dr. Mehmet Baray, Director of Institute of Engineering and Science

# ABSTRACT

USING MULTIPLE SOURCES OF INFORMATION
FOR
CONSTRAINT–BASED MORPHOLOGICAL DISAMBIGUATION

Gökhan Tür
M.S. in Computer Engineering and Information Science
Supervisor: Asst. Prof. Kemal Oflazer
July, 1996

This thesis presents a constraint-based morphological disambiguation approach that is applicable to languages with complex morphology–specifically agglutinative languages with productive inflectional and derivational morphological phenomena. For morphologically complex languages like Turkish, automatic morphological disambiguation involves selecting for each token morphological parse(s), with the right set of inflectional and derivational markers. Our system combines corpus independent hand-crafted constraint rules, constraint rules that are learned via unsupervised learning from a training corpus, and additional statistical information obtained from the corpus to be morphologically disambiguated. The hand-crafted rules are linguistically motivated and tuned to improve precision without sacrificing recall. In certain respects, our approach has been motivated by Brill's recent work [6], but with the observation that his transformational approach is not directly applicable to languages like Turkish. Our approach also uses a novel approach to unknown word processing by employing a secondary morphological processor which recovers any relevant inflectional and derivational information from a lexical item whose root is unknown. With this approach, well below 1% of the tokens remains as unknown in the texts we have experimented with. Our results indicate that by combining these hand-crafted, statistical and learned information sources, we can attain a recall of 96 to 97% with a corresponding precision of 93 to 94%, and ambiguity of 1.02 to 1.03 parses per token.

# ÖZET

DEĞİŞİK BİLGİ KAYNAKLARI KULLANARAK
BİÇİMBİRİMSEL BİRİKLEŞTİRME

Gökhan Tür
Bilgisayar ve Enformatik Mühendisliği, Yüksek Lisans
Tez Yöneticisi: Yrd. Doç. Dr. Kemal Oflazer
Temmuz, 1996

Bu tezde, karmaşık biçimbirimli dillerde (özellikle üretken yapım ve çekim eklerine sahip çekimli ve bitişken dillerde) uygulanabilecek, kurallara dayanan bir biçimbirimsel birikleştirme yaklaşımı sunulmaktadır. Türkçe gibi karmaşık biçimbirimsel yapıya sahip dillerde, otomatik biçimbirimsel birikleştirme, kelimelerin, doğru yapım ve çekim eklerini içeren biçimbirimsel çözümlerini seçmeyi amaçlar. Bu çalışmada gerçekleştirilen sistem, metinlerden bağımsız olarak elle oluşturulmuş kuralları, öğrenilmiş kuralları, ve birikleştirilecek metinden elde edilen ek istatistiksel bilgileri kullanarak biçimbirimsel birikleştirme işlevini gerçekleştirmektedir. Elle oluşturulmuş kurallar, anma'dan (recall) fedakarlık etmeden duyarlılığı (precision) artıracak şekilde düzenlenen dilbilimsel kurallardan meydana gelmiştir. Sistemin tasarımının çıkış noktası, Brill'in dönüşümsel yaklaşımının Türkçe gibi dillerde direkt olarak uygulanamayacağı gözlemi olmuştur. Ayrıca bilinmeyen kelimelerin çözümlenmesinde, ikinci bir biçimbirimsel işlemci kullanılarak ve kelimelerdeki olası yapım ve çekim ekleri belirlenerek çözümlemesi yapılmıştır. Bu yaklaşım sayesinde, deneylerde kullanılan metinlerdeki kelimelerin %1'inden çok daha azı çözümsüz kalmıştır. Elle oluşturulmuş ve öğrenilmiş kurallar ile istatistiki bilgilerin birleştirilmesi sayesinde üzerinde deney yaptığımız metinlerde kelime başına 1.02-1.03 çözüm düşerken %96-%97 anma ve buna karşılık %93-%94 duyarlılık sağlanmıştır.

**To my family**

# ACKNOWLEDGEMENTS

I am very grateful to my supervisor, Assistant Professor Kemal Oflazer, who has provided a stimulating research environment and invaluable guidance during this study. His instruction will be the closest and most important reference in my future research.

I would also like to thank Assoc. Prof. Halil Altay Güvenir and Asst. Prof. İlyas Çiçekli for their valuable comments and guidance on this thesis.

I would like to thank Xerox Advanced Document Systems, and Lauri Karttunen of Xerox Parc and of Rank Xerox Research Centre (Grenoble) for providing us with the two-level transducer development software on which the morphological and unknown word recognizer were implemented. This research has been supported in part by a NATO Science for Stability Project Grant TU–LANGUAGE.

I would like to thank everybody who has in some way contributed to this study by lending me moral, technical and intellectual support, including my colleagues Mehmet Surav who taught me even how to use this editor, Kemal Ülkü, A. Kurtuluş Yorulmaz, Yücel Saygın, Murat Bayraktar, and many others who are not mentioned here by name.

I would like to thank to my family. I am very grateful for their moral support, motivation and hope-giving. They are always with me, especially when I need them. I dedicate this thesis to these persons.

Finally, I would like to thank to Ms. Dilek Z. Hakkani. I cannot forget her invaluable technical and moral support which continued during my study. It is Dilek and her friendship that deserve the biggest thanks for the existence of this thesis.

# Contents

# List of Figures

# List of Tables

# Chapter 1

# Introduction

For morphologically complex languages like Turkish, automatic morphological disambiguation involves selecting for each token, morphological parse(s) with the right set of inflectional and derivational markers in the given context. We take a token to be a lexical form occurring in a text, like a word, a punctuation mark, a date, a numeric structure, etc. Such disambiguation is a very crucial component in higher level analysis of natural language text corpora. For example, morphological disambiguation facilitates parsing, essentially by performing a certain amount of ambiguity resolution using relatively cheaper methods (e.g., Güngördü and Oflazer [12], report that parsing with disambiguated text is twice as fast and generates one half ambiguities in general.) Figure 1.1 shows the place of morphological disambiguation in an abstract context.

Typical applications that can benefit from disambiguated text are:

- corpus analysis, e.g. to gather language statistics,
- syntactic parsing, e.g. prior reduction of sentence ambiguity,
- spelling correction, e.g. context sensitive selection of pronunciation,
- speech synthesis, e.g. selection of true spellings.

There has been a large number of studies in morphological disambiguation and part-of-speech tagging – assigning every token its proper part-of-speech based upon the context it appears in – using various techniques. These systems have used either a statistical approach where a large corpora has been used to train a statistical model which then has been used to tag new text, assigning the most likely tag for a given word in a given context (e.g., Church [7], Cutting et. al [9], DeRose [10]), or a constraint-based approach, recently most prominently exemplified by the Constraint Grammar work [15, 28, 29, 30], where a large number of hand-crafted linguistic constraints are used to eliminate impossible tags or morphological parses for a given word in a given context. Using the constraint grammar, it is claimed that an English text can be morphologically disambiguated with 99.77% recall and 95.54% precision[1]. This ratio is better than all of the statistical approaches, which result in 96-97% accuracy. It is also possible to use a hybrid approach, which disambiguate an English text with 98.5% accuracy. Brill [2, 4, 5] has presented a transformation-based learning approach, which induces rules from tagged corpora. Recently he has extended this work so that learning can proceed in an unsupervised manner using an untagged corpus [6]. Levinger et al. [20] have recently reported on an approach that learns morpho-lexical probabilities from untagged corpus and have used the resulting information in morphological disambiguation in Hebrew.

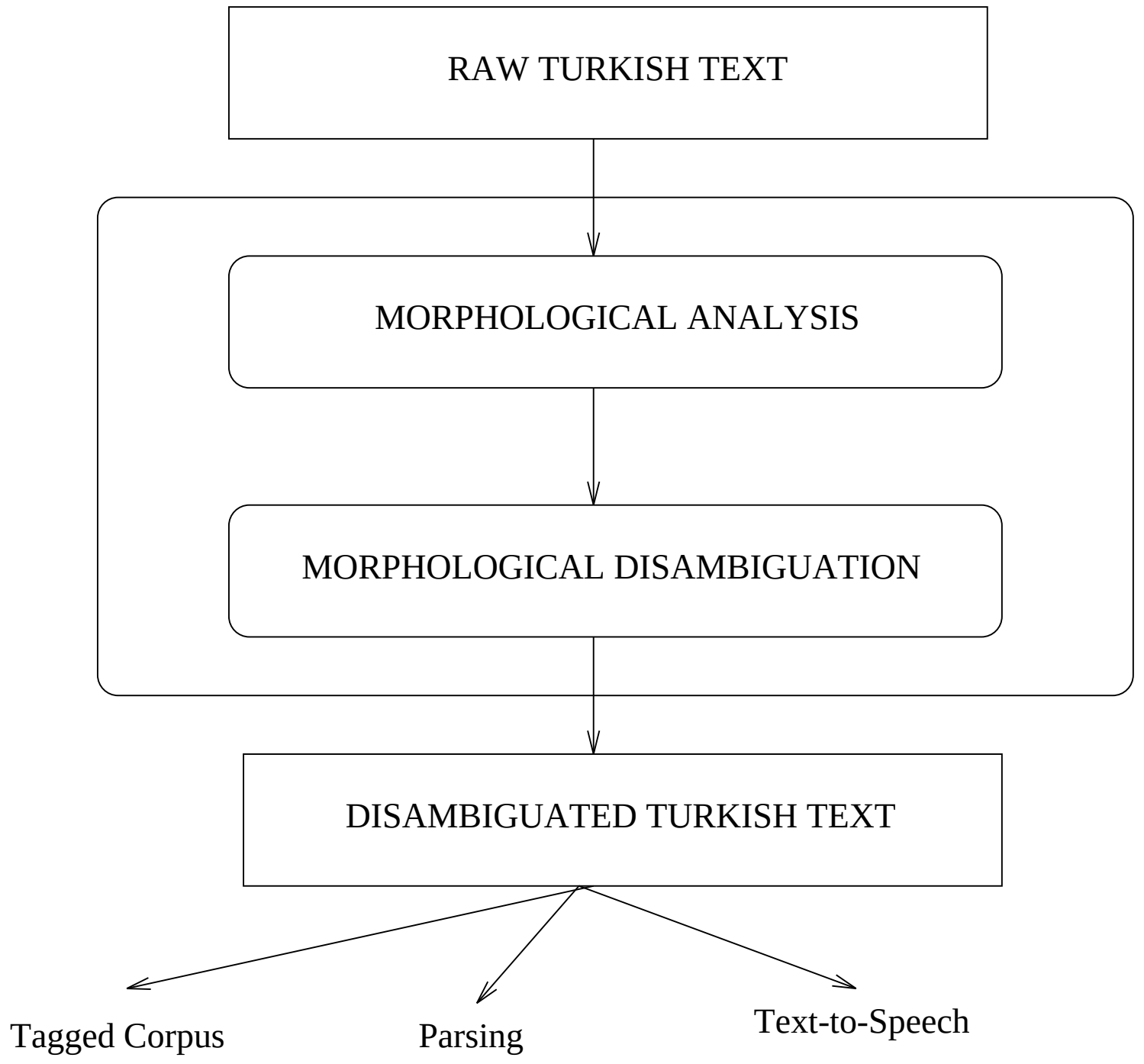

Figure 1.1: The place of morphological disambiguation in an abstract context.

[1]These metrics will be defined in detail in the next Chapter, but attaining both a 100% recall and 100% precision concurrently is the ultimate desired goal.

In contrast to languages like English, for which there is a very small number of possible word forms with a given root word, and a small number of tags associated with a given lexical form, languages like Turkish or Finnish with very productive agglutinative morphology where it is possible to produce thousands of forms (or even millions [13]) for a given root word, and this poses a challenging problem for morphological disambiguation. In English, for example, a word such as *make* or *set* can be verb or a noun. In Turkish, even though there are ambiguities of such sort, the agglutinative nature of the language usually helps resolution of such ambiguities due to restrictions on morphotactics. On the other hand, this very nature introduces another kind of ambiguity, where a lexical form can be morphologically interpreted in many ways, some with totally unrelated roots and morphological features, as will be exemplified in the next chapter.

The previous approach to tagging and morphological disambiguation for Turkish text had employed a constraint-based approach [24, 19] along the general lines of similar previous work for English [15, 26, 27, 28, 29, 30]. Although the results obtained there were reasonable, the fact that all constraint rules were hand crafted, posed a rather serious impediment to the generality and improvement of the system.

This thesis presents the morphological disambiguation of a Turkish text, based on constraints. The tokens, on which the disambiguation will be performed are determined using a preprocessing module, which will be covered in detail in Chapter 3.

Although we have used a constraint-based approach, we also make use of some constraint rules that are learned by a learning module. This module is capable of incrementally proposing and evaluating additional (possibly corpus dependent) constraints for disambiguation of morphological parses using the constraints imposed by unambiguous contexts. These rules choose or delete parses with specified features. This learning is achieved using a corpus, which is first disambiguated by the hand-crafted rules. In certain respects, our approach has been motivated by Brill's recent work [6], but with the observation that his transformational approach is not directly applicable to languages like Turkish, where all tags associated with forms are not predictable in advance.

In our approach, we use the following sources of information:

- Linguistic constraints,
- Contextual statistics and
- Root word preference statistics.

The following chapter presents an overview of the morphological disambiguation problem, highlighted with examples from Turkish in addition to the approaches to part-of-speech tagging and morphological disambiguation with the evaluation metrics, like recall, precision and ambiguity. Chapter 3 describes the details of our approach. The experimental results are presented in Chapter 4, with a discussion on the results. The last chapter concludes this thesis.

# Chapter 2

# Tagging and Morphological Disambiguation

In almost all languages, words are usually ambiguous in their parts-of-speech or other lexical features, and may represent lexical items of different syntactic categories, or morphological structures depending on the syntactic and semantic context. Part-of-speech (POS) tagging involves assigning every word its proper part-of-speech based upon the context the word appears in. In English, for example a word such as *set* can be a verb in certain contexts (e.g., He *set* the table for dinner) and a noun in some others (e.g., We are now facing a whole *set* of problems). According to Church, it is commonly believed that most words have just one part-of-speech, and that the few exceptions such as *set* are easily disambiguated by the context in most cases [7]. But in contrast, lexical disambiguation is a major issue in computational linguistics. Introductory texts are full of ambiguous sentences, where no amount of syntactic parsing will help, such as in the sentences:

| | | | | |
|---|---|---|---|---|
| Time | flies | like | an | arrow |
| NOUN | VERB+AOR | PREP | DET | NOUN |
| NOUN | NOUN+PLU | VERB | DET | NOUN |
| Flying | planes | can | be | dangerous |
| ADJ | NOUN+PLU | MODAL | VERB | ADJ |
| VERB | NOUN+PLU | MODAL | VERB | ADJ |

In Turkish, there are ambiguities of the sort above. However, the agglutinative nature of the language usually helps resolution of such ambiguities due to the restrictions on morphotactics. On the other hand, this very nature introduces another kind of ambiguity, where a whole lexical form can be morphologically interpreted in many ways not predictable in advance. For instance, our full-scale morphological analyzer for Turkish returns the following set of parses for the word *oysa*:[1]

```
1. [[CAT=CONN][ROOT=oysa]]
   (on the other hand)

2. [[CAT=NOUN][ROOT=oy][AGR=3SG][POSS=NONE][CASE=NOM]
   [CONV=VERB=NONE][TAM1=COND][AGR=3SG]]
   (if it is a vote)

3. [[CAT=PRONOUN][ROOT=o][TYPE=DEMONS][AGR=3SG][POSS=NONE][CASE=NOM]
   [CONV=VERB=NONE][TAM1=COND][AGR=3SG]]
   (if it is)

4. [[CAT=PRONOUN][ROOT=o][TYPE=PERSONAL][AGR=3SG][POSS=NONE][CASE=NOM]
   [CONV=VERB=NONE][TAM1=COND][AGR=3SG]]
   (if s/he is)

5. [[CAT=VERB][ROOT=oy][SENSE=POS][TAM1=DES][AGR=3SG]]
   (wish  s/he would  carve)
```

On the other hand, the form *oya* gives rise to the following parses:

[1]Glosses are given as linear feature value sequences corresponding to the morphemes (which are not shown). The feature names are as follows: `CAT`-major category, `TYPE`-minor category, `ROOT`-main root form, `AGR` -number and person agreement, `POSS` - possessive agreement, `CASE` - surface case, `CONV` - conversion to the category following with a certain suffix indicated by the argument after that, `TAM1`-tense, aspect, mood marker 1, `SENSE`-verbal polarity, `DES`- desire mood, `IMP`-imperative mood, `OPT`- optative mood, `COND`-Conditional

1. `[[CAT=NOUN][ROOT=oya][AGR=3SG][POSS=NONE][CASE=NOM]]`
   `(lace)`

2. `[[CAT=NOUN][ROOT=oy][AGR=3SG][POSS=NONE][CASE=DAT]]`
   `(to the vote)`

3. `[[CAT=VERB][ROOT=oy][SENSE=POS][TAM1=OPT][AGR=3SG]]`
   `(let him carve)`

and the form *oyun* gives rise to the following parses:

1. `[[CAT=NOUN][ROOT=oyun][AGR=3SG][POSS=NONE][CASE=NOM]]`
   `(game)`

2. `[[CAT=NOUN][ROOT=oy][AGR=3SG][POSS=NONE][CASE=GEN]]`
   `(of the vote)`

3. `[[CAT=NOUN][ROOT=oy][AGR=3SG][POSS=2SG][CASE=NOM]]`
   `(your vote)`

4. `[[CAT=VERB][ROOT=oy][SENSE=POS][TAM1=IMP][AGR=2PL]]`
   `(carve it!)`

However, the local syntactic context may help reduce some of the ambiguity above, as in:[2]

| | |
|---|---|
| sen-in | **oy-un** |
| PRON(you)+GEN | NOUN(vote)+POSS-2SG |
| *'your vote'* | |
| **oy-un** | reng-i |
| NOUN(vote)+GEN | NOUN(color)+POSS-3SG |
| *'color of the vote'* | |

[2]With a slightly different but nevertheless common glossing convention.

| | |
|---|---|
| **oyun** | reng-i |
| NOUN(game) | NOUN(color)+POSS-3SG |
| *'game color'* | |

using some very basic noun phrase agreement constraints in Turkish. In the first case, the two word form a simple noun phrase (NP) and the constraints are such that the possessive marking on the second form has to be the same as the agreement of the first instance, which is also case marked genitive, while in the second case, ambiguity still can not be resolved, since both *color of the vote* and *game color* readings are possible. Such ambiguities can be resolved, using the root word preference statistics. Obviously in other similar cases, it may be possible to resolve the ambiguity completely.

There are also numerous other examples of word forms where productive derivational processes come into play:[3]

```
geliSindeki
(at the time of his/your coming)

1. [[CAT=VERB][ROOT=gel][SENSE=POS]
    (basic form)
   [CONV=NOUN=YIS][AGR=3SG][POSS=2SG][CASE=LOC]
    (participle form)
   [CONV=ADJ=REL]]
    (final adjectivalization by the relative ``ki'' suffix)

2. [[CAT=VERB][ROOT=gel][SENSE=POS]
    (basic form)
   [CONV=NOUN=YIS][AGR=3SG][POSS=3SG][CASE=LOC]
    (participle form)
   [CONV=ADJ=REL]]
    (final adjectivalization by the relative ``ki'' suffix)
```

Here, the original root is verbal but the final part-of-speech is adjectival. In general, the ambiguities of the forms that come before such a form in text can be

[3]Upper cases in morphological output indicates one of the non-ASCII special Turkish characters: e.g., `G` denotes ğ, `U` denotes ü, etc.

resolved with respect to its original (or intermediate) parts-of-speech (and inflectional features), while the ambiguities of the forms that follow can be resolved based on its final part-of-speech. Consider the noun phrase:

senin gelişindeki gecikme
`your come+INF+POSS-2SG delay`
*'the delay in your coming'*

In this phrase, the previous word, *senin* (your) implies that the possessive marker in the next token *gelişindeki* is 2SG, instead of 3SG, and the final category of the token *gelişindeki*, i.e. adjective, implies that the next word *gecikme* (delay) is a noun, instead of a verb with an imperative reading, meaning '*do not be late!*'.

## 2.1 Approaches to Tagging and Morphological Disambiguation

Part-of-speech taggers and morphological disambiguators generally use two kinds of approaches:

- *Constraint-based Approaches*, where a large number of hand-crafted linguistic constraints are used to eliminate impossible tags or morphological parses for a given word in a given context.
- *Statistical Approaches*, where a large corpora is used to train a statistical model which then to be used to tag a new text.

Brill introduced a method to induce the constraints from tagged corpora, called *transformation-based error-driven learning* [2, 3, 4, 5]. Recently, this method is extended so that, no tagged corpus is needed [6].

It is also possible to use some or all of these approaches together in a morphological disambiguation system, which we investigate in this thesis.

### 2.1.1 Constraint-based Approaches

The earliest tagger was developed in 1963 by Klein and Simmons [17], and this was an initial categorial tagger rather than a disambiguator. Its primary goal was to avoid *the labor of constructing a very large dictionary*; which was more important in those days. Their algorithm uses a palette of 30 categories, and it is claimed that, this algorithm correctly and unambiguously tags about 90% of the words in several pages of the Golden Book Encyclopedia. The algorithm first seeks each word in dictionaries of about 400 function words, and of about 1,500 words which are exceptions to the computational rules used. The program, then, checks for suffixes and special characters as clues. Finally, context frame tests are applied. These work on scopes bounded by unambiguous words, like later algorithms. However, Klein and Simmons impose an explicit limit of three ambiguous words in a row. For each such *span* of ambiguous words, the pair of unambiguous categories bounding it is mapped into a list. This list includes all known sequences of tags occurring between the particular bounding tags; all such sequences of the correct length become candidates. The program then matches the candidate sequences against the ambiguities remaining from earlier steps of the algorithm. When only one sequence is possible, disambiguation is successful. This approach works, because since the number of different POS categories is too limited, and this reduces the ambiguity obviously, and also their test sample is a very small text, a larger sample would contain both low frequency ambiguities and many long spans with a higher probability.

The next important tagger, TAGGIT, was developed by Greene and Rubin in 1971 [11]. This tagger correctly tags approximately 77% of the million words in the Brown Corpus (the rest is completed by human post-editors). TAGGIT uses 86 part-of-speech (POS) tags. TAGGIT first consults an exception dictionary of about 3,000 words, which contains all known closed-class words among other items. It then handles various special cases, such as special symbols, capitalized words, etc. The word's ending is then checked against a suffix list of about 450 strings. If TAGGIT has not assigned some tag(s) after these steps, the word is tagged *noun, verb or adjective* in order that the disambiguation routine may have something to work with. The disambiguation routine then applies a set of 3,300 context frame rules. Each rule, when its context is satisfied, has the effect of deleting one or more candidates from the list of possible tags for one word. Each rule can include a context of up to two unambiguous words on each side of the ambiguous word to which rule is being applied.

TAGGIT is important in the sense that, it is the first tagger, that deals with such a large and varied corpus. The decision of examining only one ambiguity at a time with up to two unambiguous words on either side is derived from an experiment made on a sample text of 900 sentences. Moreover, while less than 25% of TAGGIT's context frame rules are concerned with only the immediate preceding or succeeding word, these rules were applied in about 80% of all attempts to apply rules.

A very successful constraint-based approach for morphological disambiguation was developed in Finland. From 1989 to 1992, four researchers – Fred Karlsson, Arto Anttila, Juha Heikkila and Atro Voutilainen – from the Research Unit for Computational Linguistics at the University of Helsinki participated in the ESPRIT II project No. 2083 SIMPR (Structured Information Management: Processing and Retrieval). The task was to make an operational parser for running English text, mainly for information retrieval purposes. The parsing framework, known as Constraint Grammar was originally proposed by Karlsson upon which the English Constraint Grammar description ENGCG was written [14].

In this framework, the problem of parsing was broken into seven subproblems or *modules*, four of them are related to morphological disambiguation, the rest are used for parsing the running text.

1. *Preprocessing*: This part deals with idioms and other more or less fixed multi-word expressions like *in spite of*, etc. We have also a similar called, *preprocessor*, which is defined in the next chapter.

2. *Morphological Analysis*: Koskenniemi's two-level model was used in the morphological analyzer [18].

3. *Local Disambiguation*: This step precedes context-based morphological disambiguation and deals with the local inspection of the current token without invoking any contextual information. An example rule is: *Choose the parse* *which includes the minimum number of derivations.* It is claimed that this principle is very close to perfect.

4. *Context–Dependent Disambiguation Constraints*: Ambiguity is resolved using some context-dependent constraints. Each constraint is a quadruple consisting of domain, operator, target and context condition(s). For example:

    (@w=0 "PREP" (-1 DET))

    state that if a word (@w) has a reading with the feature "PREP", this very reading is discarded (=0) if the preceding word (i.e. the word position -1) has a reading with feature "DET"

The constraint-based morphological disambiguator for English was implemented by Voutilainen [25, 26, 27, 28, 29, 30]. The present grammar consists of 1,100 constraints. Of all words, 93-97% became unambiguous and at least 99.7% of all words retained the contextually most appropriate morphological reading with 1.04 morphological readings per word on the average after morphological disambiguation, and with an optionally applicable heuristic grammar of 200 constraints resolves about half of the remaining ambiguities 96-97% reliably. These numbers also include errors due to the ENGTWOL lexicon which contains 80,000 lexical entries, and *morphological heuristics*, a rule-based module that assigns ENGTWOL-style analyses to those words not represented in ENGTWOL itself. Currently, ENGCG contains no module that disambiguates the remaining 2-4%. If a blind-guessing module was used, the overall precision and recall of the entire system with no ambiguity in the output would be claimed as 98% or a little more[4].

Later, Tapainen from Rank Xerox Research Center in France, and Voutilainen combined ENGCG and Xerox Tagger. In a 27,000 word unseen text, they reached an accuracy of about 98.5%, with no ambiguous word. This result is significantly better than 95-97% accuracy which state-of-the-art statistical taggers reach alone.

There are several other part-of-speech disambiguators for English. Among the best known are CLAWS1 by the UCREL team (Garside, Leech, Sampson,

[4] Definitions of *recall* and *precision* can be found in the Section 2.2

Marshall) and Parts-of-speech by Church.[5] An experiment was performed on 5 unseen texts, with a total of 2167 words. The results are summarized in the Table 2.1.

| Method | Recall | Precision |
|---|---|---|
| CLAWS | 96.95 | 96.95 |
| Parts-of-speech | 96.21 | 96.21 |
| ENGCG | 99.77 | 95.54 |

Table 2.1: Comparison of the taggers

Kuruöz and Oflazer's work, deals with the morphological disambiguation of Turkish by using some constraints [19, 24]. Although the results obtained there are reasonable, the fact that all constraint rules are hand-crafted, has posed a rather serious impediment to the generality and improvement of the system. But this work is important in the sense that it formed a framework for this thesis, and experiences gained in that work lead us to the ideas implemented and presented in this thesis. It is claimed that their morphological disambiguator can disambiguate about 97% to 99% of the texts accurately with very minimal user intervention, but that system lacks many features of the approach presented in this thesis.

### 2.1.2 Statistical Approaches

In 1983, a tagging algorithm for Lancaster-Oslo-Bergen (LOB) Corpus, called CLAWS was described [21]. The main innovation of CLAWS was the use of a matrix of *collocational probabilities*, indicating the relative likelihood of co-occurrence of all ordered pairs of tags. The matrix could be mechanically derived from any pre-tagged corpus. CLAWS used a large portion of the Brown Corpus, with 200,000 words. The tag set is very similar to TAGGIT, but somewhat larger, at about 130 tags. The dictionary is derived also from the Brown Corpus. It contains 7,000 rather than 3,000 entries and 700 rather than 450 suffixes.

When an ambiguous token is encountered, the algorithm computes the probabilities of each path using a collocation matrix. Each path is a combination of selecting one tag for each ambiguous token which occur side by side. The path with the maximal probability is chosen.

[5] Both of these systems are statistical, and will be mentioned in the following subsection

Before the disambiguation, a program called IDIOMTAG is used to deal with the idiosyncratic word sequences, like *in-spite-of*. This module tags approximately 1% of the running text.

CLAWS has been applied to the entire LOB Corpus with an accuracy of between 96% and 97%. The contribution of IDIOMTAG is 3%.

Later, in 1988, DeRose [10] proposed an advanced version of CLAWS, called VOLSUNGA. This algorithm reached an accuracy of 96% without an idiom listing, and is claimed to be more time and space efficient than CLAWS.

Church has described a successful probabilistic tagger, which uses also a tagged corpus, namely Brown Corpus [7]. This tagger makes use of both lexical and contextual probabilities. The gist of this work can be explained best by an example. Consider the sentence:[6]

---

[6]Church tells of an interesting story in an interview, which is published in the EACL special of Ta!, the Dutch students' magazine for computational linguistics [8].

> One day I was going to give a tutorial to the speech guys on chart parsing. I just put together a very tiny parser for pedagogical purposes, maybe a quarter inch of code, I had the simplest possible grammar I could come up with, I'd use the simplest possible sentence I could think of and I had a complete trace of the whole thing; we could go through this in the lecture. The sentence I first picked was: 'I saw a bird'. What I did though, just for fun, I replaced my simple little lexicon with Webster's Dictionary, which I happened to have on line. I tried the sentence 'I saw a bird', and it came out ambiguous. Not only was it what you would hope, but it also came out as a noun phrase. 'I' and 'a' are letters of the alphabet, and 'saw' and 'bird' could both be nouns. So four nouns, and I had the rule that said: NP goes to any number of nouns. So if that isn't a good example, let me try an easier one. How about 'I see a bird'. That one couldn't be ambiguous. Well, it turns out it's exactly the same. Why? Well, 'see' is listed in the Websters Dictionary as the holy See. And it dawned on me that the problem here is, is that the dictionary is just full of absurdly unlikely things. Look in the Brown corpus and 'I' and 'a' don't appear as nouns anywhere in it. The idea was that there was something fundamentally wrong with the idea that everything that's in the dictionary is on an equal footing.
>
> At that point I started looking at these statistical methods for doing part of speech tagging and they just cleaned up. At the time most people weren't doing very well with part of speech tagging. They had all declared it a solved problem. They had also declared all of syntax a solved problem. I was told when I started working on CL that it was no longer possible to get a PhD thesis in any kind of computational syntax. All the problems have been solved. And then ten years after that... there I was, really nervous about getting up in front of the ACL and saying that I had a statistical method on part of speech. Not only of course was statistics heresy, but

I see a bird

The lexical probabilities gathered for this sentence are as follows:

| **Word** | **Parts of Speech** | |
|---|---|---|
| I | PRONOUN (5837) | NOUN (1) |
| see | VERB (771) | INTERJECTION (1) |
| a | ARTICLE (23013) | PREPOSITION (French)(6) |
| bird | NOUN (26) | |

Table 2.2: Lexical frequencies of the words

Church states that, for all practical purposes every word in a sentence is unambiguous , however, according to the Webster'e Dictionary, every word is ambiguous. This is the situation shown in this example sentence. The word *I* is said to be a noun since it is a character in the alphabet; the word *a* might be a French preposition, and the word *see* can be used as an interjection. Also, these words have some other readings in the dictionary. For example the word *bird* can also be used as an intransitive verb, and *a* is also a noun since it is also a character in the alphabet.

The lexical probability of a word is calculated in the obvious way. For example, the lexical probability that *see* is a VERB is:

$$Prob(VERB|see) = \frac{freq(VERB|see)}{freq(see)} = \frac{771}{772}$$

The contextual probability, the probability of observing part of speech X given the following two parts of speech Y and Z, is estimated by dividing the trigram frequency XYZ by the bigram frequency YZ. Thus, for example, the probability of observing a VERB before an ARTICLE and a NOUN is estimated as:

$$\frac{freq(VERB, ARTICLE, NOUN)}{freq(ARTICLE, NOUN)}$$

in addition to that I was talking about a problem that had long since been declared solved. And here I was going to say : Well you may not like the methods, and you may not have known it was a problem, but ...

A search is performed in order to find the assignment of part of speech tags to words that optimizes the product of the lexical and contextual probabilities. Conceptually, the search enumerates all possible assignments of parts of speech to input words. For example, in the above example, there are 4 input words, three of which are 2 way ambiguous, producing a set of 2*2*2*1=8 possible part of speech assignments of input words. Each of them is then scored by the product of the lexical probabilities and the contextual probabilities, and the best sequence is selected.

Church claims an accuracy of 95% to 99%. But there is no detail how these percentages are obtained, and the given range is so large, that it is almost impossible to make a comment on this system. But his system became very popular, and formed the basis of statistical computational linguistics.

Among the other studies in developing automatically trained part of speech taggers, that use Hidden-Markov-Models, Cutting et al., Merialdo, DeRose, and chedel et al. can be considered [9, 10, 22, 31].

### 2.1.3 Transformation-Based Tagging

During his Ph.D. thesis in the University of Pennsylvania, Eric Brill presented an innovative learning algorithm, called as *transformation-based error-driven learning* [2, 3, 4, 5]. A *transformation* is an instantiation of a predefined template, depending on the application it is used in. The aim of this algorithm is to automatically discover the structural information about a language using corpus. This approach has been applied to a number of natural language problems, including part of speech tagging, prepositional phrase attachment disambiguation and syntactic parsing.

In one sentence this approach can be explained as:

> The distribution of errors produced by an imperfect annotator is examined to learn an ordered list of transformations that can be applied to provide an accurate structural annotation.

Learning natural language from large corpora is not a new concept. It is worthwhile considering whether corpus-based learning algorithms can be implemented, because of the following reasons:

- Building a knowledge base manually is a very expensive, difficult process,
- These knowledge bases have not been used effectively in structurally parsing the sentences, except the highly restricted domains,
- The advent of very fast computers and the availability of annotated on-line corpora.

**Brill's Algorithm**

This algorithm starts with a small structurally annotated corpus and a larger unannotated corpus, and uses these corpora to learn an ordered list of transformations that can be used to accurately annotate fresh text.

The system begins in a language-naive start state. From the start state, it is given an annotated corpus of text as input and it arrives at an end state. In this work, the end-state is an ordered list of transformations for each particular learning module. Transformations depend on predefined transformation templates. The learner is defined by the set of allowable transformations, the scoring function used for learning and the search method carried out in learning. Basically, greedy search is used in learning. At each stage of learning, the learner finds the transformation whose application to the corpus results in the best scoring corpus. Learning proceeds on the corpus, that results from applying the learned transformation. This continues until no more transformations can be found whose application results in improvement. Once an ordered list of transformations has been learned, new text is annotated by simply applying each transformation, in order, to the entire corpus.

Figure 2.1 summarizes the framework of this approach. Unannotated text is first presented to the system. The system uses its prespecified initial state knowledge to annotate the text. This initial state can be at any level of sophistication. For example, the initial state can assume that, every unknown word is a noun.

Rather than manually creating a system with mature linguistic knowledge, the system begins in a naive initial state and then learns linguistic knowledge automatically from a corpus. After the text is annotated by the initial state annotator, it is then compared to the true annotation assigned in the manually annotated training corpus.

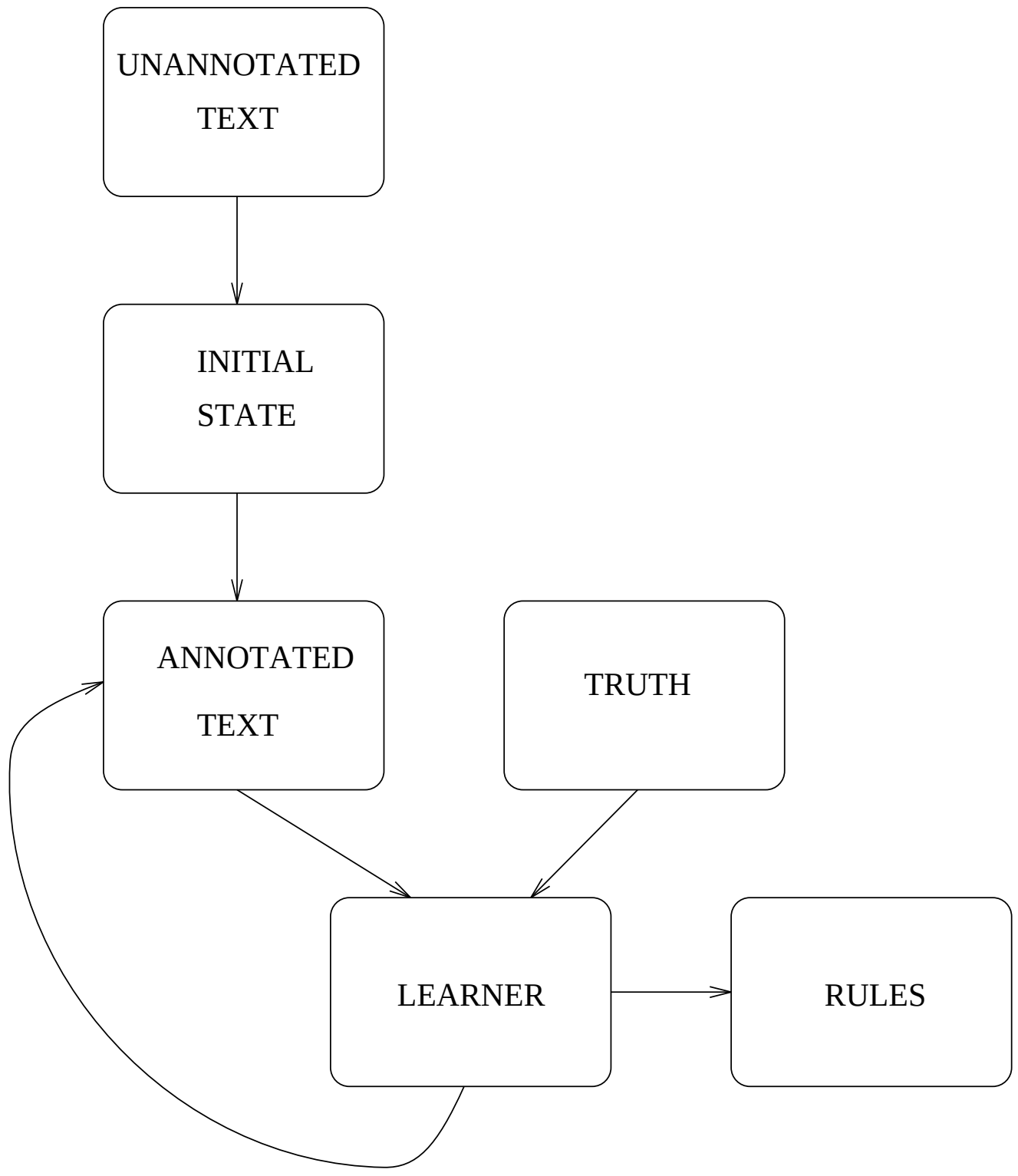

Figure 2.1: Transformation-Based Error-Driven Learning.

In the empirical evaluation, Brill uses 3 different manually created corpora: The Penn Treebank, original Brown Corpus and a corpus of old English. At most 45,000 words of annotated text were used in the experiments. By comparing the output of the naive start state annotator to the true annotation indicated in the manually annotated corpus, something can be learned about the errors produced by the naive annotator. Transformations then can be learned which can be applied to the naively annotated text to make it resemble the manual annotation more. A set of transformation templates specifying the types of transformations which can be applied to the corpus is prespecified. In all of the learning modules described in this dissertation, the transformation templates are very simple, and do not contain any deep linguistic knowledge. The number of transformation templates is also small. These templates contain uninstantiated variables. For example, in the template:

Change a tag from X to Y, if the previous tag is Z.

X, Y and Z are variables. All possible instantiations of all specified templates define the set of allowable transformations.

Some transformations result in better, and some result in worse accuracy. So the system looks for the best transformation and adds it to its transformation list. The criteria is the number of errors in the automatically annotated text.

Learning stops when no more effective transformations can be found, meaning either no transformations are found that improve performance or none improve performance above some threshold.

An example application of this algorithm is outlined in Figure 2.2. The initial corpus results in 532 errors, found by comparing the annotated corpus to a manually annotated corpus. At time T-0, all possible transformations are tested. Transformation T-0-1 (transformation T1 applied at time 0) is applied to the corpus, resulting in a new corpus, Corpus 1.1. There are 341 errors in this corpus. Transformation T-0-2, obtained by applying transformation T2 to corpus C-0, results in Corpus-1-2, which has 379 errors. The third transformation results in an annotated corpus with 711 errors. Because Corpus-1-1 has the lowest error rate, the transformation T1 becomes the first learned transformation, and learning continues on Corpus-1-1. Figure 2.3 shows the resulting corpora at each iteration of this algorithm.

Brill used both lexical and contextual information. The templates used in lexical information is as follows:

- Change the most likely tag to X if:
  - Deleting (adding) the prefix (suffix) x, $|x| < 5$ results in a word.
  - The first (last) 1,2,3 or 4 characters of the word are x.

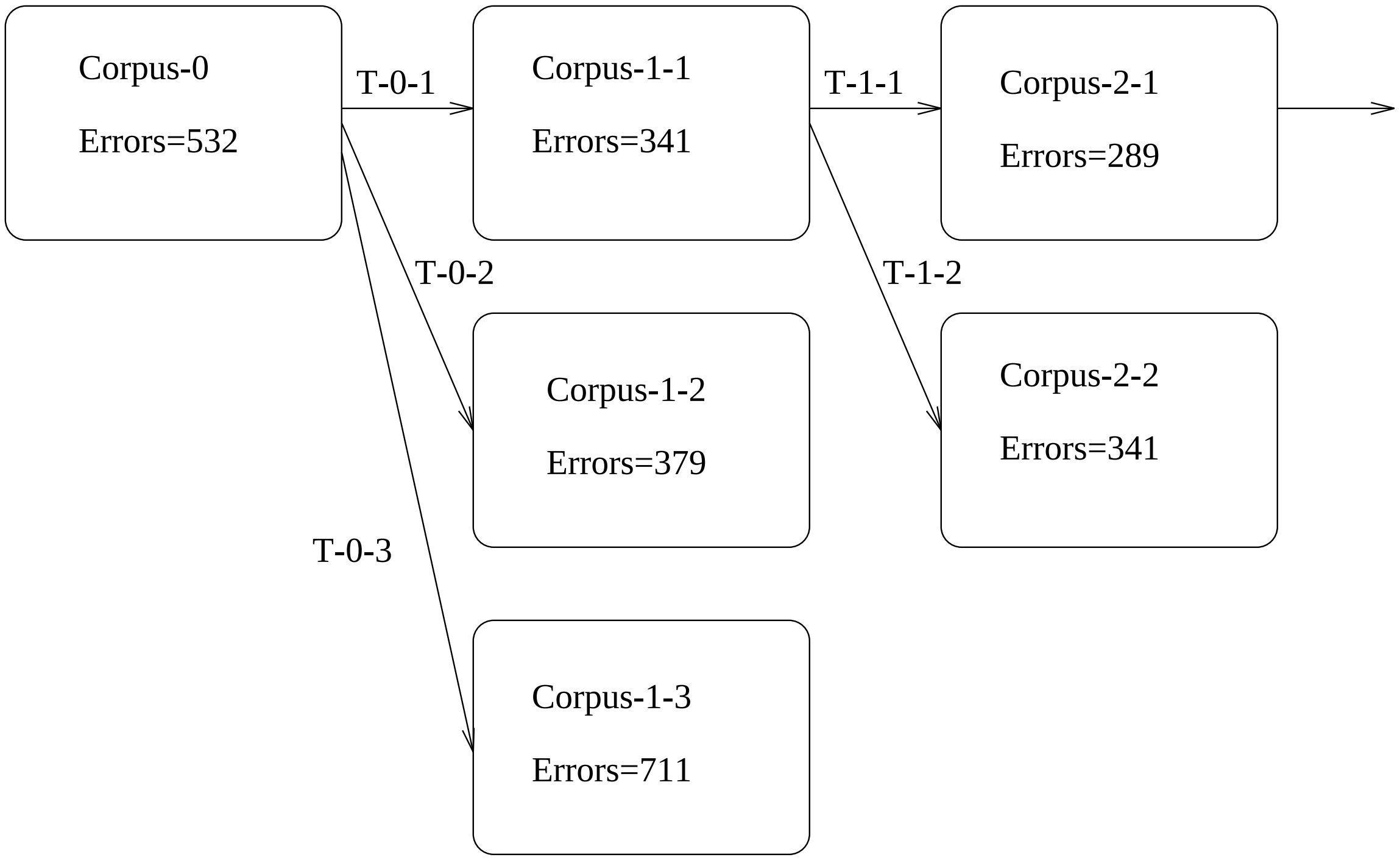

Figure 2.2: Learning Transformations.

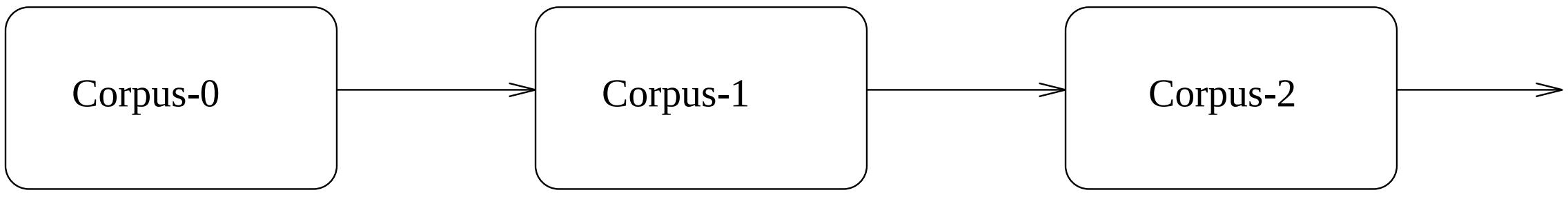

Figure 2.3: Applying Transformations.

- Adding the character string x as a prefix (suffix) results in a word ($|x| < 5$).
  - Word Y ever appears immediately to the left (right) of the word.
  - Character Z appears in the word.

- All of the above transformations modified to say: Change the most likely tag from Y to X if...

Some learned transformations are shown in Table 2.3.

| **From** | **To** | **Condition** |
|---|---|---|
| ? | Plural Noun | Suffix is 's' |
| Noun | Proper Noun | Appear at the start of sentence |
| ? | Past Part. Verb | Suffix is 'ed' |
| ? | Cardinal Number | Appear to the right of '$' |
| ? | Present Part. Verb | Suffix is 'ing' |

Table 2.3: The first 5 transformations from the Wall Street Journal Corpus

After learning the lexical transformations, the next step is to use contextual cues to disambiguate word tokens. This is nothing, but another application of transformation-based error-driven learning. The following templates are used:

Change a tag from X to Y if:

- The previous (following) word is tagged as Z.
- The previous word is tagged as Z, and the following as W.
- The following (preceding) 2 words are tagged as Z.
- one of the 2 (3) preceding (following) words is tagged as Z.
- The word, two words before (after) is tagged as Z.

An example of a learned transformation is:

Change the tag of a word from VERB to NOUN if the previous word is a DETERMINER.

Later in 1994, Brill extended this learning paradigm to capture relationships between words by adding contextual transformations that could make reference to the words as well as part-of-speech tags. Used transformation templates are as follows:

Change a tag from X to Y if:

- The preceding (following) word is W.
- The current word is W and the preceding (following) word is X.
- The current word is W and the preceding (following) word is tagged as Z.

Some results obtained from the experiments are summarized in Table 2.4:

| **Method** | **Corpus Size** | **# Rules** | **Accuracy %** |
|---|---|---|---|
| Statistical | 64 K | 6170 | 96.3 |
| Statistical | 1 M | 10000 | 96.7 |
| w/o Lex. Rules | 600 K | 219 | 96.9 |
| with Lex. Rules | 600 K | 267 | 97.2 |

Table 2.4: Results of the tagger

Transformation-based approach is different from other approaches in language learning in the following aspects:

- There is very little linguistic knowledge, and no language-specific knowledge built into the system.
- Learning is statistical, but only *weakly* so.
- The end-state is completely symbolic.
- A small annotated corpus is necessary for learning to succeed.

The run-time of the algorithm is $O(|op| \times |env| \times |n|)$ where $|op|$ is the number of allowable transformation operations, $|env|$ is the number of possible triggering environments, and $|n|$ is the training corpus size (the number of word types in the annotated lexical training corpus). Applying the transformations to the corpus runs in linear time, $O(|T| \times |n|)$, where $|T|$ is the transformation size, and $|n|$ is the size of the test corpus.

The accuracy of the algorithm is not too high, it can reach only to 97%, this is almost the same as other statistical tagging methods. The important point is that, his approach can be used with rule-based approaches, since it produces rules with an order, but note that, this is a greedy algorithm, it suffers from the *horizon effect*, that is, since you can see only one transformation ahead, you cannot catch a better transformation, more than one step ahead.

### Unsupervised Learning of Disambiguation Rules

In 1995, Brill improved this algorithm so that, it no longer requires a manually annotated training corpus [6]. Instead, all needed is the allowable part-of-speech tags for each token, and the initial state annotator tags each token in the corpus with a list of all allowable tags.

The main idea can be explained best with the following example. Given the sentence:

> The *can* will be crushed.

using an unannotated corpus it could be discovered that of the unambiguous tokens (i.e. that have only one possible tag) that appear after *the* in the corpus, nouns are much more common than verbs or modals. From this, the following rule could be learned:

> Change the tag of a word from (*modal OR noun OR verb*) to noun if the previous word is *the*.

Unlike supervised learning, in this approach, main aim is not to change the tag of a token, but reduce the ambiguity, by choosing a tag for the words in a particular context. So all learned transformations have the form:

Change the tag of a word from $\chi$ to Y in context C

where $\chi$ is a set of two or more part-of-speech tags, and Y is one of them.

Brill used 4 templates in his implementation:

Change the tag of a word from $\chi$ to Y if:

- The previous tag is T.
- The next tag is T.
- The previous word is W.
- The next word is W.

The scoring function is also different from supervised approach. With unsupervised learning, the learner does not have a gold standard training corpus with which accuracy can be measured. Instead, unambiguous words are used in the scoring. In order to score the transformation *Change the tag of a word from* $\chi$ *to Y in context C*, the following is done. Compute:

$$R = argmax_Z \frac{count(Y)}{count(Z)} \times incontext(Z, C, )$$

where $Z \in \chi, Z \neq Y$. The score of the candidate rule is then computed as:

$$Score = incontext(Y, C) - \frac{count(Y)}{count(R)} \times incontext(R, C)$$

A good transformation for removing part-of-speech ambiguity of a word is the one for which one of the possible tags appears much more frequently as measured by unambiguously tagged words than all others in the context, after adjusting for the differences in relative frequency between the different tags. In each learning iteration, the learner searches for the transformation which maximizes this function. Learning stops when no positive scoring transformations can be found.

Brill reports an accuracy of 95.1%-96.0% in using unsupervised learning. Later he has completed this work by combining both supervised and unsupervised learning approaches. In that case, he has reached an accuracy of 96.8% with a training corpus size of 88,200 words.

The main advantage of unsupervised learning is that, it does not require a manually tagged training corpus. Intuitively, it can be thought that using unambiguous words in the scoring result in very insufficient results, but if the algorithm is modified as to terminate when the score is below some threshold, the learned rules are very interesting.

## 2.2 Evaluation Metrics

The main intent of our system is to achieve a *morphological ambiguity reduction* in the text by choosing for a given ambiguous token, a subset of its parses which are not disallowed by the syntactic context it appears in. It is certainly possible that a given token may have multiple correct parses, usually with the same inflectional features or with inflectional features not ruled out by the syntactic context. These can only be disambiguated usually on semantic or discourse constraint grounds.

We consider a token *fully disambiguated* if it has only one morphological parse remaining after automatic disambiguation. We consider a token as correctly disambiguated, if one of the parses remaining for that token is the *correct* intended parse.[7]

In this thesis, we use the metrics of the ENGCG team from the University of Helsinki:

*Recall:* The ratio "received appropriate readings/intended appropriate readings"

*Precision:* The ratio "received appropriate readings/all received readings"

Thus, a recall of 100% means that all tokens have received an appropriate reading, so initially before any disambiguation, (assuming no unknown words) recall is 100%.[8] A precision of 100% means that there is no superfluous reading, *noise* in the output. If recall and precision are same, then this value is called *accuracy*, which happens when all tokens have exactly one parse. The aim of a morphological disambiguator or a tagger is 100% accuracy.

[7] It is certainly possible that, a parse that is deleted may also be a valid parse in that context.

Let us explain these terms with an example. Consider the sentence:

| | | | |
|---|---|---|---|
| bunun | üzerinde | duralım | . |
| `PRONOUN(this)+GEN` | `NOUN(on)+POSS-3SG` | `VERB(focus)+OPT+1PL` | `PUNCT` |

*'Let's focus on this.'*

```
[[bunun,
[[cat:noun,root:bun,agr:'3SG',poss:'NONE',case:gen],
 [cat:noun,root:bun,agr:'3SG',poss:'2SG',case:nom],
 [cat:pronoun,root:bu,type:demons,agr:'3SG',poss:'NONE',case:gen],
 [cat:verb,root:bun,sense:pos,tam1:imp,agr:'2PL']]],

['Uzerinde',
[[cat:noun,root:'Uzer',agr:'3SG',poss:'2SG',case:loc],
 [cat:noun,root:'Uzer',agr:'3SG',poss:'3SG',case:loc]]],

[duralIm,
[[cat:noun,stem:[cat:adj,root:dural],suffix:none,agr:'3SG',poss:'1SG',case:nom],
 [cat:verb,stem:[cat:adj,root:dural],suffix:none,tam2:pres,agr:'1SG'],
 [cat:verb,root:dur,sense:pos,tam1:opt,agr:'1PL']]],

['.',
[[cat:punct,root:'.']]]].
```

The output of an ideal morphological disambiguator, with a 100% recall and precision would be as follows:

[8] In our system, we ignore the unknown words, called unknown also in the gold standard, but their effect is negligible.

```
[[bunun,
[[cat:pronoun,root:bu,type:demons,agr:'3SG',poss:'NONE',case:gen]]],

['Uzerinde',
[[cat:noun,root:'Uzer',agr:'3SG',poss:'3SG',case:loc]]],

[duralIm,
[[cat:verb,root:dur,sense:pos,tam1:opt,agr:'1PL']]],

['.',
[[cat:punct,root:'.']]]].
```

In this case, the number of intended appropriate readings, received appropriate readings and all received readings are same, namely 4, since there are 4 tokens. Now, assume that, our morphological disambiguator has an output of:

```
[[bunun,
[[cat:noun,root:bun,agr:'3SG',poss:'NONE',case:gen]]],


['Uzerinde',
[[cat:noun,root:'Uzer',agr:'3SG',poss:'3SG',case:loc]]],

[duralIm,
[[cat:verb,stem:[cat:adj,root:dural],suffix:none,tam2:pres,agr:'1SG'],
 [cat:verb,root:dur,sense:pos,tam1:opt,agr:'1PL']]],

['.',
[[cat:punct,root:'.']]]].
```

where, one token, *bunun*, is incorrectly disambiguated and another token *duralım* has 2 parses. Now, the number of *received appropriate readings* is 3 out of 4 *intended readings*, because the parses of one token do not contain the correct reading. So, recall is 3/4 or 66.67%. Totally, 5 parses are received, because one token has one extra parse. So, precision is 3/5 or 60%.

# Chapter 3

# Morphological Disambiguation

The morphological disambiguation of a Turkish text, explained in this thesis is based on constraints. The tokens, on which the disambiguation will be performed are determined using a preprocessing module.

Given a new text annotated with all morphological parses of the tokens, the initial choose and delete rules are applied first, then contextual and root word preference statistics are applied, and last of all, the learned choose and delete rules are used to discard further parses.

## 3.1 The Preprocessor

Early studies on automatic text tagging for Turkish had shown that some preprocessing on the raw text is necessary before analyzing the words in a morphological analyzer [19, 24].

This preprocessing module (shown in Figure 3.1) includes:

- *Tokenization*, in which raw text is split into its tokens, which are not necessarily separated by blank characters or punctuation marks;
- *Morphological Analyzer*, which is used for processing the tokens, obtained from the tokenization module, using the morphological analyzer;

- *Lexical and Non-lexical Collocation Recognizer*, in which lexical and non-lexical collocations are recognized and packeted;
- *Unknown Word Processor*, in which the tokens, which are marked as *unknown* after the lexical and non-lexical collocation recognizer, are parsed;
- *Format Conversion*, in which each parse of a token is converted into a hierarchical feature structure;
- *Projection*, in which each feature structure is projected on a subset of its features to be used in the training.

See Appendix A for a sample text, whose preprocessed form is presented in Appendix C.

### 3.1.1 Tokenization

This module is used to divide up the raw text into its tokens. A *token* belongs to one of:

- **Words**: e.g.,

  evde (at home)
  geliyorum (I am coming)

- **Numeric structures**: In most of the texts, numeric tokens are very frequent. It is also possible to have suffixes after these numeric tokens, for example, *32.si* or *32.'si*, (of the $32^{nd}$) are valid tokens we handle. The following numeric structures are handled in the tokenization and also in the morphological analysis:

  - **Cardinals**: e.g.,

    32542432

    1. [[CAT=ADJ][TYPE=CARDINAL][ROOT=32542432]]

    2. [[CAT=ADJ][TYPE=CARDINAL][ROOT=32542432][CONV=NOUN=NONE][AGR=3SG]

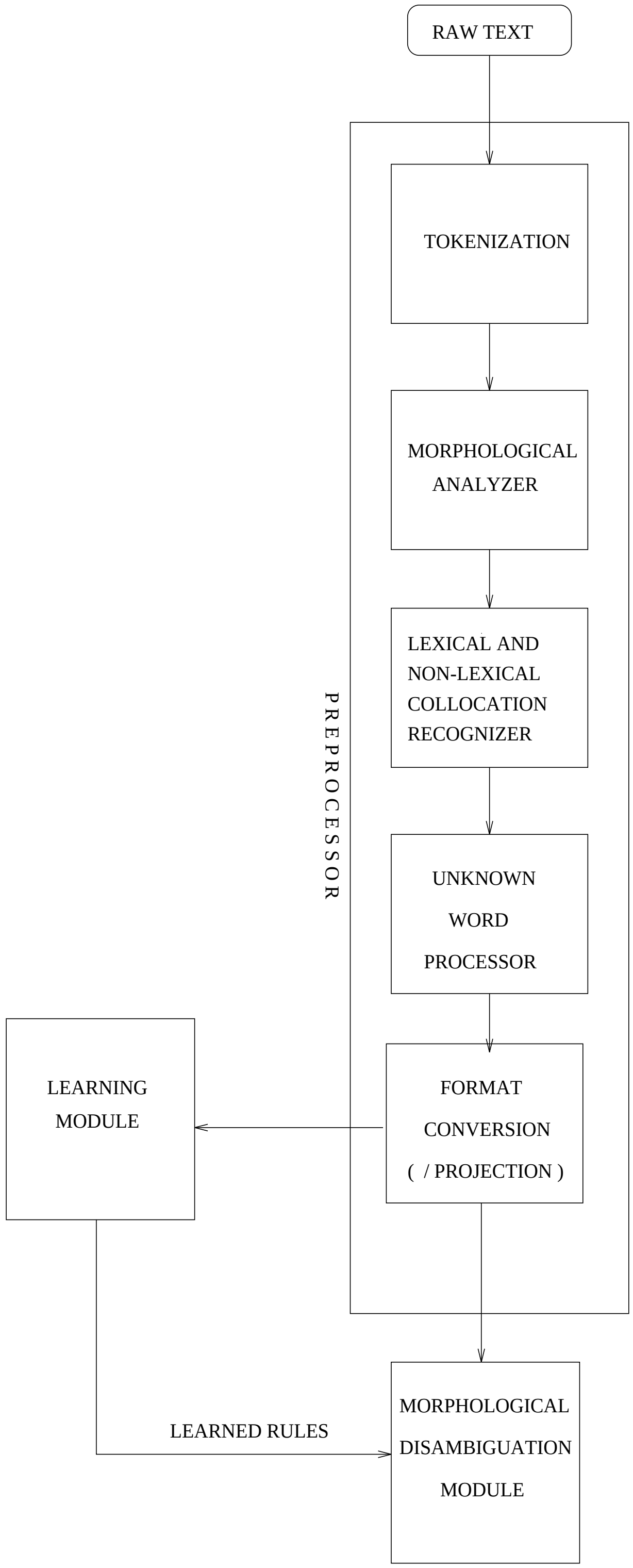

Figure 3.1: The structure of the preprocessor

32.542.432'nin

1. `[[CAT=ADJ][TYPE=CARDINAL][ROOT=32.542.432'][CONV=NOUN=NONE]`
   `[AGR=3SG][POSS=NONE][CASE=GEN]]`

2. `[[CAT=ADJ][TYPE=CARDINAL][ROOT=32.542.432'][CONV=NOUN=NONE]`
   `[AGR=3SG][POSS=2SG][CASE=GEN]]`

– **Ordinals**: e.g.,

3783.

1. `[[CAT=ADJ][TYPE=ORDINAL][ROOT=3783.]]`

2. `[[CAT=ADJ][TYPE=ORDINAL][ROOT=3783.][CONV=NOUN=NONE]`
   `[AGR=3SG][POSS=NONE][CASE=NOM]]`

3785.si

1. `[[CAT=ADJ][TYPE=ORDINAL][ROOT=3785.][CONV=NOUN=NONE]`
   `[AGR=3SG][POSS=3SG][CASE=NOM]]`

– **Reals**: e.g.,

0,23'ten

1. `[[CAT=ADJ][TYPE=REAL][ROOT=0,23'][CONV=NOUN=NONE]`
   `[AGR=3SG][POSS=NONE][CASE=ABLy]]`

– **Percentages**: e.g.,

%7'sinin

1. `[[CAT=ADJ][TYPE=PERCENTAGE][ROOT=7'][CONV=NOUN=NONE]`
   `[AGR=3SG][POSS=3SG][CASE=GEN]]`

– **Time**: e.g.,

23:15'te

1. `[[CAT=ADJ][TYPE=TIME][ROOT=23:15'][CONV=NOUN=NONE]`
   `[AGR=3SG][POSS=NONE][CASE=LOCy]]`

– **Ratios**: e.g.,

3:5'i (üç bölü beşi)

(of three over five)

1. [[CAT=ADJ][TYPE=RATIO][ROOT=3:5'][CONV=NOUN=NONE]
   [AGR=3SG][POSS=NONE][CASE=ACCy]]

2. [[CAT=NUMBER][TYPE=RATIO][ROOT=3:5'][CONV=NOUN=NONE]
   [AGR=3SG][POSS=3SG][CASE=NOM]]

3/5'i (üç bölü beşi)

(of three over five)

1. [[CAT=ADJ][TYPE=RATIO][ROOT=3/5'][CONV=NOUN=NONE]
   [AGR=3SG][POSS=NONE][CASE=ACCy]]

2. [[CAT=NUMBER][TYPE=RATIO][ROOT=3/5'][CONV=NOUN=NONE]
   [AGR=3SG][POSS=3SG][CASE=NOM]]

3/5'ü (beşte üçü)

(of three of five)

1. [[CAT=ADJ][TYPE=RATIO][ROOT=3/5'][CONV=NOUN=NONE]
   [AGR=3SG][POSS=NONE][CASE=ACCy]]

2. [[CAT=NUMBER][TYPE=RATIO][ROOT=3:5'][CONV=NOUN=NONE]
   [AGR=3SG][POSS=3SG][CASE=NOM]]

Note that, some tokens have more than one parses. For example *7'nin* can be used in the text as *senin 7'nin* (your 7) or *7'nin 2'si* (2 of 7). It is no different than the word *evin* , which can be used as *senin evin* (your house) or *evin solu* (left of the house). Also, there is a problem in the reading of ratios. Consider the token *2/4'ü* (2 over 4), that can be read in Turkish as *iki bölü dördü*. It is also possible to read this token as *2/4'si*, that is *dörtte ikisi*, (two of four), and suffixation process according to the selected pronounciation.

- **Punctuation** : All punctuation marks are behaved as a distinct token. The apostrophe (') is an exception, which separates the suffixes of proper nouns.
- **Abbreviations** : These are classified into 3 classes:
  1. One capital letter followed by one or more small letters and a period. e.g. Dr. Prof.

2. Capital letters, each followed by a period.
   e.g. T.B.M.M.
3. One or two small letters followed by a period.
   e.g. cm. m.

- **Double quote** : Double quote cannot be in a part of the token, it is a punctuation marker. So, the occurrences like *"ev"in* are syntactically wrong, and must be corrected as *"evin"*.

Our tokenizer is implemented using *lex* which reads raw text and sends its output to the morphological analyzer.

### 3.1.2 Morphological Analyzer

We used the morphological analyzer developed by Oflazer [23], using the two-level transducer development software developed by Lauri Karttunen of Xerox PARC and of Rank Xerox Research Centre at Grenoble [16]. The morphological analyzer has about 30,000 root words and about 35,000 proper names. It can analyze about 2,000 words in one second in an Ultra-Sparc. Its output is a linear feature value pair sequence for morphemes.

Mainly, it gives the legitimate parses of the words. On the average, Turkish tokens have 2 such parses, because of the reasons described in the previous chapter.

### 3.1.3 Lexical and Non-lexical Collocation Recognizer

This module behaves like a multi-word construct processor. Turkish, like other languages, has many lexical and non-lexical collocations. Such a processor is needed, because certain words, which appear together, may behave very different, as a group. An example is the non-lexicalized collocation *koşa koşa* (running). The morphological analyzer considers the word *koşa* (let him run) as an optative verb, but if it is repeated, then, these two words together, have the grammatical role of a manner adverb in the sentence. The output of the morphological analyzer is as follows:

```
koSa

[[CAT=VERB][ROOT=koS][SENSE=POS][TAM1=OPT][AGR=3SG]]
(let him run)

koSa

[[CAT=VERB][ROOT=koS][SENSE=POS][TAM1=OPT][AGR=3SG]]
(let him run)
```

The collocation recognizer composes these two words into a very simple form:

```
koSa koSa

['koSa koSa',[[[cat,adverb],[root,'koSa koSa']]]]
(running)
```

There are a number of such non-lexicalized forms, which are specified in the collocation database and have already been defined by Oflazer and Kuruöz [24, 19]. Almost all of these collocations involve duplications, and have forms like $w + x \;\; w + y$ where $w$ is the duplicated string comprising the root and certain sequence of suffixes and $x$ and $y$ are possibly different (or empty) sequences of other suffixes. The following is a list of non-lexicalized collocations for Turkish that we handle in our preprocessor:

- duplicated optative and 3SG verbal forms functioning as manner adverb. This is the one described above.
- aorist verbal forms with root duplications and sense negation, functioning as temporal adverbs. For instance for the non-lexicalized collocation *yapar yapmaz*, where items have the parses

```
[[CAT=VERB] [ROOT=yap] [SENSE=POS]
     [TAM1=AORIST] [AGR=3SG]]
 (does)

[[CAT=VERB] [ROOT=yap] [SENSE=NEG]
     [TAM1=AORIST] [AGR=3SG]]
 (does not do)
```

respectively, the preprocessor generates the feature sequence

```
[[CAT=ADVERB] [ROOT=yapar yapmaz]]
 (as soon as s/he does)
```

- duplicated verbal and derived adverbial forms with the same verbal root acting as temporal adverbs, e.g.,

  gitti gideli
  `went go+since`
  *'since he went'*

- emphatic adjectival forms involving duplication and question clitic, e.g.,

  güzel mi güzel
  `beautiful question-clitic beautiful`
  *'very beautiful'*

- adjective or noun duplications that act as manner adverbs, e.g.,

  hızlı hızlı
  ADJ(`fast`) ADJ(`fast`)
  AVDERB(*'fast'*)

  ev ev
  `house house`
  *'house by house'*

- duplicated nominal forms, with word *be* between them, behaving as manner adverb, e.g.,

  ev be ev
  `house be house`
  *'house by house'*

- duplicated verbal forms, first one has the suffix *yip*, and second has negative sense, with necessitative modality, e.g.,

  gelip gelmemesi

  `come+`*yip* `come+NEG+INF+POSS-3SG`

  *'whether comes or do not come'*

- idiomatic forms, which are never used singularly, like *gürül gürül*,

This module also recognizes lexicalized collocations. A typical example is the group *yanı sıra* (besides). If these words are considered alone, we get the following parses:

```
yanI

[[CAT=ADJ][ROOT=yan][CONV=NOUN=NONE][AGR=3SG][POSS=NONE][CASE=ACCy]]
 (side+ACC)

[[CAT=ADJ][ROOT=yan][CONV=NOUN=NONE][AGR=3SG][POSS=3SG][CASE=NOM]]
 (side+GEN)

sIra

[[CAT=NOUN][ROOT=sIra][AGR=3SG][POSS=NONE][CASE=NOM]]
 (desk)

[[CAT=NOUN][ROOT=sIr][AGR=3SG][POSS=NONE][CASE=DATy]]
 (secret+DAT)
```

The collocation recognizer output for these words is as follows:

```
yanI sIra

['yanI sIra',[[[cat,postp],[root,'yanI sIra'],[subcat,gen]]]]
(besides)
```

Such lexical collocations are also defined in the collocation database. They are not allowed to take suffixes. Certain lexical collocations, like proper name groups, compound verbs, or idiomatic groups can take suffixes. So such lexical collocations are defined separately. Consider the collocation, *Mustafa Kemal Ataürk'ün*:

```
Mustafa

[[CAT=NOUN][ROOT=mustafa'][TYPE=RPROPER][AGR=3SG][POSS=NONE][CASE=NOM]]

Kemal

[[CAT=NOUN][ROOT=kemVl'][TYPE=RPROPER][AGR=3SG][POSS=NONE][CASE=NOM]]

[[CAT=NOUN][ROOT=kemVl][AGR=3SG][POSS=NONE][CASE=NOM]]

AtatUrk'Un

[[CAT=NOUN][ROOT=atatUrk'][TYPE=RPROPER][AGR=3SG][POSS=NONE][CASE=GEN]]

[[CAT=NOUN][ROOT=atatUrk'][TYPE=RPROPER][AGR=3SG][POSS=2SG][CASE=NOM]]
```

and as a whole, this word group has the same analyze as the last word, *Atatürk'ün*.

```
['mustafa kemal atatUrk''Un',
[
[[cat,noun],[root,'mustafa kemal atatUrk'''],[type,rproper],
 [agr,'3SG'],[poss,'NONE'],[case,gen]],

[[cat,noun],[root,'mustafa kemal atatUrk'''],[type,rproper],
 [agr,'3SG'],[poss,'2SG'],[case,nom]]
]]
```

It is worthwhile noting that for the lexical collocations, the program tries to find the longest group. For example, although both *Mustafa Kemal* and *Mustafa*

*Kemal Atatürk* are in the database, regardless of their order, if the longer one exists in the text, the program chooses that one. In a previous study, Kuruöz requires that shorter one must exist before the longer one, in the database [19].

Consider the lexical collocation of *devam et* (continue), if we can group these words in one item, the morphological disambiguator's work will be much easier in such a sentence:

| | | | | |
|---|---|---|---|---|
| benim | devam etmem | artık | | imkansızlaşmıştı. |
| `my` | `continuation` | `any more` | | `became impossible` |

*'it was impossible for me to continue any more.'*

Morphological disambiguator will not deal with a wrong parse, *benim devam* (my remedy), since *devam etmem* is already grouped as a unit.

Recognition of the dates and percentages are handled also in this module. In the recognition of the dates, the maximum length of the date expression is tried to packeted like:

```
2 Subat 1915'te

[[cat:date,root:'2 Subat 1915''',type:temp1,
  agr:'3SG',poss:'NONE',case:loc]]

(on 2 February 1915)
```

The collocations in the form, the token *yüzde* (percent) followed by a numeric are also recognized here, e.g.,

```
yUzde 12

[[cat:adj,root:'yUzde 12',type:percentage]]

(12 percent)
```

Figure 3.2 summarizes the form of the collocation database. Also, this collocation database can be seen in Appendix B

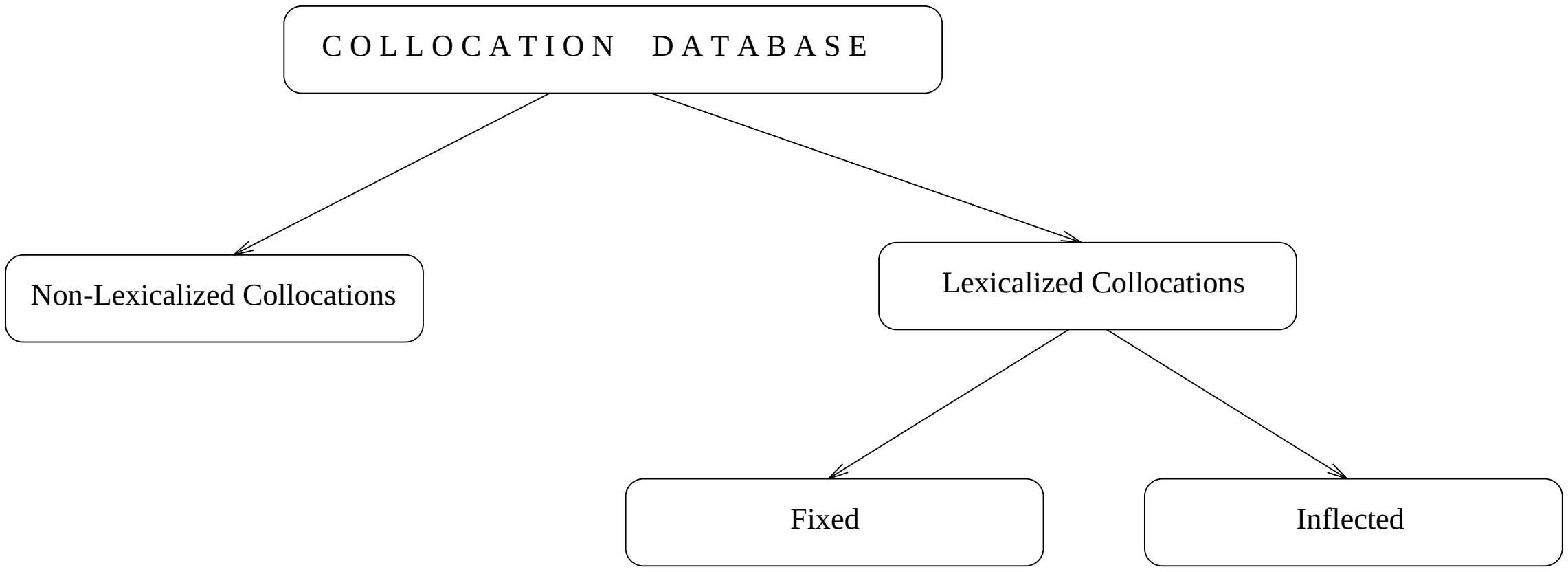

Figure 3.2: The Collocation Database

This module also handles certain tokens, like punctuation marks and roman numbers, which can not be handled in the morphological analyzer. The following are some examples to such tokens:

```
!

[[cat:punct,root:'!']]

xv.

[[cat:adj,root:'xv.',type:ordinal]]
```

While performing these functions, this module also gathers the following statistics from the raw text:

- Number of tokens,
- Number of parses,
- Parse distribution, and

- Total calls to morphological analyzer.

These statistics are used mainly, when evaluating the experimental results, presented in the next chapter.

The output of this module for each sentence is in the form:

```
[

[token 1,[[parse 11],[parse 12],...,[parse 1i]]],

[token 2,[[parse 21],[parse 22],...,[parse 2j]]],

...

[token n,[[parse n1],[parse n2],...,[parse nk]]]

].
```

For instance, for the sentence

benim devam etmem artık imkansızlaşmıştı.
`my continuation any more became impossible`
*'it was impossible for me to continue any more.'*

the output is as follows:

```
['benim',

[[[cat,noun],[root,'ben'],[agr,'3SG'],[poss,'NONE'],[case,nom],
  [conv,verb,none],[tam1,pres],[agr,'1SG']],

 [[cat,noun],[root,'ben'],[agr,'3SG'],[poss,'1SG'],[case,nom]],

 [[cat,pronoun],[root,'ben'],[type,personal],[agr,'1SG'],[poss,'NONE'],[case,gen]],
```

```
 [[cat,pronoun],[root,'ben'],[type,personal],[agr,'1SG'],[poss,'NONE'],
  [case,nom],[conv,verb,none],[tam1,pres],[agr,'1SG']]]],


['devam etmem',

[[[cat,verb],[root,'devam ed'],[sense,neg],[tam1,aorist],[agr,'1SG']],

 [[cat,verb],[root,'devam ed'],[sense,pos],[conv,noun,ma],
  [type,infinitive],[agr,'3SG'],[poss,'1SG'],[case,nom]]]],


['artIk',

[[[cat,adj],[root,'artIk']],

 [[cat,adverb],[root,'artIk']]]],


['imkansIzlaSmIStI',

[[[cat,noun],[root,'imkan'],[conv,adj,siz],[conv,verb,las],
  [sense,pos],[tam1,narr],[tam2,past],[agr,'3SG']]]],


['.',

[[[cat,punct],[root,'.']]]]
```

### 3.1.4 Unknown Word Processor

Although the coverage of our morphological analyzer for Turkish [23], with about 30,000 root words and about 35,000 proper names, is very satisfactory, it is inevitable that there will be forms in the corpora being processed that are not recognized by the morphological analyzer. These are almost always foreign proper names, words adapted into the language and not in the lexicon, or very obscure technical words. These are nevertheless inflected (using Turkish word formation paradigms) with inflectional features demanded by the syntactic context and sometimes even go through derivational processes. For improved disambiguation, one has to at least recover any morphological features even if the root word is unknown. To deal with this, we have made the assumption that all unknown words have nominal roots, and built a second morphological analyzer whose (nominal) root lexicon recognizes $S^+$ where $S$ is the Turkish surface alphabet (in the two-level morphology sense), but then tries to interpret an arbitrary postfix of the unknown word as a sequence of Turkish suffixes subject to all morphographemic constraints. For instance when a form such as *talkshowumun* is entered, this second analyzer hypothesizes the following analyses:

```
1. [[CAT NOUN] [ROOT talkshowumun]
     [AGR 3SG] [POSS NONE] [CASE NOM]]

2. [[CAT NOUN] [ROOT talkshowumu]
     [AGR 3SG] [POSS 2SG] [CASE NOM]]

3. [[CAT NOUN] [ROOT talkshowum]
     [AGR 3SG] [POSS NONE] [CASE GEN]]

4. [[CAT NOUN] [ROOT talkshowum]
     [AGR 3SG] [POSS 2SG] [CASE NOM]]

5. [[CAT NOUN] [ROOT talkshowu]
     [AGR 3SG] [POSS 1SG] [CASE GEN]]

6. [[CAT NOUN] [ROOT talkshow]
     [AGR 3SG] [POSS 1SG] [CASE GEN]]
```

which are then processed just like any other during disambiguation.[1]

[1]Incidentally, the correct analysis is the $6^{th}$ meaning *of my talk show*. The $5^{th}$ one has the same morphological features except for the root.

Another example is the token *kermezdere'deki* (at Kermezdere). Kermezdere is a proper noun, and not known by the morphological analyzer. But it is possible to guess, using the apostrophe and the suffixes *de* and *ki*. As a result unknown word recognizer suggests the following parses for this token:

```
1. [[cat,noun],[root,'kermezdere'],[agr,'3SG'],[poss,'NONE'],
    [type,proper],[case,locy],[conv,adj,rel]]

2. [[cat,noun],[root,'kermezdere'],[agr,'3SG'],[poss,'NONE'],
    [type,proper],[case,locy],[conv,adj,rel],
    [conv,noun,none],[agr,'3SG'],[poss,'NONE'],[case,nom]]
```

This however is not a sufficient solution for some very obscure situations where for the foreign word is written using its, say, English orthography, while suffixation goes on according to its English pronunciation, which may make some constraints like vowel harmony inapplicable on the graphemic representation, though harmony is in effect in the pronunciation. For instance one sees the form $Carter'a$ where the last vowel in $Carter$ is pronounced so that it harmonizes with $a$ in Turkish, while the $e$ in the surface form does not harmonize with $a$.

We are nevertheless rather satisfied with our solution as in our experiments we have noted that *well below* 1% of the forms remain as unknown and these are usually item markers in formatted or itemized lists, or obscure foreign acronyms. Our experimental results also indicate that more than 90% of the processed words, which are not recognized by the morphological analyzer have got their intended readings after the morphological disambiguation.

### 3.1.5 Format Conversion

The preprocessor then converts each parse into a hierarchical feature structure so that the inflectional feature of the form with the last category conversion (if any) is at the top level. Thus in the example above, *gelişindeki*, whose morphological analysis is:

```
[[CAT VERB] [ROOT gel] [SENSE POS]
 [CONV NOUN YIS] [AGR 3SG]
 [POSS 2SG] [CASE LOC]
 [CONV ADJ REL]]
```

the following feature structure is generated.

$$\begin{bmatrix} \text{CAT} & \text{ADJ} \\ \text{STEM} & \begin{bmatrix} \text{CAT} & \text{NOUN} \\ \text{AGR} & \text{3SG} \\ \text{POSS} & \text{2SG} \\ \text{CASE} & \text{LOC} \\ \text{STEM} & \begin{bmatrix} \text{CAT} & \text{VERB} \\ \text{ROOT} & \text{gel} \\ \text{SENSE} & \text{POS} \end{bmatrix} \\ \text{SUFFIX} & \text{YIS} \end{bmatrix} \\ \text{SUFFIX} & \text{REL} \end{bmatrix}$$

Consider again, the token *imkansızlaşmıştı* (it had become impossible) in the example sentence above. The feature structure for this token is as follows:

$$\begin{bmatrix} \text{CAT} & \text{VERB} \\ \text{SENSE} & \text{POS} \\ \text{TAM1} & \text{NARR} \\ \text{TAM2} & \text{PAST} \\ \text{AGR} & \text{3SG} \\ \text{STEM} & \begin{bmatrix} \text{STEM} & \begin{bmatrix} \text{CAT} & \text{NOUN} \\ \text{ROOT} & \text{imkan} \end{bmatrix} \\ \text{SUFFIX} & \text{SIZ} \end{bmatrix} \\ \text{SUFFIX} & \text{LAS} \end{bmatrix}$$

The root of this word is the noun *imkan* (possibility). The suffix *sız* makes an adjective, converting it into the word *impossible.* Then the next suffix *laş* makes a verb, *became impossible.* The rest of the word defines the tense, aspect and agreement of the verbal form. The output of the format conversion module for the example sentence is as follows:

```
[benim,

[[cat:verb,stem:[cat:noun,root:ben,agr:'3SG',poss:'NONE',case:nom],
  suffix:none,tam2:pres,agr:'1SG'],

 [cat:noun,root:ben,agr:'3SG',poss:'1SG',case:nom],

 [cat:pronoun,root:ben,type:personal,agr:'1SG',poss:'NONE',case:gen],

 [cat:verb,stem:[cat:pronoun,root:ben,type:personal,agr:'1SG',poss:'NONE',
  case:nom],suffix:none,tam2:pres,agr:'1SG']]],


['devam etmem',

[[cat:verb,root:'devam ed',sense:neg,tam1:aorist,agr:'1SG'],

 [cat:noun,stem:[cat:verb,root:'devam ed',sense:pos],
  suffix:ma,type:infinitive,agr:'3SG',poss:'1SG',case:nom]]],


[artIk,

[[cat:adj,root:artIk],

 [cat:noun,stem:[cat:adj,root:artIk],suffix:none,agr:'3SG',poss:'NONE',case:nom],

 [cat:adverb,root:artIk]]],


[imkansIzlaSmIStI,

[[cat:verb,stem:[cat:adj,stem:[cat:noun,root:imkan],suffix:siz],
  suffix:las,sense:pos,tam1:narr,tam2:past,agr:'3SG']]],


['.',
```

```
[[cat:punct,root:'.']]]
```

### 3.1.6 Projection

In the learning modules, which will be explained in the sections 3.3 and 3.4, we used projected parses of the tokens instead of using all of the features of a parse. The motivation behind this is as follows: We need certain statistical information in order to proceed in learning, such as the number of occurences of a certain parse in a certain context unambiguously. Considering the whole parse would be meaningless, because of the large number of features, including the root. In such a case, although we have omitted the root, the performance of the learning would have been well below satisfactory. On the other hand, if we have considered only the category of the tokens, then there would have been no difference. So, we decided to *project* each such feature structure on a subset of its features. The features selected are

- inflectional and certain derivational markers, and stems for open class of words,
- roots and certain relevant features such as subcategorization requirements for closed classes of words such as connectives, postpositions, etc.

The set of features selected for each part-of-speech category is determined by a template and hence is controllable, permitting experimentation with differing levels of information. The information selected for stems are determined by the category of the stem itself recursively.

Under certain circumstances where a token has two or more parses that agree in the selected features, those parses will be represented by a single projected parse, hence the number of parses in the (projected) training corpus may be smaller than the number of parses in the original corpus. For example, the feature structure above is projected into a feature structure such as:

$$\begin{bmatrix} \text{CAT} & \text{ADJ} \\ \text{STEM} & \begin{bmatrix} \text{CAT} & \text{NOUN} \\ \text{AGR} & \text{3SG} \\ \text{POSS} & \text{1SG} \\ \text{CASE} & \text{LOC} \\ \text{STEM} & \begin{bmatrix} \text{CAT} & \text{VERB} \end{bmatrix} \\ \text{SUFFIX} & \text{DIK} \end{bmatrix} \\ \text{SUFFIX} & \text{REL} \end{bmatrix}$$

The sentence above

| benim | devam etmem | artık | imkansızlaşmıştı. |
|---|---|---|---|
| `my` | `continuation` | `any more` | `became impossible` |

*'it was impossible for me to continue any more.'*

is projected as follows:

```
[benim,

[[cat:verb,stem:[cat:noun,agr:'3SG',poss:'NONE',case:nom]],

 [cat:noun,agr:'3SG',poss:'1SG',case:nom],

 [cat:pronoun,agr:'1SG',poss:'NONE',case:gen],

 [cat:verb,stem:[cat:pronoun,agr:'1SG',poss:'NONE',case:nom]]]],


['devam etmem',

[[cat:verb],

 [cat:noun,agr:'3SG',poss:'1SG',case:nom]]],


[artIk,
```

```
[[cat:adj],

 [cat:noun,agr:'3SG',poss:'NONE',case:nom],

 [cat:adverb]]],


[imkansIzlaSmIStI,

[[cat:verb,stem:[cat:adj]]]],


['.',

[[cat:punct,root:'.']]]
```

## 3.2 Constraint Rules

The system uses rules of the sort

if `LC` and `RC` then `choose PARSE`

or

if `LC` and `RC` then `delete PARSE`

where `LC` and `RC` are feature constraints on unambiguous left and right contexts of a given token, and `PARSE` is a feature constraint on the parse(s) that is (are) chosen (or deleted) in that context if they are subsumed by that constraint. Currently, the left and right contexts can be at most 2 tokens, hence we look at a window of at most 5 tokens of which one is ambiguous. We refer to the unambiguous tokens in the context as `llc` (left-left context) `lc` (left context), `rc` (right context) and `rrc` (right-right context). Depending on the amount of unambiguous tokens in a context, our rules can have one of the following context structures, listed in order of decreasing specificity:

```
1.     llc, lc  ____  rc, rrc

2.     llc, lc  ____
                ____  rc, rrc

3.          lc  ____  rc

4.          lc  ____
                ____  rc
```

To illustrate the flavor of our rules we can give the following examples. The first example chooses parses with case feature ablative, preceding an unambiguous postposition which subcategorizes for an ablative nominal form.

```
[llc:[],lc:[],
    choose:[case:abl],
        rc:[[cat:postp,subcat:abl]],rrc:[]]
```

| | | |
|---|---|---|
| | candan | önce |
| * | ADVERB(friendly) | POSTP(before) |
| | NOUN(soul)+ABL | |

A second example rule is

```
  [llc:[[cat:adj,type:determiner]],
    lc:[[cat:adj,stem:[cat:noun]]],
choose:[cat:adj],
    rc:[[cat:noun,poss:'NONE']], rrc:[]].
```

| | | | | |
|---|---|---|---|---|
| | bir | odadaki | kırmızı | top |
| | DET(a) | NOUN(oda)+LOC+ADJ(ki) | red | ball |
| * | ADVERB(only if) | | | |

which selects and adjective parse following a determiner, adjective sequence, and before a noun without a possessive marker.

Another example rule is:

```
[llc:[],lc:[[agr:'2SG',case:gen]],
    choose:[cat:noun,poss:'2SG'],
        rc:[],rrc:[]]
```

| | | |
|---|---|---|
| | senin | gelişinin |
| | your | coming+POSS-2SG |
| * | your | coming+POSS-3SG |

which chooses a nominal form with a possessive marker `2SG` following a pronoun with `2SG` agreement and genitive case, enforcing the simplest form of noun–noun form noun phrase constraints.

Our system uses two hand-crafted sets of rules:

1. We use an initial set of hand-crafted *choose rules* to speed-up the learning process by creating disambiguated contexts over which statistics can be collected. These rules (examples of which are given above) are independent of the corpus that is to be tagged, and are linguistically motivated. They enforce some very common feature patterns especially where word order is rather strict as in NP's or PP's.[2] The motivation behind these rules is that they should improve precision without sacrificing recall. *These are rules which impose very tight constraints so as not to make any recall errors.* Our experience is that after processing with these rules, the recall is above 99% while precision improves by about 20 percentage points. *Another important feature of these rules is that they are applied even if the contexts are also ambiguous*, as the constraints are tight. That is, if each token in a sequence of, say, three ambiguous tokens have a parse matching one of the context constraints (in the proper order), then all of them are simultaneously disambiguated. In hand crafting these rules, we have used our experience from an earlier tagger [24]. Currently we use 288 hand-crafted choose rules.

2. We also use a set of hand-crafted heuristic *delete rules* to get rid of any very low probability parses. For instance, in Turkish, postpositions have rather strict contextual constraints and if there are tokens remaining with multiple parses one of which is a postposition reading, we delete that reading. Our experience is that these rules improve precision by about 10 to 12 additional percentage points with negligible impact on recall. Currently we use 43 hand-crafted delete rules.

[2]Turkish is a free constituent order language whose unmarked order is SOV.

Appendix D presents the hand-crafted choose and delete rules.

## 3.3 Learning Choose Rules

Given a training corpus, with tokens annotated with possible parses (projected over selected features), we first apply the hand-crafted rules. Learning then goes on as a number of iterations over the training corpus. We proceed with the following schema which is an adaptation of Brill's formulation [6]:

1. We generate a table, called *incontext*, of all possible unambiguous contexts which contain a token with an unambiguous (projected) parse, along with a count of how many times this parse occurs unambiguously in exactly the same context in the corpus. We refer to an entry in table with a context $C$ and parse $P$ as $incontext(C, P)$.

2. We also generate a table, called *count*, of all unambiguous parses in the corpus along with a count of how many times this parse occurs in the corpus. We refer to an entry in this table with a given parse $P$, as $count(P)$.

3. We then start going over the corpus token by token generating contexts as we go.

4. For each unambiguous context encountered, $C = (\texttt{LC}, \texttt{RC})$[3] around an *ambiguous* token $w$ with parses $P_1, \ldots P_k$, and for each parse $P_i$, we generate a candidate rule of the sort

   if `LC` and `RC` then `choose` $P_i$

5. Every such candidate rule is then scored in the following fashion:

[3] Either of `LC` or `RC` may be empty.

(a) We compute

$$P_{max} = argmax_{P_j\ (j \neq i)} \frac{count(P_i)}{count(P_j)} \cdot incontext(C, P_j).$$

(b) The score of the candidate rule is then computed as:

$$Score_i = incontext(C, P_i) - \frac{count(P_i)}{count(P_{max})} \cdot incontext(C, P_{max})$$

6. We order all candidate rules generated during one pass over the corpus, along two dimensions:

   (a) we group candidate rules by *context specificity* (given by the order in Section 3.2),

   (b) in each group, we order rules by descending *score.*

   We maintain score thresholds associated with each context specificity group: the threshold of a less specific group being higher than that of a more specific group. We then choose the top scoring rule from any group whose score equals or exceeds the threshold associated with that group. The reasoning is that we prefer more specific and/or high scoring rules: high scoring rules are applicable, in general, in more places; while more specific rules have more strict constraints and more accurate morphological parse selections, We have noted that choosing the highest scoring rule at every step may sometimes make premature commitments which can not be undone later.

7. The selected rules are then applied in the matching contexts and ambiguity in those contexts is reduced. During this application the following are also performed:

   (a) If the application results in an unambiguous parse in the context of the applied rule, we increment the count associated with this parse in table *count*. We also update the *incontext* table for the same context, and other contexts which contains the disambiguated parse.

   (b) We also generate any new unambiguous contexts that this newly disambiguated token may give rise to, and add it to the *incontext* table along with count 1.

Note that for efficiency reasons, rule candidates are not generated repeatedly during each pass over the corpus, but rather once at the beginning, and then when selected rules are applied to very specific portions of the corpus.

8. If there are no rules in any group that exceed its threshold, group thresholds are reduced by multiplying by a damping constant $d$ $(0 < d < 1)$ and iterations are continued.

9. If the threshold for the most specific context falls below a given lower limit, the learning process is terminated.

Some of the rules that have been generated by this learning process are given below:

1. Disambiguate around a coordinating conjunction:

```
[llc:[],lc:[],
    choose:[cat:noun,agr:3SG,case:nom],
        rc:[[cat:conn,root:ve]],
       rrc:[[cat:noun,agr:3SG,poss:NONE]]]
```

| | | | |
|---|---|---|---|
| | kazan | ve | tencerede |
| | NOUN(cauldron) | CONN(and) | NOUN(pot)+LOC |
| * | ADJ(digging) | | |

2. Choose participle form adjectival over a nominal reading:

```
[llc:[],lc:[],
    choose:[cat:adj,suffix:yan],
        rc:[[cat:noun,agr:3SG,poss:NONE]],
       rrc:[[cat:noun,agr:3SG,poss:3SG]]].
```

| | | | |
|---|---|---|---|
| | kazan | yol | işçisi |
| | ADJ(digging) | NOUN(road) | NOUN(worker)+POSS-3SG |
| * | NOUN(cauldron) | | |

3. Choose a nominal reading (over an adjectival) if a three token compound noun agreement can be established with the next two tokens:

```
[llc:[],lc:[],
    choose:[cat:noun,agr:3SG,case:nom],
        rc:[[cat:noun,agr:3SG,poss:3SG]],
       rrc:[[cat:noun,agr:3SG,poss:3SG]]]
```

| | | | |
|---|---|---|---|
| kitap | | kapağı | | resmi |
| NOUN(book) | | NOUN(cover)+POSS-3SG | | NOUN(picture)+POSS-3SG |
| | * | NOUN(cover)+ACC | * | NOUN(picture)+ACC |

### 3.3.1 Contexts induced by morphological derivation

The procedure outlined in the previous section has to be modified slightly in the case when the unambiguous token in the `rc` position is a morphologically derived form. For such cases one has to take into consideration additional pieces of information. We will motivate this using a simple example from Turkish. Consider the example fragment:

... bir masa+dır.
... *a* *table+is*
... *is* *a table*

where the first token has the morphological parses:

```
1. [[CAT ADJ] [ROOT bir] [TYPE CARDINAL]]
   (one)

2. [[CAT ADJ] [ROOT bir] [TYPE DETERMINER]]
   (a)

3. [[CAT ADVERB] [ROOT bir]]
   (only/merely)
```

and the second form has the unambiguous morphological parse:

```
1. [[CAT NOUN] [ROOT masa] [AGR 3SG] [POSS NONE]
      [CASE NOM] [CONV VERB NONE]
      [TAM1 PRES] [AGR 3SG]]  (is table)
```

which in hierarchical form corresponds to the feature structure:

$$\begin{bmatrix} \text{CAT} & \text{VERB} \\ \text{TAM1} & \text{PRES} \\ \text{AGR} & \text{3SG} \\ \text{STEM} & \begin{bmatrix} \text{CAT} & \text{NOUN} \\ \text{ROOT} & \text{masa} \\ \text{AGR} & \text{3SG} \\ \text{POSS} & \text{NONE} \\ \text{CASE} & \text{NOM} \end{bmatrix} \\ \text{SUFFIX} & \text{NONE} \end{bmatrix}$$

In the syntactic context this fragment is interpreted as

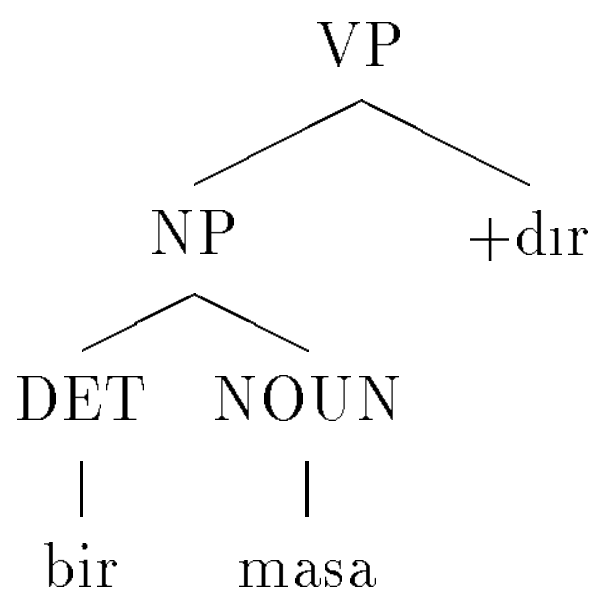

where the determiner is attached to the noun and the whole phrase is then taken as a VP although the verbal marker is on the second lexical item. If, in this case, the token *bir* is considered to neighbor a token whose top level inflectional features indicate it is a verb, it is likely that *bir* will be chosen as an adverb as it precedes a verb, whereas the correct parse is the determiner reading.

In such a case where the right context of an ambiguous token is a derived form, one has to consider as the right context, both the top level features of final form, and the *stem* from which it was derived. During the set-up of the *incontext* table, such a context is entered twice: once with the top level feature constraints of the immediate unambiguous right-context, and once with the feature constraints of the stem. The unambiguous token in the right context is also entered to the *count* table once with its top level feature structure and once with the feature structure of the stem.

When generating candidate choose or delete rules, for contexts where `rc` is a derived form and `rrc` is empty, we actually generate two candidates rules for each ambiguous token in that context:

1. if `llc, lc` and `rc` then `choose/delete` $P_i$.
2. if `llc, lc` and *stem*(`rc`) then `choose/delete` $P_i$.

These candidate rules are then evaluated as described above. In general all derivations in a lexical form have to be considered though we have noted that considering one level gives satisfactory results.

### 3.3.2 Ignoring Features

Some morphological features are only meaningful or relevant for disambiguation only when they appear to the left or to the right of the token to be disambiguated. For instance, in the case of Turkish, the `CASE` feature of a nominal form is only useful in the immediate left context, while the `POSS` (the possessive agreement marker) is useful only in the right context. If these features along with their possible values are included in context positions where they are not relevant, they "split" scores and hence cause the selection of some other irrelevant rule. Using the maxim that union gives strength, we create contexts so that features not relevant to a context position are not included, thereby treating context that differ in these features as same.[4]

[4]Obviously these features are specific to a language.

## 3.4 Learning Delete Rules

For choosing delete rules we have experimented with two approaches. One obvious approach is to use the formulation described above for learning choose rules, but instead of generating choose rules, pick the parses that score (significantly) worse than and generate delete rules for such parses. We have implemented this approach and found that it is not very desirable due to two reasons:

1. it generates far too many delete rules, and
2. it impacts recall seriously without a corresponding increase in precision.

The second approach that we have used is considerably simpler. We first re-process the training corpus but this time use a second set of projection templates, and apply initial rules, learned choose rules and heuristic delete rules. Then for every unambiguous context $C = (\texttt{LC},\texttt{RC})$, with either an immediate left, or an immediate right components or both (so the contexts used here are the last 3 in Section 3.2), a score

$$\frac{incontext(C, P_i)}{count(P_i)}$$

for each parse $P_i$ of the (still) ambiguous token, is computed. Then, delete rules of the sort

if `LC` and `RC` then `delete` $P_i$

are generated for all parses with a score below a certain fraction (0.2 in our experiments) of the highest scoring parse. In this process, our main goal is to remove any seriously improbable parses which may somehow survive all the previous choose and delete constraints applied so far. Using a second set of templates which are more specific than the templates used during the learning of the choose rules, we introduce features we were originally projected out. Our experience has been that less strict contexts (e.g., just a `lc` or `rc`) generate very useful delete rules, which basically weed out what can (almost) never happen as it is certainly not very feasible to formulate hand-crafted rules that specify what sequences of features are not possible.

Some of the interesting delete rules learned here are:

1. Delete the first of two consecutive verb parses:

```
[llc:[],lc:[],
    delete:[cat:verb],
        rc:[[cat:verb]],rrc:[]]
```

| | | |
|---|---|---|
| | yıkanmış | elbisedir |
| | ADJ(washed) | is..dress |
| * | VERB(is washed) | |

2. Delete accusative case marked noun parse before a postposition that subcategorizes for a nominative noun:

```
[llc:[],lc:[],
 delete:[cat:noun,agr:3SG,poss:NONE,case:acc],
     rc:[[cat:postp,subcat:nom]],rrc:[]].
```

| | | |
|---|---|---|
| | kazı | gibi |
| | NOUN(excavation) | POSTP(like) |
| * | NOUN(goose)+ACC | |

3. Delete the accusative case marked parse without any possessive marking, if the previous form has genitive case marking (signaling a genitive–possessive NP construction):

```
[llc:[],
 lc:[[cat:noun,agr:3SG,poss:NONE,case:gen]],
delete:[cat:noun,agr:3SG,poss:NONE,case:acc],
    rc:[],rrc:[]].
```

| | | |
|---|---|---|
| çocuğun | | kitabı |
| NOUN(child)+GEN | | NOUN(book)+POSS-3SG |
| | * | NOUN(book)+ACC |

## 3.5 Using Context Statistics to Delete Parses

After applying hand-crafted rules to a text to be disambiguated we arrive at a state where ambiguity is about 1.10 to 1.15 parses per token (down from 1.70 to 1.80 parses per token) without any serious loss on recall. This state allows statistics to be collected over unambiguous contexts. To remove additional parses which never appear in any unambiguous context we use the scoring described above for choosing delete rules, to discard parses on the current text based on context statistics.[5] We make three passes over the current text, scoring parses in unambiguous contexts of the form used in generating delete rules, and discarding parses whose score is below a certain fraction of the maximum scoring parse, on the fly. The only difference with the scoring used for delete rules, is that the score of a parse $P_i$ here is a weighted sum of the quantity

$$\frac{incontext(C, P_i)}{count(P_i)}$$

evaluated for three contexts in the case both the `lc` and `rc` are unambiguous.

## 3.6 Using Root Word Statistics

Using root word statistics is a well known technique, and mainly used in tagging English texts. The idea beyond this technique is the following: Although most of the words are ambiguous in the dictionary, for the real texts, generally, the usage frequencies of these readings are very different. In the sentence:

I see a bird

the word *see* is ambiguous according to the Webster's dictionary, it can be used as the Holy See. But this reading is too rare that, it is possible to discard it.

Similarly, for Turkish, it is possible to detect such ambiguities left in the sentence, and using a statistics database. A typical example for Turkish is the word *ama* (but). In a 2,000 sentence text, this word is used 270 times as a connective, and never used in its old meaning, *blind.* So, it is possible to do this operation automatically. Hand-crafted rules contain some similar lexical issues, like *ama* (but/blind), *de* (too/say) or *diye* (for that reason/let him say).

[5] Please note that delete rules learned may be applied to future texts to be disambiguated, while this step is applied to the current text on which disambiguation is performed.

The functionality of this module is to resolve some ambiguities left using such statistics. In fact, after the contextual statistics, the ambiguity is about 1.04 parses per tokens, and the parses of the ambiguities left are generally differ in the roots and some features like type or voice. There is also another type of tokens, which are still ambiguous: Most of the time, a lexicalized word, which is considered as a separate parse is a derived reading of another parse. A typical example for this is the word *deyiş* (utterance). It can also be derived from the verb *de* (to say), with a suffix *yiş*, forming deyiş (saying). In this case, the higher level feature structures are generally the same, but one parse is derived, the other is lexicalized. The root statistics have tendency to select the derived parse. This is the major drawback of this method.

Some example tokens disambiguated using this module are:

```
[[cat:noun,root:gOze,agr:3SG,poss:NONE,case:nom],  (cell)
 [cat:noun,root:gOz,agr:3SG,poss:NONE,case:dat]]   (eye)
```

The root *göz* (eye) is much more frequent than *göze* (cell), so *göze* may be deleted.

```
[[cat:noun,root:alan,agr:3SG,poss:NONE,case:gen], (area)
 [cat:noun,root:ala,agr:3SG,poss:NONE,case:gen]]  (the one with red)
```

In this example above, the root *alan* is much more frequent, so it is chosen instead of the other reading with the root *ala*.

This is a very practical and effective method, but unfortunately, it is very dangerous to apply this on the earlier steps. It must be applied to the tokens, which are still left ambiguous, when ambiguity is too low.

# Chapter 4

# Experimental Results

We have applied our learning system to four Turkish texts, two of them are previously unseen.[1] Table 4.1 gives some statistics on these texts. The first text labeled ARK is a short text on near eastern archaeology. The second text from which fragments whose labels start with C are derived, is a book on early $20^{th}$ history of Turkish Republic. The third text, MANUAL is a technical manual, originally translated from English. The last text, EMBASSY is a guideline document for the embassy staff.

Given a new text annotated with all morphological parses of the tokens, the disambiguation process goes through the following steps:

1. The initial hand-crafted choose rules are applied first. These rules always constrain top level inflectional features, and hence, any stems from derivational processes are not considered unless explicitly indicated in the constraint itself.
2. The hand-crafted delete clean-up rules are applied.
3. Context statistics described in the preceding section are used to discard further parses.

[1] During our experiments, we have used a tool developed by Nihat Özkan as a senior project in order to create a gold standard disambiguated text and compare an automatically disambiguated text with this gold standard. This tool has made the whole evaluation much faster and easier.

4. Root word preference statistics are used before applying the learned constraints.

5. The choose rules that have been learned earlier, are then repeatedly applied to unambiguous contexts, until no more ambiguity reduction is possible. During the application of these rules, if the immediate right context of a token is a derived form, then the stem of the right context is also checked against the constraint imposed by the rule. So if the rule right context constraint subsumes the top level feature structure or the stem feature structure, then the rule succeeds and is applied if all other constraints are also satisfied.

6. Finally, the delete rules that have been learned are applied repeatedly to unambiguous contexts, until no more ambiguity reduction is possible.

In Table 4.1, the tokens considered are that are generated after morphological analysis, unknown word processing and any lexical coalescing is done. The words that are unknown are those that could not even be processed by the unknown noun processor. Whenever an unknown word had more than one parse it was counted under the appropriate group.[2]

| Text | Sentences | Tokens | Distribution of Morphological Parses | | | | | |
|---|---|---|---|---|---|---|---|---|
| | | | 0 | 1 | 2 | 3 | 4 | $> 4$ |
| ARK | 492 | 7,928 | 0.15% | 49.34% | 30.93% | 9.19% | 8.46% | 1.93% |
| C2400 | 2,407 | 39,800 | 0.03% | 50.56% | 28.66% | 10.12% | 8.16% | 2.47% |
| C270 | 270 | 5212 | 0.02% | 50.63% | 30.68% | 8.62% | 8.36% | 1.69% |
| EMBASSY | 198 | 5177 | 0.09% | 43.94% | 34.58% | 9.60% | 9.46% | 2.33% |
| MANUAL | 204 | 2756 | 0.65% | 49.01% | 31.70% | 6.37% | 8.91% | 3.36% |

Table 4.1: Statistics on Texts

We learned rules from ARK itself, and on the first 500, 1000, and 2000 sentence portions of C2400. C270 which was from the remaining 400 sentences of C2400 was set aside for testing. Gold standard disambiguated versions for ARK, C270 were prepared manually to evaluate the automatically tagged versions.

[2]For the text MANUAL, 18 of the tokens are left as unknown, but the roots of 16 of them are the function keys, like *f10*. These tokens are said to be illegal by the morphological analyzer. This is why the unknown word ratio is too high.

Our results are summarized in the following set of tables. Tables 4.2 and 4.3 give the ambiguity, recall and precision initially, after hand-crafted rules are applied, and after the contextual statistics are used to remove parses – all applications being cumulative. The rows labeled BASE give the initial state of the text to be tagged. The rows labeled INITIAL CHOOSE give the state after hand-crafted choose rules are applied, while the rows labeled INITIAL DELETE give the state after the hand-crafted choose and delete rules are applied. The rows labeled CONTEXT STATISTICS give the state after the rules are applied and context statistics are used (as described earlier) to remove additional parses.

| **Disambiguation Stage** | **Ambiguity** | **Recall** (%) | **Pre.** (%) |
|---|---|---|---|
| BASE | 1.828 | 100.00 | 54.69 |
| INITIAL CHOOSE | 1.339 | 99.28 | 74.13 |
| INITIAL DELETE | 1.110 | 99.08 | 88.91 |
| CONTEXT STATISTICS | 1.032 | 97.38 | 94.35 |

Table 4.2: Average parses, recall and precision for text ARK

| **Disambiguation Stage** | **Ambiguity** | **Recall** (%) | **Pre.** (%) |
|---|---|---|---|
| BASE | 1.719 | 100.00 | 58.18 |
| INITIAL CHOOSE | 1.353 | 99.16 | 73.27 |
| INITIAL DELETE | 1.130 | 98.73 | 87.24 |
| CONTEXT STATISTICS | 1.038 | 96.70 | 93.15 |

Table 4.3: Average parses, recall and precision for text C270

Tables 4.4 and 4.5 present the results of further disambiguation of ARK, and C270 using rules learned from training texts C500, C1000, C2000 and ARK. These rules are applied after the last stage in the tables above.[3] The number of rules learned are given in Table 4.6.[4]

Our next information source is the root word statistics, obtained by disambiguating the full version of the text C2400. In Section 3.6 the algorithm that uses these statistics is explained. But the main problem arises in the determination of the correct application place of this algorithm. So, we have made some experiments, whose results are shown in Tables 4.7 and 4.8. Note that, all of the parameters existing in contextual statistics, root-word statistics and the determination of the learned rules are chosen after some experimentation. For example, the optimum ratios in the application of the contextual statistics, for each tour is found experimentally. But there is no need to deal with the intermediate results.

[3]Please note for ARK, in the first two rows, the training and the test texts are the same.
[4]Learning iterations have been stopped when the maximum rule score fell below 7.

| **Disambiguation Stage** | **Ambiguity** | **Recall (%)** | **Pre. (%)** |
|---|---|---|---|
| Training Set ARK | | | |
| LEARNED CHOOSE | 1.029 | 97.31 | 94.52 |
| LEARNED DELETE | 1.027 | 97.20 | 94.63 |
| Training Set C500 | | | |
| LEARNED CHOOSE | 1.031 | 97.30 | 94.45 |
| LEARNED DELETE | 1.028 | 97.30 | 94.61 |
| Training Set C1000 | | | |
| LEARNED CHOOSE | 1.028 | 97.29 | 94.58 |
| LEARNED DELETE | 1.026 | 97.18 | 94.68 |
| Training Set C2000 | | | |
| LEARNED CHOOSE | 1.028 | 97.24 | 94.60 |
| LEARNED DELETE | 1.025 | 97.13 | 94.71 |

Table 4.4: Average parses, recall and precision for text ARK after applying learned rules.

| **Disambiguation Stage** | **Ambiguity** | **Recall (%)** | **Pre. (%)** |
|---|---|---|---|
| Training Set ARK | | | |
| LEARNED CHOOSE | 1.035 | 96.64 | 93.36 |
| LEARNED DELETE | 1.029 | 96.40 | 93.71 |
| Training Set C500 | | | |
| LEARNED CHOOSE | 1.035 | 96.66 | 93.32 |
| LEARNED DELETE | 1.029 | 96.40 | 93.66 |
| Training Set C1000 | | | |
| LEARNED CHOOSE | 1.035 | 96.66 | 93.34 |
| LEARNED DELETE | 1.029 | 96.42 | 93.64 |
| Training Set C2000 | | | |
| LEARNED CHOOSE | 1.034 | 96.64 | 93.41 |
| LEARNED DELETE | 1.030 | 96.52 | 93.70 |

Table 4.5: Average parses, recall and precision for text 270 after applying learned rules.

| **Training Text** | **Choose Rules** | **Delete Rules** |
|---|---|---|
| ARK | 23 | 89 |
| C500 | 11 | 113 |
| C1000 | 29 | 195 |
| C2000 | 61 | 245 |

Table 4.6: Number of choose and delete rules learned from training texts.

| **Disambiguation Stage** | **Ambiguity** | **Recall (%)** | **Pre. (%)** |
|---|---|---|---|
| BASE | 1.719 | 100.00 | 58.18 |
| INITIAL CHOOSE | 1.353 | 99.16 | 73.27 |
| INITIAL DELETE | 1.130 | 98.73 | 87.24 |
| ROOT WORD STATISTICS | 1.116 | 98.32 | 88.10 |
| CONTEXT STATISTICS | 1.031 | 96.44 | 93.52 |
| LEARNED CHOOSE | 1.027 | 96.40 | 93.84 |
| LEARNED DELETE | 1.023 | 96.21 | 94.06 |

Table 4.7: Average parses, recall and precision for text C270, root word statistics applied after hand-crafted initial rules

We conclude from these numbers that, it will be much more better to apply the root word statistics after the contextual statistics. So, the order of the steps explained in the beginning of the previous chapter is completed. The suggested order is applying the hand-crafted rules first, then contextual and root word statistics, and last of all the learned rules.

We continued the disambiguation process of the two unseen texts, according to this order. The results for these texts are presented in tables 4.9 and 4.10.

| Disambiguation Stage | Ambiguity | Recall (%) | Pre. (%) |
|---|---|---|---|
| BASE | 1.719 | 100.00 | 58.18 |
| INITIAL CHOOSE | 1.353 | 99.16 | 73.27 |
| INITIAL DELETE | 1.130 | 98.73 | 87.24 |
| CONTEXT STATISTICS | 1.038 | 96.70 | 93.15 |
| ROOT WORD STATISTICS | 1.034 | 96.66 | 93.50 |
| LEARNED CHOOSE | 1.029 | 96.60 | 93.81 |
| LEARNED DELETE | 1.024 | 96.40 | 94.11 |

Table 4.8: Average parses, recall and precision for text C270, root word statistics applied after contextual statistics

| Disambiguation Stage | Ambiguity | Recall (%) | Pre. (%) |
|---|---|---|---|
| BASE | 1.982 | 100.00 | 50.526 |
| INITIAL CHOOSE | 1.384 | 97.00 | 69.94 |
| INITIAL DELETE | 1.192 | 96.77 | 80.73 |
| CONTEXT STATISTICS | 1.054 | 93.87 | 89.75 |
| ROOT WORD STATISTICS | 1.046 | 93.69 | 89.99 |
| LEARNED CHOOSE | 1.039 | 93.53 | 90.15 |
| LEARNED DELETE | 1.034 | 93.39 | 90.28 |

Table 4.9: Average parses, recall and precision for text EMBASSY

| Disambiguation Stage | Ambiguity | Recall (%) | Pre. (%) |
|---|---|---|---|
| BASE | 1.719 | 100.00 | 58.18 |
| INITIAL CHOOSE | 1.340 | 98.22 | 73.25 |
| INITIAL DELETE | 1.152 | 97.74 | 84.81 |
| CONTEXT STATISTICS | 1.071 | 95.93 | 89.53 |
| ROOT WORD STATISTICS | 1.054 | 95.71 | 90.74 |
| LEARNED CHOOSE | 1.052 | 95.67 | 90.89 |
| LEARNED DELETE | 1.048 | 95.48 | 91.07 |

Table 4.10: Average parses, recall and precision for text MANUAL

Tables 4.11 and 4.12 gives some additional statistical results at the sentence level, for each of the test texts. The columns labeled UA/C and A/C give the number and percentage of the sentences that are correctly disambiguated with one parse per token, and with more than one parse for at least one token, respectively. The columns labeled 1, 2, 3, and >3 denote the number and percentage of sentences that have 1, 2, 3, and >3 tokens, with all remaining parses incorrect. It can be seen that well 60% of the sentences are correctly morphologically disambiguated with very small number of ambiguous parses remaining.

| **Text** | **Sentences** | | | |
|---|---|---|---|---|
| | **Total** | **UA/C** | **A/C** | **UA/C+A/C** |
| ARK | 494 | 44.53% | 19.64% | 64.17% |
| C270 | 270 | 42.96% | 18.52% | 61.48% |
| EMBASSY | 198 | 21.21% | 12.12% | 33.33% |
| MANUAL | 204 | 34.80% | 24.51% | 59.31% |

**UA/C** : Percentage of the sentences, correctly and unambiguously disambiguated.
**A/C** : Percentage of the sentences, correctly disambiguated with at least one ambiguous token.

Table 4.11: Disambiguation results at the sentence level using rules learned from C2000.

| **Text** | **Sentences** | | | | |
|---|---|---|---|---|---|
| | **Total** | **1** | **2** | **3** | **>3** |
| ARK | 494 | 26.92% | 8.30% | 0.61% | 0.00% |
| C270 | 270 | 20.37% | 10.00% | 6.30% | 1.85% |
| EMBASSY | 198 | 25.25% | 18.18% | 11.61% | 11.61% |
| MANUAL | 204 | 26.47% | 11.77% | 0.00% | 2.45% |

Table 4.12: The distribution of the number of wrongly disambiguated tokens in the sentences

In order to see the effectiveness of the unknown word processor, we trace the words that are processed by the unknown word processor through the morphological disambiguation. As seen in Table 4.13, more than 90% of the tokens which were unknown by the morphological processor, have got their correct reading after the morphological disambiguation process.

| Text | Tokens | | | | | |
|---|---|---|---|---|---|---|
| | **Total** | **U** | **PU** | **CD** | **PU/U** | **CD/U** |
| ARK | 7928 | 219 | 207 | 198 | 94.52% | 90.41% |

**U** : Number of tokens, unknown by the morphological processor
**PU** : Number of unknown tokens, processed by the unknown word processor
**CD** : Number of unknown tokens, correctly disambiguated during the morphological disambiguation

Table 4.13: The effectiveness of the unknown word processor

## 4.1 Discussion of Results

We can make a number of observations from our results: Hand-crafted rules go a long way in improving precision substantially, but in a language like Turkish, one has to code rules that allow no, or only carefully controlled derivations, otherwise lots of things go massively wrong. Thus we have used very tight and conservative rules in hand-crafting. Although the additional impact of choose and rules that are induced by the unsupervised learning is not substantial, this is to be expected as the stage at which they are used is when all the "easy" work has been done and the more notorious cases remain. An important class of rules we explicitly have avoided hand crafting are rules for disambiguating around coordinating conjunctions. We have noted that while learning choose rules, the system zeroes in rather quickly on these contexts and comes up with rather successful rules for conjunctions. Similarly, the delete rules find some interesting situations which would be virtually impossible to enumerate. Although it is easy to formulate what things can go together in a context, it is rather impossible to formulate what things can not go together.

We have also attempted to learn rules directly without applying any hand-crafted rules, but this has resulted in a failure with the learning process getting stuck fairly early. This is mainly due to the lack of sufficient unambiguous contexts to bootstrap the whole disambiguation process.

The unseen texts MANUAL and EMBASSY are morphologically disambiguated with a very satisfactory recall and precision. Please also note that, EMBASSY is a very hard text, in the sense that MANUAL has 13.5 tokens per sentence on the average while this number is more than 26 tokens per sentence for the text EMBASSY. That means EMBASSY contains more complex sentences, that is harder to disambiguate. This is why the sentence level results are less satisfactory for the text EMBASSY.

From analysis of our results we have noted that trying to choose one correct parse for every token is rather ambitious (at least for Turkish). There are a number of reasons for this:

- There are genuine ambiguities. The word *o* is either a personal or a demonstrative pronoun (in addition to being a determiner). One simply can not choose among the first two using any amount of contextual information.

- A given word may be interpreted in more than one way but with the same inflectional features, or with features not inconsistent with the syntactic context. This usually happens when the root of one of the forms is a proper prefix of the root of the other one. One would need serious amounts of semantic, or statistical root word and word form preference information for resolving these. For instance, in

  koyun    sürüsü
  koyun    sürü+sü
  sheep    herd+POSS-3SG
  (sheep   herd)

  koy+un      sürü+sü
  bay+GEN     herd+POSS-3SG
  (bay's      herd)

  both noun phrases are syntactically possible, though the second one is obviously nonsense. It is not clear how one would disambiguate this using just contextual or syntactic information.

  Another similar example is:

| | | |
|---|---|---|
| kurmaya | yardım | etti |
| kur+ma+ya | yardım | et+ti |
| construct+INF+DAT | help | make+PAST |
| helped | construct | (something) |
| kurmay+a | yardım | et+ti |
| military-officer+DAT | help | make+PAST |
| helped | the | military-officer |

where again with have a similar problem. It may be possible to resolve this one using subcategorization constraints on the object of the verb *kur* assuming it is in the very near preceding context, but this may be very unlikely as Turkish allows arbitrary adjuncts between the object and the verb.

- Turkish allows sentences to consist of a number of sentences separated by commas. Hence locating a verb in the middle of a sentence is rather difficult, as certain verbal forms also have an adjectival reading, and punctuation is not very helpful as commas have many other uses.

- The distance between two constituents (of, say, a noun phrase) that have to agree in various morphosyntactic features may be arbitrarily long and this causes occasional mismatches, especially if the right nominal constituent has a surface plural marker which causes a 4-way ambiguity, as in *masaları*.

```
masalarI
1. [[CAT NOUN] [ROOT masa] [AGR 3PL]
    [POSS NONE] [CASE ACC]]
    (tables accusative)

2. [[CAT NOUN] [ROOT masa] [AGR 3PL]
    [POSS 3SG] [CASE NOM]]
    (his tables)

3. [[CAT NOUN] [ROOT masa] [AGR 3PL]
    [POSS 3PL] [CASE NOM]]
    (their tables)
```

```
4. [[CAT NOUN] [ROOT masa] [AGR 3SG]
   [POSS 3PL] [CASE NOM]]
   (their table)
```

Choosing among the last three is rather problematic if the corresponding genitive form to force agreement with is outside the context.

Among these problems, the most crucial is the second one which we believe can be solved to a great extent by using the scoring mechanism in the application of the hand-crafted rules. We have already designed the algorithm, and present it in the next chapter.

# Chapter 5

# Conclusions

This thesis had presented a constraint-based morphological disambiguation approach, which combines a set of hand-crafted constraint rules, corpus statistics and learns additional rules to choose and delete parses, from untagged text in an unsupervised manner. We have extended the rule learning and application schemes so that the impact of various morphological phenomena and features are selectively taken into account.

Before the disambiguation process, a very robust preprocessing is implemented on the raw text. After the tokenization module divides the text into its tokens, a lexical and non-lexical recognizer captures the predefined lexical (e.g. *devam et* (continue), Mustafa Kemal Atatürk) and non-lexical (e.g. *koşa koşa* (running), *yapar yapmaz* (as soon as s/he does)) collocations, and forms an intermediate form of the text, containing its tokens with corresponding parses. A format conversion is applied at this step, so that the inflectional feature of the parse, with the last category conversion (if any) is at the top level, and as a result, a hierarchical feature structure is obtained. The preprocessing phase also prepares a projected form in order to be used in learning additional constraints.

We have applied our approach to morphological disambiguation of Turkish, a free–constituent order language, with agglutinative morphology, exhibiting productive inflectional and derivational processes. We have also incorporated a rather sophisticated unknown form processor which extracts any relevant inflectional or derivational markers even if the root word is unknown. For a corpus, containing 39,800 tokens, only 12 of them are left as *unknown*, which are generally foreign proper names, which take suffixes according to their pronunciation, instead of spelling, and some words, which have spelling mistakes. When we trace the tokens processed by the unknown word processor, we see that more than 90% of them have received their intended readings.

Our results indicate that by combining these hand-crafted, statistical and learned information sources, we can attain a recall of 96 to 97% with a corresponding precision of 93 to 94% and ambiguity of 1.02 to 1.03 parses per token, on test texts, however the impact of the rules that are learned is not significant as hand-crafted rules do most of the easy work at the initial stages.

The results are also satisfactory, when we do the same experiments on two unseen texts, which are on completely different topics. The recall we reach is 93-95% with a corresponding precision of 90-91% and ambiguity of 1.03 to 1.04 parses per token.

Note that these results are obtained from a certain application order of our morphological disambiguation sources. We first apply the hand-crafted rules, then contextual statistics and root-category statistics and terminate with the learned rules. This order is very important in the sense that every source of morphological disambiguation is effective on its right place. If root-category statistics are applied as a first step, the result would be a disappointment unlike the English case. Consider the token *bir* (DET(one) or ADVERB(if only)), which is generally used as a determiner. If these statistical information is used initially, all of the parses of this token would be assumed to be determiner, although it has some adverbial readings, which is not too rare.

Our next aim is to improve the obtained results. We have some readily defined ideas for an enhanced morphological disambiguation. We observe that after applying the hand-crafted rules, we sacrifice about 1% from the recall, which is a very high ratio for us, because the errors made on this step affects the success of other steps, such as using context statistics for further disambiguation. The success of the learned rules is also affected, because such rules are learned from the corpus, which is initially disambiguated by these hand-crafted rules. For these reasons, we decide to make the initial disambiguation phase independent from the order of the constraints. Instead of immediately choosing or deleting a corresponding parse, which is the main cause of error in this phase, we keep a score for each parse, and if a parse of a token is chosen, its score will be incremented, and similarly if a parse is said to be deleted by some rule, its score will be decremented. After this scoring operation is completed, the parses with a very low score according to some ratio or difference will be deleted. Motivated by Alshawi and Carter's work [1], we want to incorporate weights for the contributions of the preference constraints used in disambiguation. Using the weights, the scores of the parses will be incremented or decremented according to the weight of the constraint. For example a very strong constraint will have a higher weight. These weights will be obtained according to manually created gold standard disambiguated texts. This method was applied to an English corpus ATIS by Alshawi and Carter and got more successful results than those derived by a labour intensive hand tuning effort.

# Appendix A

# Sample Text

This sample text is consisted of sentences taken from the text ARK. Upper cases indicate one of the non-ASCII special Turkish characters: e.g., `G` denotes ğ, `U` denotes ü, `I` denotes ı, `S` denotes ş, `C` denotes ç, `O` denotes ö, and there is no other capital letter used in the text.

> arkeologlar, kazI yapmanIn yanI sIra, o kazI yerini Cevreleyen alanIn eski biCimini de yeniden kurmaya CalISIrlar. ilk evler, burada kuzey Irak'ta kermezdere'deki proto neolitik dOnem evinde gOrUldUGU gibi, topraGa gOmUlU yuvarlak kulUbelerdi. taS ve tahta aletler arasInda, iGne, dikiS iGnesi, bIz, ok baSI, mIzrak ya da zIpkIn uClarI, vb., bulunuyordu. proto neolitik eriha'nIn en dikkat Cekici Ozelliklerinden biri, surlarIn iC tarafIna yapISIk taS kuleydi. 10 m. CapIndaki kulenin 8 metrenin UstUnde bir bOlUmU bugUn de ayaktadIr. doGu tarafInda 1.7 m yUksekliGinde bir kapI, her biri tek bir taS bloGundan yapIlmIS 22 basamaklI bir merdivene aCIlIr. tell brak'taki mO. 4. binyIl tapInaGInda 300'U aSkIn (ayrIca parCa halinde binlerce) taS ya da piSmiS kilden yapIlmIS "gOz putu" bulunmuStur. tapInakta, yUkseklikleri 2 ile 11 cm arasInda deGiSen bu adak simgelerinden 20,000 - 22,000 kadar bulunduGu hesaplanmIStIr. niceliksel CalISmalar yapIlacaGI zaman, temsili nitelikte Ornekler elde etmenin temel yOntemlerinden biri elemedir. kumlu toprakta kuru eleme zaman zaman mUmkUndUr.

# Appendix B

# The Collocation Database

## B.1 Non-Lexicalized Collocations

```
# e.g. koSa koSa

[ROOT=_R][SENSE=POS][TAM1=OPT][AGR=3SG]  [ROOT=_R][SENSE=POS][TAM1=OPT][AGR=3SG]
[[CAT=ADVERB][ROOT=%s][SENSE=POS][TYPE=MANNER]]

# e.g. yapar yapmaz

[ROOT=_R][SENSE=POS][TAM1=AORIST-AR][AGR=3SG]  [ROOT=_R][SENSE=NEG][TAM1=AORIST][AGR=3SG]
[[CAT=ADVERB][ROOT=%s][SENSE=POS][TYPE=MANNER]]

[ROOT=_R][SENSE=POS][TAM1=AORIST-HR][AGR=3SG]  [ROOT=_R][SENSE=NEG][TAM1=AORIST][AGR=3SG]
[[CAT=ADVERB][ROOT=%s][SENSE=POS][TYPE=MANNER]]

[ROOT=_R][SENSE=POS][TAM1=AORIST-R][AGR=3SG]  [ROOT=_R][SENSE=NEG][TAM1=AORIST][AGR=3SG]
[[CAT=ADVERB][ROOT=%s][SENSE=POS][TYPE=MANNER]]

# e.g. yapIp yapmama

[ROOT=_R][CAT=VERB][SENSE=POS][CONV=ADVERB=YIP]
         [ROOT=_R][CAT=VERB][SENSE=NEG][CONV=NOUN=MA][TYPE=INFINITIVE]
[[CAT=NOUN][ROOT=%s]]
```

```
# e.g. gUrUl gUrUl

[WORD=_W][CAT=DUP] [WORD=_W][CAT=DUP]
[[CAT=ADVERB][ROOT=%s]]

# e.g. ev ev

[WORD=_W][CAT=NOUN][CASE=NOM] [WORD=_W][CAT=NOUN][CASE=NOM]
[[CAT=ADVERB][ROOT=%s]]

# e.g. gUzel gUzel

[WORD=_W][CAT=ADJ] [WORD=_W][CAT=ADJ]
[[CAT=ADVERB][ROOT=%s]]

# e.g. gUzel mi gUzel

[WORD=_W][CAT=ADJ] [ROOT=mi] [WORD=_W][CAT=ADJ]
[[CAT=ADJ][ROOT=%s]]

# e.g. ev be ev

[WORD=_W][CAT=NOUN] [ROOT=be] [WORD=_W][CAT=NOUN]
[[CAT=ADVERB][ROOT=%s]]
```

## B.2 Fixed Lexicalized Collocations

```
baSlI baSIna
[[CAT=ADJ][ROOT=%s]]

hiC kimse
[[CAT=PRONOUN][ROOT=%s]]

ne denli
[[CAT=ADVERB][ROOT=%s]]

son derece
[[CAT=ADVERB][ROOT=%s]]
```

```
Simdiye dek
[[CAT=ADVERB][ROOT=%s]]

okur yazar
[[CAT=ADJ][ROOT=%s]]

okur - yazar
[[CAT=ADJ][ROOT=%s]]

hiC olmazsa
[[CAT=ADVERB][ROOT=%s]]

belli baSlI
[[CAT=ADJ][ROOT=%s]]

az kalsIn
[[CAT=ADVERB][ROOT=%s]]

can havliyle
[[CAT=ADVERB][ROOT=%s]]

allaha IsmarladIk
[[CAT=EXC][ROOT=%s]]

tek tUk
[[CAT=ADJ][ROOT=%s]]

Ote yandan
[[CAT=CONN][ROOT=%s]]

ya da
[[CAT=CONN][ROOT=%s]]

yanI sIra
[[CAT=POSTP][ROOT=%s][SUBCAT=GEN]]

ne menem
[[CAT=ADJ][ROOT=%s]]

ipe sapa gelmez
[[CAT=ADJ][ROOT=%s]]
```

’ ’

[[CAT=PUNCT][ROOT=%s]]

‘ ‘

[[CAT=PUNCT][ROOT=%s]]

geCmiS olsuna

[[CAT=ADVERB][ROOT=%s]]

## B.3 Inflectable Lexicalized Collocations

| | | | |
|---|---|---|---|
| new york | ismet inOnU | kazIm karabekir | afyon karahisar |
| adnan adIvar | ali fuat | Cerkez ethem | mustafa kemal atatUrk |
| mustafa kemal | kemal atatUrk | hasan pulur | mazhar mUfit kansu |
| evliya Celebi | bosna hersek | tUrk - iS | amerikan airlines |
| | | | |
| dikkat CeK | baSta gel | Onde gel | karSIlIk gel |
| ele al | yol aJ | etkili ol | gOze Carp |
| yok ol | sahip ol | iliSkili ol | rolU ol |
| yararlI ol | Onemli ol | mUmkUn ol | neden ol |
| hedef ol | yardImcI ol | deGiSik ol | mevcut ol |
| yol gOsterici ol | yol aJ | uygun ol | felaket ol |
| sOz konusu ol | sahne ol | fikir sahibi ol | mutlu ol |
| yok ol | arkadaS ol | sebep ol | destek ol |
| gUClU ol | yUrUrlUkte ol | elinde ol | belli ol |
| hazIr ol | Uye ol | teslim ol | gUvence altInda ol |
| COzUm ol | baGlI ol | kolay ol | yeterli ol |
| geCerli ol | konu ol | harap ol | kardeS ol |
| gUC ol | SikayetCi ol | yasak ol | kayIp ol |
| sorumlu ol | aday ol | kaynak ol | kUs ol |
| kUl ol | son ol | alabora ol | boyun eG |
| ait ol | mUteSekkUr ol | mUteSekkir ol | Sart ol |
| zorunlu ol | baSarIlI ol | dikkatli ol | engel ol |
| tedavi ol | kalabalIk ol | hakim ol | mahkum ol |
| felC ol | altUst ol | alt Ust ol | gOzetim altIna al |
| taklit ed | tehdit ed | devam ed | hizmet ed |
| para ed | temin ed | hareket ed | telefon ed |
| baS ed | yolculuk ed | takdir ed | takip ed |
| eSlik ed | temsil ed | tahliye ed | kabul ed |
| ifade ed | rekabet ed | hareket ed | meSgul ed |
| inSa ed | ziyaret ed | iptal ed | yardIm ed |
| tahmin ed | terk ed | tahammUl ed | tatmin ed |
| gOzardI ed | kontrol ed | ilan ed | hitap ed |
| itiraf ed | iSgal ed | mUcadele ed | ihlal ed |
| gOC ed | tedirgin ed | merak ed | ihanet ed |
| alay ed | ithal ed | ihraC ed | seyahat ed |
| yerle bir ed | yerlebir ed | teSvik ed | mUdahale ed |
| hitap ed | davet ed | arz ed | iSaret ed |
| aktive ed | tahrik ed | teShis ed | dua ed |
| iddia ed | akIn ed | teSekkUr ed | istifa ed |
| tercih ed | dans ed | protesto ed | sohbet ed |
| ikna ed | yok ed | sOz ed | tedavi ed |
| dezenfekte ed | ikram ed | tahliye ed | iptal ed |
| elde ed | sInIrdISI ed | istismar ed | endiSe ed |
| fark ed | veto ed | maGlup ed | havale ed |
| tolere ed | ayIrt ed | iade ed | sOzUnU ed |
| yerinden ed | yerlerinden ed | garanti ed | parafe ed |
| ateS ed | imha ed | idare ed | ameliyat ed |
| idam ed | not ed | gOzardI ed | tahsis ed |
| tercUme ed | tahrip ed | hayal ed | rapor ed |
| mahkum ed | modifiye ed | rica ed | akIn ed |
| merak ed | hediye ed | tanIklIk ed | veda ed |
| Sikayet ed | tavsiye ed | tercih ed | bertaraf ed |
| tesis ed | tereddUt ed | inkar ed | empoze ed |
| ibadet ed | dikkat ed | zarar ed | isabet ed |
| tahakkuk ed | OncUlUk ed | itibar ed | kat ed |
| alt ed | altUst ed | alt Ust ed | aGzInda sakIz ed |
| iltica ed | Umit ed | takviye ed | yemin ed |
| sonuC al | imza al | gOrevden al | Onlem al |
| gOze al | haber al | satIn al | esir al |
| dikkate al | gOz OnUne al | filme al | konu al |
| sonuC al | elinden al | ele al | tedavi altIna al |
| One sUr | geri al | karar al | Ornek al |
| esas al | askIya al | koruma altIna al | incelemeye al |
| ciddiye al | bilgi al | OvgU al | zevk al |
| yer al | yol al | izin al | nefes al |
| tavIr al | hedef al | zaman al | oy al |
| eGitim al | ilgi CeK | ceza CeK | acI CeK |
| fotoGraf CeK | geri CeK | sIkIntI CeK | gUClUk CeK |
| oyun oyna | rol oyna | ortaya koy | karSI koy |
| el koy | imza koy | ambargo koy | yasa koy |
| ele geJ | abayI yaK | hiCe say | sona er |
| gerek gOr | gerek kal | gerek duy | kayda geJ |
| yok ol | var ol | geri kal | karSI CIK |
| ortaya CIK | idealize ed | | |

# Appendix C

# Sample Preprocessed Text

```
[[@,
  [[cat:beginning_of_sentence]]],
 [arkeologlar,
  [[cat:noun,stem:[cat:noun,root:arkeoloji],suffix:og,agr:'3PL',poss:'NONE',case:nom],
   [cat:verb,stem:[cat:noun,stem:[cat:noun,root:arkeoloji],suffix:og,agr:'3SG',
    poss:'NONE',case:nom],suffix:none,tam2:pres,agr:'3PL']]],
 [',',
  [[cat:punct,root:',']]],
 [kazI,
  [[cat:noun,root:kazI,agr:'3SG',poss:'NONE',case:nom],
   [cat:noun,root:kaz,agr:'3SG',poss:'NONE',case:acc],
   [cat:noun,root:kaz,agr:'3SG',poss:'3SG',case:nom],
   [cat:verb,root:kazI,sense:pos,tam1:imp,agr:'2SG']]],
 [yapmanIn,
  [[cat:noun,stem:[cat:verb,root:yap,sense:pos],suffix:ma,type:infinitive,
    agr:'3SG',poss:'NONE',case:gen],
   [cat:noun,stem:[cat:verb,root:yap,sense:pos],suffix:ma,type:infinitive,
    agr:'3SG',poss:'2SG',case:gen]]],
 ['yanI sIra',
  [[cat:postp,root:'yanI sIra',subcat:gen]]],
 [',',
  [[cat:punct,root:',']]],
 [o,
  [[cat:adj,root:o,type:determiner],
   [cat:exc,root:o],
   [cat:pronoun,root:o,type:demons,agr:'3SG',poss:'NONE',case:nom],
```

```
  [cat:pronoun,root:o,type:personal,agr:'3SG',poss:'NONE',case:nom]]],
[kazI,
 [[cat:noun,root:kazI,agr:'3SG',poss:'NONE',case:nom],
  [cat:noun,root:kaz,agr:'3SG',poss:'NONE',case:acc],
  [cat:noun,root:kaz,agr:'3SG',poss:'3SG',case:nom],
  [cat:verb,root:kazI,sense:pos,tam1:imp,agr:'2SG']]],
[yerini,
 [[cat:noun,root:yer,agr:'3SG',poss:'2SG',case:acc],
  [cat:noun,root:yer,agr:'3SG',poss:'3SG',case:acc]]],
['Cevreleyen',
 [[cat:adj,stem:[cat:verb,root:'Cevrele',sense:pos],suffix:yan]]],
[alanIn,
 [[cat:noun,root:alan,agr:'3SG',poss:'NONE',case:gen],
  [cat:noun,root:alan,agr:'3SG',poss:'2SG',case:nom],
  [cat:noun,root:ala,agr:'3SG',poss:'NONE',case:gen],
  [cat:noun,root:ala,agr:'3SG',poss:'2SG',case:gen],
  [cat:noun,stem:[cat:adj,stem:[cat:verb,root:al,sense:pos],suffix:yan],suffix:none,
   agr:'3SG',poss:'NONE',case:gen],
  [cat:noun,stem:[cat:adj,stem:[cat:verb,root:al,sense:pos],suffix:yan],suffix:none,
   agr:'3SG',poss:'2SG',case:nom]]],
[eski,
 [[cat:adj,root:eski],
  [cat:noun,stem:[cat:adj,root:eski],suffix:none,agr:'3SG',poss:'NONE',case:nom],
  [cat:verb,root:eski,sense:pos,tam1:imp,agr:'2SG']]],
[biCimini,
 [[cat:noun,root:biCim,agr:'3SG',poss:'2SG',case:acc],
  [cat:noun,root:biCim,agr:'3SG',poss:'3SG',case:acc]]],
[de,
 [[cat:conn,root:de],
  [cat:verb,root:de,sense:pos,tam1:imp,agr:'2SG']]],
[yeniden,
 [[cat:noun,stem:[cat:adj,root:yeni],suffix:none,agr:'3SG',poss:'NONE',case:abl],
  [cat:adverb,root:yeniden]]],
[kurmaya,
 [[cat:noun,root:kurmay,agr:'3SG',poss:'NONE',case:dat],
  [cat:verb,root:kur,sense:neg,tam1:opt,agr:'3SG'],
  [cat:noun,stem:[cat:verb,root:kur,sense:pos],suffix:ma,type:infinitive,
   agr:'3SG',poss:'NONE',case:dat]]],
['CalISIrlar',
 [[cat:verb,root:'CalIS',sense:pos,tam1:(aorist),agr:'3PL']]],
['.',
 [[cat:punct,root:'.']]],
```

```
 [#,
  [[cat:end_of_sentence]]]].

[[@,
  [[cat:beginning_of_sentence]]],
 [ilk,
  [[cat:adj,root:ilK],
   [cat:noun,stem:[cat:adj,root:ilK],suffix:none,agr:'3SG',poss:'NONE',case:nom],
   [cat:adverb,root:ilK]]],
 [evler,
  [[cat:noun,root:ev,agr:'3PL',poss:'NONE',case:nom],
   [cat:verb,stem:[cat:noun,root:ev,agr:'3SG',poss:'NONE',case:nom],suffix:none,
    tam2:pres,agr:'3PL']]],
 [',',
  [[cat:punct,root:',']]],
 [burada,
  [[cat:noun,root:bura,agr:'3SG',poss:'NONE',case:loc]]],
 [kuzey,
  [[cat:adj,root:kuzey],
   [cat:noun,root:kuzey,agr:'3SG',poss:'NONE',case:nom]]],
 ['Irak\'ta',
  [[cat:noun,root:'Irak',type:rproper,agr:'3SG',poss:'NONE',type:proper,case:loc]]],
 ['kermezdere\'deki',
  [[cat:adj,stem:[cat:noun,root:kermezdere,agr:'3SG',poss:'NONE',type:proper,case:loc],
    suffix:rel],
   [cat:noun,stem:[cat:adj,stem:[cat:noun,root:kermezdere,agr:'3SG',poss:'NONE',
    type:proper,case:loc],suffix:rel],suffix:none,agr:'3SG',poss:'NONE',case:nom],
   [cat:adj,stem:[cat:noun,root:kermezDere,agr:'3SG',poss:'NONE',type:proper,case:loc],
    suffix:rel],
   [cat:noun,stem:[cat:adj,stem:[cat:noun,root:kermezDere,agr:'3SG',poss:'NONE',
    type:proper,case:loc],suffix:rel],suffix:none,agr:'3SG',poss:'NONE',case:nom]]],
 [proto,
  [[cat:noun,root:proto,agr:'3SG',poss:'NONE',case:nom]]],
 [neolitik,
  [[cat:adj,root:neolitik],
   [cat:noun,stem:[cat:adj,root:neolitik],suffix:none,agr:'3SG',poss:'NONE',case:nom]]],
 [dOnem,
  [[cat:noun,root:dOnem,agr:'3SG',poss:'NONE',case:nom]]],
 [evinde,
  [[cat:noun,root:evin,agr:'3SG',poss:'NONE',case:loc],
   [cat:noun,root:ev,agr:'3SG',poss:'2SG',case:loc],
   [cat:noun,root:ev,agr:'3SG',poss:'3SG',case:loc]]],
```

```
 [gOrUldUGU,
  [[cat:adj,stem:[cat:verb,root:gOr,voice:(pass),sense:pos],suffix:dik,poss:'3SG'],
   [cat:noun,stem:[cat:verb,root:gOr,voice:(pass),sense:pos],suffix:dik,
    agr:'3SG',poss:'3SG',case:nom]]],
 [gibi,
  [[cat:postp,root:gibi,subcat:nom]]],
 [',',
  [[cat:punct,root:',']]],
 [topraGa,
  [[cat:noun,root:toprak,agr:'3SG',poss:'NONE',case:dat]]],
 [gOmUlU,
  [[cat:adj,root:gOmUlU],
   [cat:noun,stem:[cat:adj,root:gOmUlU],suffix:none,
    agr:'3SG',poss:'NONE',case:nom],
   [cat:adj,stem:[cat:noun,root:gOmU],suffix:li],
   [cat:noun,stem:[cat:adj,stem:[cat:noun,root:gOmU],suffix:li],suffix:none,
    agr:'3SG',poss:'NONE',case:nom]]],
 [yuvarlak,
  [[cat:adj,root:yuvarlak],
   [cat:noun,stem:[cat:adj,root:yuvarlak],suffix:none,agr:'3SG',poss:'NONE',case:nom]]],
 [kulUbelerdi,
  [[cat:verb,stem:[cat:noun,root:kulUbe,agr:'3PL',poss:'NONE',case:nom],suffix:none,
    tam2:past,agr:'3SG']]],
 ['.',
  [[cat:punct,root:'.']]],
 [#,
  [[cat:end_of_sentence]]]].

[[@,
  [[cat:beginning_of_sentence]]],
 [taS,
  [[cat:adj,root:taS],
   [cat:noun,root:taS,agr:'3SG',poss:'NONE',case:nom],
   [cat:verb,root:taS,sense:pos,tam1:imp,agr:'2SG']]],
 [ve,
  [[cat:conn,root:ve]]],
 [tahta,
  [[cat:adj,root:tahta],
   [cat:noun,root:tahta,agr:'3SG',poss:'NONE',case:nom],
   [cat:noun,root:taht,agr:'3SG',poss:'NONE',case:dat]]],
 [aletler,
  [[cat:noun,root:alet,agr:'3PL',poss:'NONE',case:nom],
```

```
  [cat:verb,stem:[cat:noun,root:alet,agr:'3SG',poss:'NONE',case:nom],suffix:none,
   tam2:pres,agr:'3PL']]],
[arasInda,
 [[cat:noun,root:ara,agr:'3SG',poss:'3SG',case:loc]]],
[',',
 [[cat:punct,root:',']]],
[iGne,
 [[cat:noun,root:iGne,agr:'3SG',poss:'NONE',case:nom]]],
[',',
 [[cat:punct,root:',']]],
[dikiS,
 [[cat:noun,stem:[cat:verb,root:diK,sense:pos],suffix:yis,agr:'3SG',poss:'NONE',
   case:nom],
  [cat:verb,root:diK,voice:recip,sense:pos,tam1:imp,agr:'2SG']]],
[iGnesi,
 [[cat:noun,root:iGne,agr:'3SG',poss:'3SG',case:nom]]],
[',',
 [[cat:punct,root:',']]],
[bIz,
 [[cat:noun,root:bIz,agr:'3SG',poss:'NONE',case:nom]]],
[',',
 [[cat:punct,root:',']]],
[ok,
 [[cat:noun,root:oK,agr:'3SG',poss:'NONE',case:nom]]],
[baSI,
 [[cat:noun,root:baS,agr:'3SG',poss:'NONE',case:acc],
  [cat:noun,root:baS,agr:'3SG',poss:'3SG',case:nom]]],
[',',
 [[cat:punct,root:',']]],
[mIzrak,
 [[cat:noun,root:mIzrak,agr:'3SG',poss:'NONE',case:nom]]],
['ya da',
 [[cat:conn,root:'ya da']]],
[zIpkIn,
 [[cat:noun,root:zIpkIn,agr:'3SG',poss:'NONE',case:nom]]],
[uClarI,
 [[cat:noun,root:uC,agr:'3PL',poss:'NONE',case:acc],
  [cat:noun,root:uC,agr:'3PL',poss:'3SG',case:nom],
  [cat:noun,root:uC,agr:'3PL',poss:'3PL',case:nom],
  [cat:noun,root:uC,agr:'3SG',poss:'3PL',case:nom]]],
[',',
 [[cat:punct,root:',']]],
```

```
 ['vb.',
  [[cat:noun,root:'vb.E',type:rproper,agr:'3SG',poss:'NONE',case:nom]]],
 [',',
  [[cat:punct,root:',']]],
 [bulunuyordu,
  [[cat:verb,root:bulun,sense:pos,tam1:prog1,tam2:past,agr:'3SG'],
   [cat:verb,root:bul,voice:(pass),sense:pos,tam1:prog1,tam2:past,agr:'3SG']]],
 ['.',
  [[cat:punct,root:'.']]],
 [#,
  [[cat:end_of_sentence]]]].

[[@,
  [[cat:beginning_of_sentence]]],
 [proto,
  [[cat:noun,root:proto,agr:'3SG',poss:'NONE',case:nom]]],
 [neolitik,
  [[cat:adj,root:neolitik],
   [cat:noun,stem:[cat:adj,root:neolitik],suffix:none,agr:'3SG',poss:'NONE',case:nom]]],
 ['eriha\'nIn',
  [[cat:noun,root:eriha,type:rproper,agr:'3SG',poss:'NONE',type:proper,case:gen],
   [cat:noun,root:eriha,type:rproper,agr:'3SG',type:proper,poss:'2SG',case:gen]]],
 [en,
  [[cat:adverb,root:en],
   [cat:noun,root:en,agr:'3SG',poss:'NONE',case:nom]]],
 ['dikkat Cekici',
  [[cat:adj,stem:[cat:verb,root:'dikkat
 CeK',sense:pos],suffix:yici],
   [cat:noun,stem:[cat:verb,root:'dikkat
 CeK',sense:pos],suffix:yici,agr:'3SG',poss:'NONE',case:nom]]],
 ['Ozelliklerinden',
  [[cat:noun,stem:[cat:adj,root:'Ozel'],suffix:lik,agr:'3PL',poss:'2SG',case:abl],
   [cat:noun,stem:[cat:adj,root:'Ozel'],suffix:lik,agr:'3PL',poss:'3SG',case:abl],
   [cat:noun,stem:[cat:adj,root:'Ozel'],suffix:lik,agr:'3PL',poss:'3PL',case:abl],
   [cat:noun,stem:[cat:adj,root:'Ozel'],suffix:lik,agr:'3SG',poss:'3PL',case:abl],
   [cat:noun,root:'Ozellik',agr:'3PL',poss:'2SG',case:abl],
   [cat:noun,root:'Ozellik',agr:'3PL',poss:'3SG',case:abl],
   [cat:noun,root:'Ozellik',agr:'3PL',poss:'3PL',case:abl],
   [cat:noun,root:'Ozellik',agr:'3SG',poss:'3PL',case:abl]]],
 [biri,
  [[cat:noun,stem:[cat:adj,root:bir,type:cardinal],suffix:none,agr:'3SG',poss:'NONE',
    case:acc],
```

```
    [cat:noun,stem:[cat:adj,root:bir,type:cardinal],suffix:none,agr:'3SG',poss:'3SG',
     case:nom],
    [cat:pronoun,root:biri,type:quant,agr:'3SG',poss:'3SG',case:nom]]],
  [',',
   [[cat:punct,root:',']]],
  [surlarIn,
   [[cat:noun,root:sur,agr:'3PL',poss:'NONE',case:gen],
    [cat:noun,root:sur,agr:'3PL',poss:'2SG',case:nom]]],
  [iC,
   [[cat:adj,root:iJ],
    [cat:noun,root:iJ,agr:'3SG',poss:'NONE',case:nom],
    [cat:verb,root:iJ,sense:pos,tam1:imp,agr:'2SG']]],
  [tarafIna,
   [[cat:noun,root:taraf,agr:'3SG',poss:'2SG',case:dat],
    [cat:noun,root:taraf,agr:'3SG',poss:'3SG',case:dat]]],
  [yapISIk,
   [[cat:adj,root:yapISIk],
    [cat:noun,stem:[cat:adj,root:yapISIk],suffix:none,agr:'3SG',poss:'NONE',case:nom]]],
  [taS,
   [[cat:adj,root:taS],
    [cat:noun,root:taS,agr:'3SG',poss:'NONE',case:nom],
    [cat:verb,root:taS,sense:pos,tam1:imp,agr:'2SG']]],
  [kuleydi,
   [[cat:verb,stem:[cat:noun,root:kule,agr:'3SG',poss:'NONE',case:nom],suffix:none,
     tam2:past,agr:'3SG']]],
  ['.',
   [[cat:punct,root:'.']]],
  [#,
   [[cat:end_of_sentence]]]].

[[@,
   [[cat:beginning_of_sentence]]],
  ['10',
   [[cat:adj,type:cardinal,root:'10']]],
  ['m.',
   [[cat:noun,root:'m.E',type:rproper,agr:'3SG',poss:'NONE',case:nom]]],
  ['CapIndaki',
   [[cat:adj,stem:[cat:noun,root:'Cap',agr:'3SG',poss:'2SG',case:loc],suffix:rel],
    [cat:adj,stem:[cat:noun,root:'Cap',agr:'3SG',poss:'3SG',case:loc],suffix:rel]]],
  [kulenin,
   [[cat:noun,root:kule,agr:'3SG',poss:'NONE',case:gen],
    [cat:noun,root:kule,agr:'3SG',poss:'2SG',case:gen]]],
```

```
 ['8',
  [[cat:adj,type:cardinal,root:'8']]],
 [metrenin,
  [[cat:noun,root:metre,agr:'3SG',poss:'NONE',case:gen],
   [cat:noun,root:metre,agr:'3SG',poss:'2SG',case:gen]]],
 ['UstUnde',
  [[cat:noun,stem:[cat:adj,root:'UstUn'],suffix:none,agr:'3SG',poss:'NONE',case:loc],
   [cat:noun,root:'Ust',agr:'3SG',poss:'2SG',case:loc],
   [cat:noun,root:'Ust',agr:'3SG',poss:'3SG',case:loc]]],
 [bir,
  [[cat:adj,root:bir,type:cardinal],
   [cat:adj,root:bir,type:determiner],
   [cat:adverb,root:bir]]],
 [bOlUmU,
  [[cat:noun,root:bOlUm,agr:'3SG',poss:'NONE',case:acc],
   [cat:noun,root:bOlUm,agr:'3SG',poss:'3SG',case:nom],
   [cat:noun,root:bOlU,agr:'3SG',poss:'1SG',case:acc]]],
 [bugUn,
  [[cat:adverb,root:bugUn],
   [cat:noun,root:bugUn,type:temp2,agr:'3SG',poss:'NONE',case:nom]]],
 [de,
  [[cat:conn,root:de],
   [cat:verb,root:de,sense:pos,tam1:imp,agr:'2SG']]],
 [ayaktadIr,
  [[cat:verb,stem:[cat:noun,root:ayak,agr:'3SG',poss:'NONE',case:loc],suffix:none,
    tam2:pres,copula:'2',agr:'3SG']]],
 ['.',
  [[cat:punct,root:'.']]],
 [#,
  [[cat:end_of_sentence]]]].

[[@,
  [[cat:beginning_of_sentence]]],
 [doGu,
  [[cat:adj,root:doGu],
   [cat:noun,root:doGu,agr:'3SG',poss:'NONE',case:nom]]],
 [tarafInda,
  [[cat:noun,root:taraf,agr:'3SG',poss:'2SG',case:loc],
   [cat:noun,root:taraf,agr:'3SG',poss:'3SG',case:loc]]],
 ['1.7',
  [[cat:adj,type:real,root:'1.7']]],
 ['m
```

```
',
 [[cat:noun,root:mE,type:rproper,agr:'3SG',poss:'NONE',case:nom]]],
[yUksekliGinde,
 [[cat:noun,stem:[cat:adj,root:yUksek],suffix:lik,agr:'3SG',poss:'2SG',case:loc],
  [cat:noun,stem:[cat:adj,root:yUksek],suffix:lik,agr:'3SG',poss:'3SG',case:loc]]],
[bir,
 [[cat:adj,root:bir,type:cardinal],
  [cat:adj,root:bir,type:determiner],
  [cat:adverb,root:bir]]],
[kapI,
 [[cat:noun,root:kapI,agr:'3SG',poss:'NONE',case:nom]]],
[',',
 [[cat:punct,root:',']]],
[her,
 [[cat:adj,root:her,type:determiner]]],
[biri,
 [[cat:noun,stem:[cat:adj,root:bir,type:cardinal],suffix:none,agr:'3SG',poss:'NONE',
   case:acc],
  [cat:noun,stem:[cat:adj,root:bir,type:cardinal],suffix:none,agr:'3SG',poss:'3SG',
   case:nom],
  [cat:pronoun,root:biri,type:quant,agr:'3SG',poss:'3SG',case:nom]]],
[tek,
 [[cat:adj,root:teK],
  [cat:noun,stem:[cat:adj,root:teK],suffix:none,agr:'3SG',poss:'NONE',case:nom]]],
[bir,
 [[cat:adj,root:bir,type:cardinal],
  [cat:adj,root:bir,type:determiner],
  [cat:adverb,root:bir]]],
[taS,
 [[cat:adj,root:taS],
  [cat:noun,root:taS,agr:'3SG',poss:'NONE',case:nom],
  [cat:verb,root:taS,sense:pos,tam1:imp,agr:'2SG']]],
[bloGundan,
 [[cat:noun,root:blok,agr:'3SG',poss:'2SG',case:abl],
  [cat:noun,root:blok,agr:'3SG',poss:'3SG',case:abl]]],
[yapIlmIS,
 [[cat:verb,root:yap,voice:(pass),sense:pos,tam1:narr,agr:'3SG'],
  [cat:adj,stem:[cat:verb,root:yap,voice:(pass),sense:pos,tam1:narr],suffix:none]]],
['22',
 [[cat:adj,type:cardinal,root:'22']]],
[basamaklI,
 [[cat:adj,stem:[cat:noun,root:basamak],suffix:li],
```

```
  [cat:noun,stem:[cat:adj,stem:[cat:noun,root:basamak],suffix:li],suffix:none,
   agr:'3SG',poss:'NONE',case:nom]]],
 [bir,
  [[cat:adj,root:bir,type:cardinal],
   [cat:adj,root:bir,type:determiner],
   [cat:adverb,root:bir]]],
 [merdivene,
  [[cat:noun,root:merdiven,agr:'3SG',poss:'NONE',case:dat]]],
 [aCIlIr,
  [[cat:verb,root:aJ,voice:(pass),sense:pos,tam1:(aorist),agr:'3SG'],
   [cat:adj,stem:[cat:verb,root:aJ,voice:(pass),sense:pos,tam1:(aorist)],suffix:none]]],
 ['.',
  [[cat:punct,root:'.']]],
 [#,
  [[cat:end_of_sentence]]]].

[[@,
  [[cat:beginning_of_sentence]]],
 [tell,
  [[cat:noun,root:tell,agr:'3SG',poss:'NONE',case:nom]]],
 ['brak\'taki',
  [[cat:adj,stem:[cat:noun,root:brak,agr:'3SG',poss:'NONE',type:proper,case:loc],
    suffix:rel],
   [cat:noun,stem:[cat:adj,stem:[cat:noun,root:brak,agr:'3SG',poss:'NONE',type:proper,
    case:loc],suffix:rel],suffix:none,agr:'3SG',poss:'NONE',case:nom]]],
 [mO,
  [[cat:adj,root:mO]]],
 ['.',
  [[cat:punct,root:'.']]],
 [#,
  [[cat:end_of_sentence]]]].

[[@,
  [[cat:beginning_of_sentence]]],
 ['4.',
  [[cat:adj,type:ordinal,root:'4.']]],
 [binyIl,
  [[cat:noun,root:binyIl,agr:'3SG',poss:'NONE',case:nom]]],
 [tapInaGInda,
  [[cat:noun,root:tapInak,agr:'3SG',poss:'2SG',case:loc],
   [cat:noun,root:tapInak,agr:'3SG',poss:'3SG',case:loc]]],
 ['300\'U',
```

```
 [[cat:noun,stem:[cat:adj,type:cardinal,root:'300\'''],suffix:none,
   agr:'3SG',poss:'NONE',case:acc],
  [cat:noun,stem:[cat:adj,type:cardinal,root:'300\'''],suffix:none,
   agr:'3SG',poss:'3SG',case:nom]]],
[aSkIn,
 [[cat:noun,root:aSK,agr:'3SG',poss:'NONE',case:gen],
  [cat:noun,root:aSK,agr:'3SG',poss:'2SG',case:nom],
  [cat:postp,root:aSkIn,subcat:acc]]],
['(',
 [[cat:punct,root:'(']]],
[ayrIca,
 [[cat:adverb,stem:[cat:adj,root:ayrI],suffix:ca,type:manner],
  [cat:noun,root:ayrIC,agr:'3SG',poss:'NONE',case:dat]]],
[parCa,
 [[cat:noun,root:parCa,agr:'3SG',poss:'NONE',case:nom]]],
[halinde,
 [[cat:noun,root:hVl,agr:'3SG',poss:'2SG',case:loc],
  [cat:noun,root:hVl,agr:'3SG',poss:'3SG',case:loc]]],
[binlerce,
 [[cat:adj,root:binlerce,type:cardinal],
  [cat:noun,stem:[cat:adj,root:bin,type:cardinal],suffix:none,
   agr:'3PL',poss:'NONE',case:equ]]],
[')',
 [[cat:punct,root:')']]],
[taS,
 [[cat:adj,root:taS],
  [cat:noun,root:taS,agr:'3SG',poss:'NONE',case:nom],
  [cat:verb,root:taS,sense:pos,tam1:imp,agr:'2SG']]],
['ya da',
 [[cat:conn,root:'ya da']]],
[piSmiS,
 [[cat:verb,root:piS,sense:pos,tam1:narr,agr:'3SG'],
  [cat:adj,stem:[cat:verb,root:piS,sense:pos,tam1:narr],suffix:none]]],
[kilden,
 [[cat:noun,root:kil,agr:'3SG',poss:'NONE',case:abl]]],
[yapIlmIS,
 [[cat:verb,root:yap,voice:(pass),sense:pos,tam1:narr,agr:'3SG'],
  [cat:adj,stem:[cat:verb,root:yap,voice:(pass),sense:pos,tam1:narr],suffix:none]]],
['"',
 [[cat:punct,root:'"']]],
[gOz,
 [[cat:noun,root:gOz,agr:'3SG',poss:'NONE',case:nom]]],
```

```
 [putu,
  [[cat:noun,root:put,agr:'3SG',poss:'NONE',case:acc],
   [cat:noun,root:put,agr:'3SG',poss:'3SG',case:nom]]],
 ['"',
  [[cat:punct,root:'"']]],
 [bulunmuStur,
  [[cat:verb,root:bulun,sense:pos,tam1:narr,copula:'2',agr:'3SG'],
   [cat:verb,root:bul,voice:(pass),sense:pos,tam1:narr,copula:'2',agr:'3SG']]],
 ['.',
  [[cat:punct,root:'.']]],
 [#,
  [[cat:end_of_sentence]]]].

[[@,
  [[cat:beginning_of_sentence]]],
 [tapInakta,
  [[cat:noun,root:tapInak,agr:'3SG',poss:'NONE',case:loc]]],
 [',',
  [[cat:punct,root:',']]],
 [yUkseklikleri,
  [[cat:noun,stem:[cat:adj,root:yUksek],suffix:lik,agr:'3PL',poss:'NONE',case:acc],
   [cat:noun,stem:[cat:adj,root:yUksek],suffix:lik,agr:'3PL',poss:'3SG',case:nom],
   [cat:noun,stem:[cat:adj,root:yUksek],suffix:lik,agr:'3PL',poss:'3PL',case:nom],
   [cat:noun,stem:[cat:adj,root:yUksek],suffix:lik,agr:'3SG',poss:'3PL',case:nom]]],
 ['2',
  [[cat:adj,type:cardinal,root:'2']]],
 [ile,
  [[cat:conn,root:ile],
   [cat:noun,root:il,agr:'3SG',poss:'NONE',case:dat],
   [cat:postp,root:ile,subcat:nom]]],
 ['11',
  [[cat:adj,type:cardinal,root:'11']]],
 [cm,
  [[cat:noun,root:cmE,type:rproper,agr:'3SG',poss:'NONE',case:nom]]],
 [arasInda,
  [[cat:noun,root:ara,agr:'3SG',poss:'3SG',case:loc]]],
 [deGiSen,
  [[cat:adj,stem:[cat:verb,root:deGiS,sense:pos],suffix:yan],
   [cat:adj,stem:[cat:verb,root:deG,voice:recip,sense:pos],suffix:yan]]],
 [bu,
  [[cat:adj,root:bu,type:determiner],
   [cat:pronoun,root:bu,type:demons,agr:'3SG',poss:'NONE',case:nom]]],
```

```
 [adak,
  [[cat:noun,root:adak,agr:'3SG',poss:'NONE',case:nom]]],
 [simgelerinden,
  [[cat:noun,root:simge,agr:'3PL',poss:'2SG',case:abl],
   [cat:noun,root:simge,agr:'3PL',poss:'3SG',case:abl],
   [cat:noun,root:simge,agr:'3PL',poss:'3PL',case:abl],
   [cat:noun,root:simge,agr:'3SG',poss:'3PL',case:abl]]],
 ['20,000 - 22,000',
  [[cat:adj,type:range,root:'20,000-22,000']]],
 [kadar,
  [[cat:postp,root:kadar,subcat:dat,type:temp2]]],
 [bulunduGu,
  [[cat:adj,stem:[cat:verb,root:bulun,sense:pos],suffix:dik,poss:'3SG'],
   [cat:noun,stem:[cat:verb,root:bulun,sense:pos],suffix:dik,agr:'3SG',poss:'3SG',
    case:nom],
   [cat:adj,stem:[cat:verb,root:bul,voice:(pass),sense:pos],suffix:dik,poss:'3SG'],
   [cat:noun,stem:[cat:verb,root:bul,voice:(pass),sense:pos],suffix:dik,
    agr:'3SG',poss:'3SG',case:nom]]],
 [hesaplanmIStIr,
  [[cat:verb,stem:[cat:noun,root:hesab],suffix:lan,sense:pos,tam1:narr,
    copula:'2',agr:'3SG'],
   [cat:verb,root:hesapla,voice:(pass),sense:pos,tam1:narr,copula:'2',agr:'3SG']]],
 ['.',
  [[cat:punct,root:'.']]],
 [#,
  [[cat:end_of_sentence]]]].

[[@,
  [[cat:beginning_of_sentence]]],
 [niceliksel,
  [[cat:adj,root:niceliksel],
   [cat:noun,stem:[cat:adj,root:niceliksel],suffix:none,agr:'3SG',poss:'NONE',
    case:nom]]],
 ['CalISmalar',
  [[cat:noun,stem:[cat:verb,root:'CalIS',sense:pos],suffix:ma,
    type:infinitive,agr:'3PL',poss:'NONE',case:nom],
   [cat:verb,stem:[cat:noun,stem:[cat:verb,root:'CalIS',sense:pos],suffix:ma,
    type:infinitive,agr:'3SG',poss:'NONE',case:nom],suffix:none,tam2:pres,agr:'3PL']]],
 [yapIlacaGI,
  [[cat:adj,stem:[cat:verb,root:yap,voice:(pass),sense:pos],suffix:yacak,poss:'3SG'],
   [cat:noun,stem:[cat:verb,root:yap,voice:(pass),sense:pos],suffix:yacak,
    agr:'3SG',poss:'3SG',case:nom]]],
```

```
[zaman,
 [[cat:noun,root:zaman,type:temp1,agr:'3SG',poss:'NONE',case:nom]]],
[',',
 [[cat:punct,root:',']]],
[temsili,
 [[cat:adj,root:temsili],
  [cat:noun,stem:[cat:adj,root:temsili],suffix:none,agr:'3SG',poss:'NONE',case:nom],
  [cat:noun,root:temsil,agr:'3SG',poss:'NONE',case:acc],
  [cat:noun,root:temsil,agr:'3SG',poss:'3SG',case:nom]]],
[nitelikte,
 [[cat:noun,root:nitelik,agr:'3SG',poss:'NONE',case:loc]]],
['Ornekler',
 [[cat:noun,root:'Ornek',agr:'3PL',poss:'NONE',case:nom],
  [cat:verb,stem:[cat:noun,root:'Ornek',agr:'3SG',poss:'NONE',case:nom],suffix:none,
   tam2:pres,agr:'3PL'],
  [cat:verb,root:'Ornekle',sense:pos,tam1:(aorist),agr:'3SG'],
  [cat:adj,stem:[cat:verb,root:'Ornekle',sense:pos,tam1:(aorist)],suffix:none]]],
['elde etmenin',
 [[cat:noun,stem:[cat:verb,root:'elde
ed',sense:pos],suffix:ma,type:infinitive,agr:'3SG',poss:'NONE',case:gen],
  [cat:noun,stem:[cat:verb,root:'elde ed',sense:pos],suffix:ma,
   type:infinitive,agr:'3SG',poss:'2SG',case:gen]]],
[temel,
 [[cat:adj,root:temel],
  [cat:noun,root:temel,agr:'3SG',poss:'NONE',case:nom]]],
[yOntemlerinden,
 [[cat:noun,root:yOntem,agr:'3PL',poss:'2SG',case:abl],
  [cat:noun,root:yOntem,agr:'3PL',poss:'3SG',case:abl],
  [cat:noun,root:yOntem,agr:'3PL',poss:'3PL',case:abl],
  [cat:noun,root:yOntem,agr:'3SG',poss:'3PL',case:abl]]],
[biri,
 [[cat:noun,stem:[cat:adj,root:bir,type:cardinal],suffix:none,agr:'3SG',poss:'NONE',
   case:acc],
  [cat:noun,stem:[cat:adj,root:bir,type:cardinal],suffix:none,agr:'3SG',poss:'3SG',
   case:nom],
  [cat:pronoun,root:biri,type:quant,agr:'3SG',poss:'3SG',case:nom]]],
[elemedir,
 [[cat:verb,stem:[cat:noun,stem:[cat:verb,root:ele,sense:pos],suffix:ma,type:infinitive,
   agr:'3SG',poss:'NONE',case:nom],suffix:none,tam2:pres,copula:'2',agr:'3SG']]],
['.',
 [[cat:punct,root:'.']]],
[#,
```

```
  [[cat:end_of_sentence]]]].

[[@,
  [[cat:beginning_of_sentence]]],
 [kumlu,
  [[cat:adj,stem:[cat:noun,root:kum],suffix:li],
   [cat:noun,stem:[cat:adj,stem:[cat:noun,root:kum],suffix:li],suffix:none,
    agr:'3SG',poss:'NONE',case:nom]]],
 [toprakta,
  [[cat:noun,root:toprak,agr:'3SG',poss:'NONE',case:loc]]],
 [kuru,
  [[cat:adj,root:kuru],
   [cat:noun,stem:[cat:adj,root:kuru],suffix:none,agr:'3SG',poss:'NONE',case:nom],
   [cat:noun,root:kur,agr:'3SG',poss:'NONE',case:acc],
   [cat:noun,root:kur,agr:'3SG',poss:'3SG',case:nom],
   [cat:verb,root:kuru,sense:pos,tam1:imp,agr:'2SG']]],
 [eleme,
  [[cat:noun,root:elem,agr:'3SG',poss:'NONE',case:dat],
   [cat:verb,root:ele,sense:neg,tam1:imp,agr:'2SG'],
   [cat:noun,stem:[cat:verb,root:ele,sense:pos],suffix:ma,
    type:infinitive,agr:'3SG',poss:'NONE',case:nom]]],
 ['zaman zaman',
  [[cat:adverb,root:'zaman zaman']]],
 [mUmkUndUr,
  [[cat:verb,stem:[cat:adj,root:mUmkUn],suffix:none,tam2:pres,copula:'2',agr:'3SG']]],
 ['.',
  [[cat:punct,root:'.']]],
 [#,
  [[cat:end_of_sentence]]]].
```

# Appendix D

# Hand-crafted Rules

## D.1 Contextual Choose Rules

```
% Choose CONN at the beginning of the sentence
[llc:[],lc:[[cat:'beginning_of_sentence']],rc:[],rrc:[],choose:[cat:conn]].

% After a CONN, choose the CONN reading of ``ki''
[llc:[],lc:[],rc:[[cat:conn]],rrc:[],choose:[cat:conn],token:ki].

% Rules for handling the word ``var''
[llc:[],lc:[],rc:[[cat:punct]],rrc:[],choose:[cat:adj],token:var].
[llc:[],lc:[],rc:[[cat:punct]],rrc:[],choose:[cat:verb,stem:[cat:adj]],token:vardIr].
[llc:[],lc:[[case:nom]],rc:[[cat:punct]],rrc:[],
 choose:[cat:verb,stem:[cat:adj]],token:vardI].
[llc:[[case:nom]],lc:[[cat:conn,root:de]],rc:[[cat:punct]],rrc:[],
 choose:[cat:verb,stem:[cat:adj]],token:vardI].
[llc:[[case:nom]],lc:[[cat:conn,root:da]],rc:[[cat:punct]],rrc:[],
 choose:[cat:verb,stem:[cat:adj]],token:vardI].
[llc:[],lc:[[case:nom]],rc:[],rrc:[],choose:[cat:verb,stem:[cat:adj]],token:varmIS].
[llc:[],lc:[[case:nom]],rc:[],rrc:[],choose:[cat:verb,stem:[cat:adj]],token:varsa].

% Before end-of-sentence choose verb
[llc:[],lc:[],rc:[[root: '.']],rrc:[],choose:[cat:verb]].
[llc:[],lc:[],rc:[[root: '?']],rrc:[],choose:[cat:verb]].
[llc:[],lc:[],rc:[[root:'!']],rrc:[],choose:[cat:verb]].
[llc:[],lc:[],rc:[[root: '...']],rrc:[],choose:[cat:verb]].
```

```
[llc:[],lc:[],rc:[[root: ':']],rrc:[],choose:[cat:verb]].

% Rules for handling the word ``ol''
[llc:[],lc:[[cat:adj]],rc:[],rrc:[],choose:[cat:adj],token:olan].
[llc:[],lc:[[cat:adj]],rc:[],rrc:[],choose:[cat:noun],token:olmak].
[llc:[],lc:[[cat:adj]],rc:[],rrc:[],choose:[cat:verb],token:olmuStur].
[llc:[],lc:[[cat:adj]],rc:[],rrc:[],choose:[cat:verb],token:oldu].
[llc:[],lc:[[cat:adj]],rc:[],rrc:[],choose:[cat:verb,root:ol]].
[llc:[],lc:[[cat:adj]],rc:[],rrc:[],choose:[cat:adj,stem:[cat:verb,root:ol],suffix:yan]].
[llc:[],lc:[[cat:adj]],rc:[],rrc:[],choose:[cat:noun,stem:[cat:verb,root:ol],suffix:ma]].
[llc:[],lc:[[cat:adj]],rc:[],rrc:[],
 choose:[cat:noun,case:abl,stem:[cat:verb,root:ol],suffix:dik]].
[llc:[],lc:[[cat:adj]],rc:[],rrc:[],
 choose:[cat:noun,case:acc,stem:[cat:verb,root:ol],suffix:dik]].
[llc:[],lc:[[cat:adj]],rc:[],rrc:[],
 choose:[cat:noun,case:abl,stem:[cat:verb,root:ol],suffix:yacak]].
[llc:[],lc:[[cat:adj]],rc:[],rrc:[],
 choose:[cat:noun,case:acc,stem:[cat:verb,root:ol],suffix:yacak]].
[llc:[],lc:[[cat:adj]],rc:[],rrc:[],choose:[cat:noun,stem:[cat:verb,root:ol],suffix:mak]].
[llc:[],lc:[[case:nom]],rc:[],rrc:[],choose:[cat:adj],token:olan].
[llc:[],lc:[[case:nom]],rc:[],rrc:[],choose:[cat:noun],token:olmak].
[llc:[],lc:[[case:nom]],rc:[],rrc:[],choose:[cat:verb],token:olmuStur].
[llc:[],lc:[[case:nom]],rc:[],rrc:[],choose:[cat:verb],token:oldu].
[llc:[],lc:[[case:nom]],rc:[],rrc:[],choose:[cat:verb,root:ol]].
[llc:[],lc:[[case:nom]],rc:[],rrc:[],choose:[cat:adj,stem:[cat:verb,root:ol],suffix:yan]].
[llc:[],lc:[[case:nom]],rc:[],rrc:[],choose:[cat:noun,stem:[cat:verb,root:ol],suffix:ma]].
[llc:[],lc:[[case:nom]],rc:[],rrc:[],
 choose:[cat:noun,stem:[cat:verb,root:ol],suffix:mak]].

% Rules for numeric tokens
[llc:[[cat:adj,type:cardinal]],lc:[[cat:adj,type:cardinal]],rc:[[cat:adj,type:cardinal]],
 rrc:[],choose:[cat:adj,type:cardinal]].
[llc:[[cat:adj,type:cardinal]],lc:[[cat:adj,type:cardinal]],
 rc:[[cat:noun,agr:'3SG',poss:'NONE',stem:no]],rrc:[],choose:[cat:adj,type:cardinal]].
[llc:[], lc:[[cat:noun,stem:[cat:adj,type:cardinal],case:loc,poss:'NONE']],rc:[],rrc:[],
 choose:[cat:noun,stem:[cat:adj,type:cardinal]]].

% Rules for handling tokens related to date
[llc:[],lc:[[cat:date]],rc:[], rrc:[], choose:[cat:noun,case:nom,agr:'3SG'],token:gUnU].
[llc:[],lc:[[cat:date]],rc:[], rrc:[], choose:[cat:adj],token:gUnlU].
[llc:[],lc:[[type:cardinal]],rc:[],rrc:[],choose:[cat:noun,poss:'3SG'],token:yIlI].
[llc:[],lc:[[type:cardinal]],rc:[],rrc:[],
```

```
 choose:[cat:noun,poss:'3SG',case:loc],token:yIlInda].
[llc:[],lc:[[type:cardinal]],rc:[],rrc:[],
 choose:[cat:noun,poss:'3SG',case:abl],token:yIlIndan].
[llc:[],lc:[[type:cardinal]],rc:[],rrc:[],
 choose:[cat:noun,poss:'3SG',case:gen],token:yIlInIn].

% Choose the adjectival reading to the verbal reading, if the next
% token is a verb
[llc:[],lc:[[cat:adj,stem:[cat:verb,tam1:narr]]],rc:[],rrc:[],
 choose:[cat:verb,tam1:pres]].
[llc:[],lc:[[cat:adj,stem:[cat:verb,tam1:narr]]],rc:[],rrc:[],
 choose:[cat:verb,tam1:past]].
[llc:[],lc:[[cat:adj,stem:[cat:verb,tam1:narr]]],rc:[],rrc:[],
 choose:[cat:verb,tam1:future]].
[llc:[],lc:[[cat:adj,stem:[cat:verb,tam1:narr]]],rc:[],rrc:[],
 choose:[cat:verb,tam1:narr]].
[llc:[],lc:[[cat:adj,stem:[cat:verb,tam1:narr]]],rc:[],rrc:[],
 choose:[cat:verb,tam1:neces]].
[llc:[],lc:[[cat:adj,stem:[cat:verb,tam1:narr]]],rc:[],rrc:[],
 choose:[suffix:dik]].

% Rules for omitting the very infrequent reading for:
% For example ``ama'' is also a noun, and ``de'' is a verb
[llc:[],lc:[],rc:[],rrc:[],choose:[cat:conn,root:ama]].
[llc:[],lc:[],rc:[],rrc:[],choose:[cat:conn,root:de]].
[llc:[],lc:[],rc:[],rrc:[],choose:[cat:conn,root:ya]].
[llc:[],lc:[],rc:[],rrc:[],choose:[cat:conn,root:hatta]].
[llc:[],lc:[],rc:[],rrc:[],choose:[cat:adverb],token:aslInda].
[llc:[],lc:[],rc:[],rrc:[],choose:[cat:adverb],token:yalnIz].
[llc:[],lc:[],rc:[],rrc:[],choose:[cat:postp,root:diye]].
[llc:[],lc:[],rc:[],rrc:[],choose:[cat:adverb],token:hala].
[llc:[],lc:[],rc:[],rrc:[],choose:[cat:adverb],token:iSte].
[llc:[],lc:[],rc:[],rrc:[],choose:[cat:verb],token:ise].
[llc:[],lc:[],rc:[],rrc:[],choose:[cat:pronoun,case:acc],token:bunu].
[llc:[],lc:[],rc:[],rrc:[],choose:[cat:pronoun],token:bunlar].
[llc:[],lc:[],rc:[],rrc:[],choose:[cat:pronoun],token:bunun].
[llc:[],lc:[],rc:[],rrc:[],choose:[cat:pronoun],token:onlar].
[llc:[],lc:[],rc:[],rrc:[],choose:[cat:pronoun],token:bizler].
[llc:[],lc:[],rc:[],rrc:[],choose:[cat:pronoun],token:bizi].
[llc:[],lc:[],rc:[],rrc:[],choose:[cat:pronoun],token:biz].
[llc:[],lc:[],rc:[],rrc:[],choose:[cat:pronoun,case:acc],token:onlarI].
[llc:[],lc:[],rc:[],rrc:[],choose:[cat:pronoun,case:acc],token:bunlarI].
```

```
[llc:[],lc:[],rc:[],rrc:[],choose:[cat:pronoun,case:acc],token:birini].
[llc:[],lc:[],rc:[],rrc:[],choose:[cat:pronoun,case:acc],token:beni].
[llc:[],lc:[],rc:[],rrc:[],choose:[cat:pronoun,case:acc],token:onu].
[llc:[],lc:[],rc:[],rrc:[],choose:[cat:pronoun,case:acc],token:'Sunu'].
[llc:[],lc:[],rc:[],rrc:[],choose:[cat:pronoun,agr:'3PL',poss:'NONE',case:nom]].

% A noun phrase, but the modified or specified word is a pronoun
% e.g. adamlardan bazIlarI
[llc:[],lc:[[agr:'3PL',case:abl]],rc:[],rrc:[],choose:[cat:pronoun,poss:'3PL']].
[llc:[],lc:[[agr:'3PL',case:abl]],rc:[],rrc:[],choose:[cat:pronoun,poss:'3SG']].
[llc:[],lc:[[agr:'3PL',case:abl]],rc:[],rrc:[],choose:[cat:pronoun,poss:'3PL']].
[llc:[],lc:[[agr:'3PL',case:gen]],rc:[],rrc:[],choose:[cat:pronoun,poss:'3SG']].
[llc:[[agr:'3PL',case:abl]],lc:[[cat:adj]],rc:[],rrc:[],choose:[cat:pronoun,poss:'3SG']].
[llc:[[agr:'3PL',case:gen]],lc:[[cat:adj]],rc:[],rrc:[],choose:[cat:pronoun,poss:'3SG']].

% Noun phrase combinations
[llc:[],lc:[[agr:'3PL',case:abl]],rc:[],rrc:[],choose:[cat:noun,poss:'3SG']].
[llc:[[cat:adj,type:determiner]],lc:[[cat:adj,stem:[cat:noun]]],rc:[[cat:noun,poss:'NONE']],
 rrc:[],choose:[cat:adj]].
[llc:[[cat:adj]],lc:[[cat:adj,stem:[cat:noun],suffix:rel]],rc:[[cat:noun,poss:'NONE']],
 rrc:[],choose:[cat:adj]].
[llc:[[cat:adj]],lc:[[cat:adj,stem:[cat:noun],suffix:rel]],rc:[[cat:noun,poss:'3SG']],
 rrc:[],choose:[cat:noun,agr:'3SG',case:nom,stem:no]].
[llc:[[cat:adj]],lc:[[cat:adj,stem:[cat:noun],suffix:li]],rc:[[cat:noun,poss:'NONE']],
 rrc:[],choose:[cat:adj]].
[llc:[[cat:adj]],lc:[[cat:adj,stem:[cat:noun],suffix:li]],rc:[[cat:noun,poss:'3SG']],
 rrc:[],choose:[cat:noun,agr:'3SG',case:nom,stem:no]].
[llc:[[cat:adj]],lc:[[cat:adj,stem:[cat:noun],suffix:lik]],rc:[[cat:noun,poss:'3SG']],
 rrc:[],choose:[cat:noun,agr:'3SG',case:nom,stem:no]].
[llc:[[cat:noun,agr:'3SG',poss:'NONE',case:nom,stem:no]],
 lc:[[cat:adj,stem:[cat:noun,stem:no],suffix:li]],rc:[[cat:noun,poss:'NONE']],
 rrc:[],choose:[cat:adj]].

% Rules for the problem word ``ile''
% Still not working correctly
[llc:[[cat:noun,agr:'3SG',poss:'NONE',case:nom]],lc:[[cat:conn,root:ile]],rc:[],rrc:[],
 choose:[cat:noun,agr:'3SG',poss:'NONE']].
[llc:[[cat:noun,agr:'3SG',poss:'NONE',case:nom]],lc:[[cat:conn,root:ile]],rc:[],rrc:[],
 choose:[cat:noun,agr:'3PL',poss:'NONE']].
[llc:[[cat:noun,agr:'3SG',poss:'NONE',case:nom]],lc:[[cat:conn,root:ile]],rc:[],rrc:[],
 choose:[cat:noun,agr:'3SG',poss:'3SG']].
[llc:[[cat:noun,agr:'3SG',poss:'NONE',case:nom]],lc:[[cat:conn,root:ile]],rc:[],rrc:[],
```

```
 choose:[cat:noun,agr:'3PL',poss:'3PL']].

% Choose the pronoun reading, instead of any other
[llc:[],lc:[],rc:[],rrc:[],choose:[cat:pronoun,case:dat]].
[llc:[],lc:[],rc:[],rrc:[],choose:[cat:pronoun,case:abl]].
[llc:[],lc:[],rc:[],rrc:[],choose:[cat:pronoun,case:ins]].
[llc:[],lc:[],rc:[],rrc:[],choose:[cat:pronoun,case:loc]].
[llc:[],lc:[],rc:[],rrc:[],choose:[cat:pronoun,case:gen]].

% NOUN+CASE - POSTP+SUBCAT agreements
[llc:[],lc:[[suffix:madan]],rc:[],rrc:[],choose:[cat:postp,subcat:abl]].
[llc:[],lc:[],rc:[[cat:postp,root:gibi]],rrc:[],choose:[cat:pronoun,case:gen]].
[llc:[],lc:[],rc:[[cat:postp,root:kadar]],rrc:[],choose:[cat:pronoun,case:gen]].
[llc:[],lc:[],rc:[[cat:postp,root:iCin]],rrc:[],choose:[cat:pronoun,case:gen]].
[llc:[],lc:[],rc:[[cat:verb,stem:[cat:postp,root:iCin]]],rrc:[],
 choose:[cat:pronoun,case:gen]].
[llc:[],lc:[],rc:[[cat:postp,subcat:dat]],rrc:[],choose:[case:dat]].
[llc:[],lc:[],rc:[[cat:postp,subcat:ins]],rrc:[],choose:[case:ins]].
[llc:[],lc:[],rc:[[cat:postp,subcat:gen]],rrc:[],choose:[case:gen]].
[llc:[],lc:[],rc:[[cat:postp,subcat:loc]],rrc:[],choose:[case:loc]].
[llc:[],lc:[],rc:[[cat:postp,root:iCin]],rrc:[],choose:[case:nom]].
[llc:[],lc:[],rc:[[cat:postp,root:gibi]],rrc:[],choose:[case:nom]].
[llc:[],lc:[],rc:[[cat:postp,root:boyunca]],rrc:[],choose:[case:nom]].
[llc:[],lc:[],rc:[[cat:postp,root:'Uzere']],rrc:[],choose:[case:nom]].
[llc:[],lc:[],rc:[[cat:postp,root:diye]],rrc:[],choose:[case:nom]].
[llc:[],lc:[],rc:[[cat:postp,subcat:acc]],rrc:[],choose:[stem:[type:cardinal],case:acc]].
[llc:[],lc:[],rc:[[cat:postp,subcat:acc]],rrc:[],choose:[case:acc]].
[llc:[],lc:[],rc:[[cat:postp,subcat:abl]],rrc:[],choose:[case:abl]].

% Some POSTPs do not require CASE agreement, especially for tokens
% indicating date
[llc:[],lc:[],rc:[[cat:postp,root:sonra,subcat:abl]],rrc:[],
 choose:[case:nom],token:yIl].
[llc:[],lc:[],rc:[[cat:postp,root:sonra,subcat:abl]],rrc:[],
 choose:[case:nom],token:gUn].
[llc:[],lc:[],rc:[[cat:postp,root:sonra,subcat:abl]],rrc:[],
 choose:[case:nom],token:ay].
[llc:[],lc:[],rc:[[cat:postp,root:sonra,subcat:abl]],rrc:[],
 choose:[case:nom],token:saat].
[llc:[],lc:[],rc:[[cat:postp,root:sonra,subcat:abl]],rrc:[],
 choose:[case:nom],token:hafta].
[llc:[],lc:[],rc:[[cat:postp,root:sonra,subcat:abl]],rrc:[],
```

```
 choose:[case:nom],token:sUre].
[llc:[],lc:[],rc:[[cat:postp,root:'Once',subcat:abl]],rrc:[],
 choose:[case:nom],token:yIl].
[llc:[],lc:[],rc:[[cat:postp,root:'Once',subcat:abl]],rrc:[],
 choose:[case:nom],token:gUn].
[llc:[],lc:[],rc:[[cat:postp,root:'Once',subcat:abl]],rrc:[],
 choose:[case:nom],token:ay].
[llc:[],lc:[],rc:[[cat:postp,root:'Once',subcat:abl]],rrc:[],
 choose:[case:nom],token:saat].
[llc:[],lc:[],rc:[[cat:postp,root:'Once',subcat:abl]],rrc:[],
 choose:[case:nom],token:hafta].
[llc:[],lc:[],rc:[[cat:postp,root:'Once',subcat:abl]],rrc:[],
 choose:[case:nom],token:sUre].
[llc:[],lc:[],rc:[[cat:verb,stem:[cat:postp,root:sonra,subcat:abl]]],rrc:[],
 choose:[case:nom],token:yIl].
[llc:[],lc:[],rc:[[cat:verb,stem:[cat:postp,root:sonra,subcat:abl]]],rrc:[],
 choose:[case:nom],token:gUn].
[llc:[],lc:[],rc:[[cat:verb,stem:[cat:postp,root:sonra,subcat:abl]]],rrc:[],
 choose:[case:nom],token:ay].
[llc:[],lc:[],rc:[[cat:verb,stem:[cat:postp,root:sonra,subcat:abl]]],rrc:[],
 choose:[case:nom],token:saat].
[llc:[],lc:[],rc:[[cat:verb,stem:[cat:postp,root:sonra,subcat:abl]]],rrc:[],
 choose:[case:nom],token:hafta].
[llc:[],lc:[],rc:[[cat:verb,stem:[cat:postp,root:sonra,subcat:abl]]],rrc:[],
 choose:[case:nom],token:sUre].
[llc:[],lc:[],rc:[[cat:verb,stem:[cat:postp,root:'Once',subcat:abl]]],rrc:[],
 choose:[case:nom],token:yIl].
[llc:[],lc:[],rc:[[cat:verb,stem:[cat:postp,root:'Once',subcat:abl]]],rrc:[],
 choose:[case:nom],token:gUn].
[llc:[],lc:[],rc:[[cat:verb,stem:[cat:postp,root:'Once',subcat:abl]]],rrc:[],
 choose:[case:nom],token:ay].
[llc:[],lc:[],rc:[[cat:verb,stem:[cat:postp,root:'Once',subcat:abl]]],rrc:[],
 choose:[case:nom],token:saat].
[llc:[],lc:[],rc:[[cat:verb,stem:[cat:postp,root:'Once',subcat:abl]]],rrc:[],
 choose:[case:nom],token:hafta].
[llc:[],lc:[],rc:[[cat:verb,stem:[cat:postp,root:'Once',subcat:abl]]],rrc:[],
 choose:[case:nom],token:sUre].

% Rules for the word ``Cok''
[llc:[[case:abl,suffix:dik]],lc:[[cat:adverb]],rc:[[cat:noun,stem:no]],rrc:[],
 choose:[cat:adj],token:'Cok'].
[llc:[],lc:[],rc:[[cat:adj,stem:[cat:noun]]],rrc:[],choose:[cat:adj],token:'Cok'].
```

```
[llc:[],lc:[],rc:[[cat:verb,stem:[cat:adj,stem:[cat:noun]]]],rrc:[],
 choose:[cat:adj],token:'Cok'].
[llc:[],lc:[],rc:[[cat:adj,stem:no]],rrc:[],choose:[cat:adverb],token:'Cok'].
[llc:[],lc:[],rc:[[cat:verb,stem:[cat:adj]]],rrc:[],choose:[cat:adverb],token:'Cok'].
[llc:[],lc:[[case:abl]],rc:[[cat:postp,subcat:abl]],rrc:[],
 choose:[cat:adverb],token:daha].
[llc:[],lc:[[case:abl]],rc:[[cat:postp,subcat:abl]],rrc:[],
 choose:[cat:adverb],token:'Cok'].
[llc:[],lc:[[case:abl]],rc:[[cat:postp,subcat:abl]],rrc:[],
 choose:[cat:adverb],token:daha].
[llc:[],lc:[[case:abl]],rc:[[cat:postp,subcat:abl]],rrc:[],
 choose:[cat:adverb],token:'Cok'].
[llc:[],lc:[[cat:noun,stem:no,case:abl]],rc:[],rrc:[],choose:[cat:postp],token:'Cok'].
[llc:[],lc:[[cat:noun,stem:[cat:adj],case:abl]],rc:[],rrc:[],
 choose:[cat:postp],token:'Cok'].
[llc:[],lc:[],rc:[[cat:noun,stem:no]],rrc:[],choose:[cat:adj],token:'Cok'].
[llc:[],lc:[[cat:adverb]],rc:[[cat:noun,stem:no]],rrc:[],choose:[cat:adj],token:'Cok'].
[llc:[],lc:[[cat:noun,suffix:mak,case:abl]],rc:[[cat:adj]],rrc:[],
 choose:[cat:adverb],token:'Cok'].
[llc:[],lc:[[cat:noun,suffix:mak,case:abl]],rc:[[cat:verb,stem:[cat:adj]]],rrc:[],
 choose:[cat:adverb],token:'Cok'].
[llc:[],lc:[],rc:[[cat:adverb]],rrc:[],choose:[cat:adverb],token:'Cok'].
[llc:[],lc:[],rc:[[cat:postp,root:gibi]],rrc:[],choose:[cat:verb]].
[llc:[[cat:adverb,stem:no]],lc:[[cat:adj,stem:no]],rc:[[cat:noun]],rrc:[],
 choose:[cat:adj]].

% VERB - CASE agreements
% Some verbs require certain case marked nouns before them,
% e.g. eve yOnelmek
[llc:[],lc:[[case:dat]],rc:[],rrc:[],
 choose:[cat:noun,stem:[cat:verb,root:yapIS],suffix:ma]].
[llc:[],lc:[[case:dat]],rc:[],rrc:[],choose:[cat:verb,root:yapIS]].
[llc:[],lc:[[case:dat]],rc:[],rrc:[],choose:[stem:[cat:verb,root:yapIS]]].
[llc:[],lc:[[case:dat]],rc:[],rrc:[],
 choose:[cat:noun,stem:[cat:verb,root:yOnel],suffix:ma]].
[llc:[],lc:[[case:dat]],rc:[],rrc:[],choose:[cat:verb,root:yOnel]].
[llc:[],lc:[[case:dat]],rc:[],rrc:[],choose:[stem:[cat:verb,root:yOnel]]].
[llc:[],lc:[[case:acc]],rc:[],rrc:[],
 choose:[cat:noun,stem:[cat:verb,root:tarihle],suffix:ma]].
[llc:[],lc:[[case:acc]],rc:[],rrc:[],choose:[cat:verb,root:tarihle]].
[llc:[],lc:[[case:acc]],rc:[],rrc:[],choose:[stem:[cat:verb,root:tarihle]]].
[llc:[],lc:[[case:nom]],rc:[],rrc:[],
```

```
 choose:[cat:noun,stem:[cat:verb,root:tarihle],suffix:ma]].
[llc:[],lc:[[case:nom]],rc:[],rrc:[],choose:[cat:verb,root:tarihle]].
[llc:[],lc:[[case:nom]],rc:[],rrc:[],choose:[stem:[cat:verb,root:tarihle]]].

% ADJ - CASE agreements
% NOUN - CASE agreements
% Some adjectives and nouns behave as POSTPs, and require certain case
% markings before them
% e.g. bana layIk, benim tarafImdan
[llc:[],lc:[[case:dat]],rc:[],rrc:[],choose:[cat:adj],token:yapISIk].
[llc:[],lc:[[case:abl]],rc:[],rrc:[],choose:[cat:adj],token:farklI].
[llc:[],lc:[[case:abl]],rc:[],rrc:[],choose:[cat:adj],token:sorumlu].
[llc:[[case:abl]],lc:[[cat:adverb]],rc:[],rrc:[],choose:[cat:adj],token:farklI].
[llc:[],lc:[[case:gen]],rc:[],rrc:[],
 choose:[cat:noun,agr:'3SG',poss:'3SG',case:dat],token:yerine].
[llc:[],lc:[[case:nom]],rc:[],rrc:[],
 choose:[cat:noun,agr:'3SG',poss:'3SG',case:dat],token:yerine].
[llc:[],lc:[[case:dat]],rc:[],rrc:[],choose:[cat:adj],token:yakIn].
[llc:[],lc:[[case:dat]],rc:[],rrc:[],choose:[cat:adj],token:layIk].
[llc:[],lc:[[case:dat]],rc:[],rrc:[],choose:[cat:adj],token:aCIk].
[llc:[],lc:[[case:nom,stem:[cat:verb]]],rc:[],rrc:[],choose:[cat:adj],token:halde].
[llc:[],lc:[[case:nom]],rc:[],rrc:[],
 choose:[cat:noun,agr:'3SG',poss:'3SG',case:abl],token:tarafIndan].

%  ADVERB - ADJ - NOUN forms, for the tokens ``en'' and ``daha''
[llc:[],lc:[],rc:[[cat:adj]],rrc:[[cat:noun,stem:no]],choose:[cat:adverb],token:en].
[llc:[],lc:[],rc:[[cat:adj]],rrc:[[cat:noun,suffix:lik]],choose:[cat:adverb],token:en].
[llc:[],lc:[],rc:[[cat:adj]],rrc:[[cat:noun,suffix:ma]],choose:[cat:adverb],token:en].
[llc:[],lc:[],rc:[[cat:adj]],rrc:[[cat:noun,stem:no]],choose:[cat:adverb],token:daha].
[llc:[],lc:[],rc:[[cat:adj]],rrc:[[cat:noun,suffix:lik]],choose:[cat:adverb],token:daha].
[llc:[],lc:[],rc:[[cat:adj]],rrc:[[cat:noun,suffix:ma]],choose:[cat:adverb],token:daha].
[llc:[],lc:[],rc:[[cat:adj]],rrc:[[cat:verb,stem:[cat:noun,suffix:ma]]],
 choose:[cat:adverb],token:en].
[llc:[],lc:[],rc:[[cat:adverb]],rrc:[[cat:noun,stem:[cat:verb]]],
 choose:[cat:adverb],token:en].
[llc:[],lc:[],rc:[[cat:adverb]],rrc:[],choose:[cat:adverb],token:en].
[llc:[],lc:[],rc:[[cat:adj]],rrc:[],choose:[cat:adverb],token:en].
[llc:[],lc:[],rc:[[stem:[cat:adj]]],rrc:[],choose:[cat:adverb],token:en].

% Noun phrases with one or more adjectives before a noun
% Note that instead of simple rules, we used very controlled rules to
% define the nouns and the adjectives
```

```
[llc:[[cat:adj,stem:no]],lc:[[cat:adj,type:determiner]],rc:[[cat:noun,stem:no]],rrc:[],
 choose:[cat:adj,stem:no]].
[llc:[],lc:[],rc:[[cat:adj,stem:no]],rrc:[[cat:noun,case:dat,poss:'NONE',stem:no]],
 choose:[cat:adj,type:determiner]].
[llc:[],lc:[],rc:[[cat:adj,stem:no]],rrc:[[cat:noun,case:gen,poss:'NONE',stem:no]],
 choose:[cat:adj,type:determiner]].
[llc:[],lc:[],rc:[[cat:adj,stem:no]],rrc:[[cat:noun,case:abl,poss:'NONE',stem:no]],
 choose:[cat:adj,type:determiner]].
[llc:[],lc:[],rc:[[cat:adj,stem:no]],rrc:[[cat:noun,case:ins,poss:'NONE',stem:no]],
 choose:[cat:adj,type:determiner]].
[llc:[],lc:[],rc:[[cat:adj,stem:no]],rrc:[[cat:noun,case:loc,poss:'NONE',stem:no]],
 choose:[cat:adj,type:determiner]].
[llc:[],lc:[],rc:[[cat:adj,stem:no]],rrc:[[cat:noun,case:nom,stem:no]],
 choose:[cat:adj,type:determiner]].
[llc:[],lc:[],rc:[[cat:adj,type:determiner]],rrc:[[cat:noun,stem:no]],
 choose:[cat:adj,stem:no]].
[llc:[],lc:[],rc:[[cat:adj,stem:no]],rrc:[[cat:noun,suffix:ma]],
 choose:[cat:adj,type:determiner]].
[llc:[],lc:[],rc:[[cat:adj,stem:no]],rrc:[[cat:noun,suffix:yis]],
 choose:[cat:adj,type:determiner]].
[llc:[],lc:[],rc:[[cat:adj,stem:no]],rrc:[[cat:noun,suffix:lik]],
 choose:[cat:adj,type:determiner]].
[llc:[],lc:[],rc:[[cat:adj,type:determiner]],rrc:[[cat:noun,stem:no]],
 choose:[cat:adj,suffix:li]].
[llc:[],lc:[],rc:[[cat:adj,type:determiner]],rrc:[[cat:noun,stem:no]],
choose:[cat:adj,suffix:ik]].
[llc:[],lc:[],rc:[[cat:adj,type:determiner]],rrc:[[cat:noun,stem:no]],
 choose:[cat:adj,suffix:lik]].
[llc:[],lc:[[cat:adverb]],rc:[[cat:adj,type:determiner]],rrc:[[cat:noun,stem:no]],
 choose:[cat:adj,suffix:yan]].
[llc:[],lc:[],rc:[[cat:adj,type:determiner]],rrc:[[cat:noun,stem:no]],
 choose:[cat:adj,suffix:yan]].
[llc:[],lc:[],rc:[[cat:adj,type:determiner]],rrc:[[cat:noun,stem:no]],
 choose:[cat:adj,suffix:rel]].
[llc:[],lc:[],rc:[[cat:adj,type:determiner]],rrc:[[cat:noun,stem:no]],
 choose:[cat:adj,stem:[cat:verb,tam1:narr]]].
[llc:[],lc:[[cat:adj,type:determiner]],rc:[],rrc:[],choose:[cat:noun,case:gen,stem:no]].
[llc:[],lc:[[cat:adj,type:determiner]],rc:[],rrc:[],choose:[cat:noun,case:loc,stem:no]].
[llc:[],lc:[[cat:adj,type:determiner]],rc:[],rrc:[],choose:[cat:noun,case:abl,stem:no]].
[llc:[],lc:[[cat:adj,type:determiner]],rc:[],rrc:[],choose:[cat:noun,case:ins,stem:no]].
[llc:[],lc:[[cat:adj,type:determiner]],rc:[],rrc:[],choose:[cat:adj,stem:[cat:noun]]].
[llc:[],lc:[[cat:adj,type:determiner]],rc:[],rrc:[],choose:[cat:noun,stem:no]].
```

```
[llc:[],lc:[[cat:adj,type:determiner]],rc:[],rrc:[],choose:[cat:noun,suffix:ma]].
[llc:[],lc:[[cat:adj,type:determiner]],rc:[],rrc:[],choose:[cat:noun,suffix:yis]].

% A noun phrase, but the modified or specified word is a pronoun
% e.g. evlerin bazIlarI
[llc:[],lc:[[agr:'3PL',case:gen]],rc:[],rrc:[],choose:[cat:pronoun,poss:'3PL']].

% Noun phrases in the form: NOUN+GEN-ADJ-NOUN+POSS
% Also noun phrases, which have verbal reading
% e.g. benim mavi kitabIm
[llc:[[agr:'1SG',case:gen]],lc:[[cat:adj,stem:no]],rc:[],rrc:[],
 choose:[cat:noun,poss:'1SG',stem:no]].
[llc:[[agr:'1SG',case:gen]],lc:[[cat:adj,stem:no]],rc:[],rrc:[],
 choose:[cat:verb,stem:[cat:noun,poss:'1SG',stem:no]]].
[llc:[[agr:'2SG',case:gen]],lc:[[cat:adj,stem:no]],rc:[],rrc:[],
 choose:[cat:noun,poss:'2SG',stem:no]].
[llc:[[agr:'2SG',case:gen]],lc:[[cat:adj,stem:no]],rc:[],rrc:[],
 choose:[cat:verb,stem:[cat:noun,poss:'2SG',stem:no]]].
[llc:[[agr:'3PL',case:gen]],lc:[[cat:adj,stem:no]],rc:[],rrc:[],
 choose:[cat:noun,agr:'3PL',poss:'3PL',stem:no]].
[llc:[[agr:'3PL',case:gen]],lc:[[cat:adj,stem:no]],rc:[],rrc:[],
 choose:[cat:noun,agr:'3PL',poss:'3PL',stem:no]].
[llc:[[agr:'3PL',case:gen]],lc:[[cat:adj,stem:no]],rc:[],rrc:[],
 choose:[cat:noun,agr:'3PL',poss:'3PL',suffix:ma]].
[llc:[[agr:'3PL',case:gen]],lc:[[cat:adj,stem:no]],rc:[],rrc:[],
 choose:[cat:verb,stem:[cat:noun,agr:'3PL',poss:'3PL',stem:no]]].
[llc:[[agr:'3PL',case:gen]],lc:[[cat:adj,stem:no]],rc:[],rrc:[],
 choose:[cat:verb,stem:[cat:noun,poss:'3PL',stem:no]]].
[llc:[[agr:'3PL',case:gen]],lc:[[cat:adj,stem:no]],rc:[],rrc:[],
 choose:[cat:verb,stem:[cat:noun,poss:'3SG',stem:no]]].
[llc:[[agr:'3PL',case:gen]],lc:[[cat:adj,stem:no]],rc:[],rrc:[],
 choose:[cat:noun,agr:'3SG',poss:'3PL',stem:no]].
[llc:[[agr:'3PL',case:gen]],lc:[[cat:adj,stem:no]],rc:[],rrc:[],
 choose:[cat:verb,stem:[cat:noun,agr:'3SG',poss:'3PL',stem:no]]].
[llc:[[agr:'3PL',case:gen]],lc:[[cat:adj,stem:no]],rc:[],rrc:[],
 choose:[cat:noun,poss:'3PL',stem:no]].
[llc:[[agr:'3PL',case:gen]],lc:[[cat:adj,stem:no]],rc:[],rrc:[],
 choose:[cat:verb,stem:[cat:noun,poss:'3PL',stem:no]]].
[llc:[[agr:'3PL',case:gen]],lc:[[cat:adj,stem:no]],rc:[],rrc:[],
 choose:[cat:noun,poss:'3SG',stem:no]].
[llc:[[agr:'3SG',case:gen]],lc:[[cat:adverb]],rc:[[cat:noun,poss:'3SG']],rrc:[],
 choose:[cat:adj]].
```

```
[llc:[[agr:'3SG',case:gen]],lc:[[cat:adj,stem:no]],rc:[],rrc:[],
 choose:[cat:noun,poss:'3SG',stem:no]].
[llc:[[agr:'3SG',case:gen]],lc:[[cat:adj,stem:no]],rc:[],rrc:[],
 choose:[cat:verb,stem:[cat:noun,poss:'3SG',stem:no]]].
[llc:[[agr:'1PL',case:gen]],lc:[[cat:adj,stem:no]],rc:[],rrc:[],
 choose:[cat:noun,poss:'1PL',stem:no]].
[llc:[[agr:'2PL',case:gen]],lc:[[cat:adj,stem:no]],rc:[],rrc:[],
 choose:[cat:noun,poss:'2PL',stem:no]].

% Noun phrases in the form: NOUN+GEN-NOUN-NOUN+POSS
% Also noun phrases, which have verbal reading
% e.g. benim kitap kapaGIm
[llc:[],lc:[[agr:'1SG',case:gen]],rc:[[cat:noun,poss:'1SG']],rrc:[],
 choose:[cat:noun,case:nom,stem:no]].
[llc:[],lc:[[agr:'2SG',case:gen]],rc:[[cat:noun,poss:'2SG']],rrc:[],
 choose:[cat:noun,case:nom,stem:no]].
[llc:[],lc:[[agr:'3PL',case:gen]],rc:[[cat:noun,agr:'3PL',poss:'3PL']],rrc:[],
 choose:[cat:noun,case:nom,poss:'NONE',stem:no]].
[llc:[],lc:[[agr:'3PL',case:gen]],rc:[[cat:noun,poss:'3PL']],rrc:[],
 choose:[cat:noun,poss:'NONE',case:nom,stem:no]].
[llc:[],lc:[[agr:'3SG',case:gen]],rc:[[cat:noun,agr:'3PL',poss:'3SG']],rrc:[],
 choose:[cat:noun,agr:'3SG',poss:'NONE',case:nom,stem:no]].
[llc:[],lc:[[agr:'1PL',case:gen]],rc:[[cat:noun,poss:'1PL']],rrc:[],
 choose:[cat:noun,case:nom,stem:no]].
[llc:[],lc:[[agr:'2PL',case:gen]],rc:[[cat:noun,poss:'2PL']],rrc:[],
 choose:[cat:noun,case:nom,stem:no]].
[llc:[],lc:[[agr:'1SG',case:gen]],rc:[],rrc:[],choose:[cat:noun,poss:'1SG',stem:no]].
[llc:[],lc:[[agr:'2SG',case:gen]],rc:[],rrc:[],choose:[cat:noun,poss:'2SG',stem:no]].
[llc:[],lc:[[agr:'3PL',case:gen]],rc:[],rrc:[],
 choose:[cat:noun,agr:'3PL',poss:'3PL',stem:no]].
[llc:[],lc:[[agr:'3PL',case:gen]],rc:[],rrc:[],choose:[cat:noun,poss:'3PL',stem:no]].
[llc:[],lc:[[agr:'3PL',poss:'NONE',case:gen]],rc:[],rrc:[],
 choose:[cat:noun,poss:'3SG',stem:no]].
[llc:[],lc:[[agr:'3PL',poss:'3SG',case:gen]],rc:[],rrc:[],
 choose:[cat:noun,poss:'3SG',stem:no]].
[llc:[],lc:[[agr:'3PL',case:gen]],rc:[],rrc:[],choose:[cat:noun,poss:'3PL',suffix:ma]].
[llc:[],lc:[[agr:'3PL',case:gen]],rc:[],rrc:[],choose:[cat:noun,poss:'3SG',suffix:ma]].
[llc:[],lc:[[agr:'3SG',case:gen]],rc:[],rrc:[],choose:[cat:noun,poss:'3SG',suffix:ma]].
[llc:[],lc:[[agr:'3SG',case:gen]],rc:[],rrc:[],choose:[cat:noun,poss:'3SG',stem:no]].
[llc:[],lc:[[agr:'3SG',case:gen]],rc:[],rrc:[],
 choose:[cat:verb,stem:[cat:noun,poss:'3SG',stem:no]]].
[llc:[],lc:[[agr:'1PL',case:gen]],rc:[],rrc:[],choose:[cat:noun,poss:'1PL',stem:no]].
```

```
[llc:[],lc:[[agr:'2PL',case:gen]],rc:[],rrc:[],choose:[cat:noun,poss:'2PL',stem:no]].

% Some more rules for the token ``ile''
[llc:[[cat:adj,stem:no]],lc:[[cat:conn,root:ile]],rc:[],rrc:[],choose:[cat:adj,stem:no]].
[llc:[],lc:[],rc:[[root:birlikte]],rrc:[],choose:[cat:postp,root:ile]].
[llc:[],lc:[],rc:[[cat:verb,stem:no]],rrc:[],choose:[cat:postp,root:ile]].

% Some more noun phrase combinations
[llc:[[cat:adj,stem:[cat:verb,tam1:narr]]],lc:[[cat:adj,stem:no]],rc:[],rrc:[],
 choose:[cat:noun, poss:'NONE',stem:no]].
[llc:[[cat:adj,stem:[cat:verb],suffix:dik]],lc:[[cat:adj,stem:no]],rc:[],rrc:[],
 choose:[cat:noun,poss:'NONE',stem:no]].
[llc:[],lc:[[cat:adj,stem:[cat:verb],suffix:dik]],rc:[],rrc:[],
 choose:[cat:noun, poss:'NONE',case:abl,stem:no]].
[llc:[],lc:[[cat:adj,stem:[cat:verb],suffix:dik]],rc:[],rrc:[],
 choose:[cat:noun, poss:'NONE',case:nom,stem:no]].
[llc:[],lc:[[cat:adj,stem:[cat:verb],suffix:dik]],rc:[],rrc:[],
 choose:[cat:noun, poss:'NONE',case:loc,stem:no]].
[llc:[],lc:[[cat:adj,stem:[cat:verb],suffix:dik]],rc:[],rrc:[],
 choose:[cat:noun, poss:'NONE',case:ins,stem:no]].
[llc:[],lc:[[cat:adj,stem:[cat:verb,tam1:narr]]],rc:[],rrc:[],
 choose:[cat:noun, poss:'NONE',stem:no]].
[llc:[],lc:[[cat:adj,stem:[cat:verb],suffix:yacak]],rc:[],rrc:[],
 choose:[cat:noun, poss:'NONE',stem:no]].
[llc:[],lc:[[cat:noun,agr:'3SG',poss:'NONE',case:nom,stem:no]],
 rc:[[cat:noun,agr:'3SG',poss:'NONE',case:nom,stem:no]],
 rrc:[[cat:noun,poss:'3SG',agr:'3SG',case:nom,stem:no]],
 choose:[cat:noun,case:nom,agr:'3SG',poss:'3SG',stem:no]].
[llc:[],lc:[[cat:adj,stem:no]],rc:[],rrc:[],
 choose:[cat:noun,stem:no,poss:'NONE',case:gen]].
[llc:[],lc:[[cat:adj,stem:no]],rc:[],rrc:[],
 choose:[cat:noun,poss:'NONE',stem:no,case:abl]].
[llc:[],lc:[[cat:adj,stem:no]],rc:[],rrc:[],
 choose:[cat:noun,poss:'NONE',stem:no,case:loc]].
[llc:[],lc:[[cat:adj,stem:no]],rc:[],rrc:[],
 choose:[cat:noun,poss:'NONE',stem:no,case:ins]].
[llc:[],lc:[[cat:adj,stem:no]],rc:[],rrc:[],
 choose:[cat:noun,stem:no,poss:'NONE',case:nom]].
[llc:[[cat:adj,type:cardinal]],lc:[[cat:adj,type:cardinal]],rc:[],rrc:[],
 choose:[cat:noun,agr:'3SG',poss:'NONE',stem:no]].
[llc:[],lc:[[cat:adj,type:cardinal]],rc:[],rrc:[],
 choose:[cat:noun,agr:'3SG',poss:'NONE',stem:no]].
```

```
[llc:[],lc:[],rc:[[cat:adj,stem:no]],rrc:[[cat:noun,stem:no]],
 choose:[cat:adj,type:cardinal]].
[llc:[],lc:[[cat:adj,stem:no]],rc:[[cat:noun,poss:'3SG']],rrc:[],
 choose:[cat:noun,agr:'3SG',poss:'NONE',case:nom, stem:no]].
[llc:[],lc:[[cat:noun,stem:[cat:verb],suffix:mak,case:nom]],rc:[],rrc:[],
 choose:[cat:verb,agr:'3SG',stem:[cat:adj,stem:no]]].
[llc:[],lc:[[cat:noun,stem:[cat:verb],case:nom]],rc:[],rrc:[],
 choose:[cat:verb,stem:[cat:noun,poss:'NONE',suffix:mak,case:loc]]].

% Rules for the connectives ``ve'' and ``veya''
[llc:[[cat:adj,stem:no]],lc:[[cat:conn,root:ve]],rc:[],rrc:[],choose:[cat:adj,stem:no]].
[llc:[[cat:adj,stem:no]],lc:[[cat:conn,root:veya]],rc:[],rrc:[],choose:[cat:adj,stem:no]].

% Noun phrases, with no genitive case marking
[llc:[[cat:noun,agr:'3SG',case:nom,stem:no]],
 lc:[[cat:noun,agr:'3SG',poss:'3SG',case:nom,stem:no]],rc:[],rrc:[],
 choose:[cat:noun,agr:'3PL',poss:'3SG',stem:no]].
[llc:[[cat:noun,agr:'3SG',case:nomm,stem:no]],
 lc:[[cat:noun,agr:'3SG',poss:'3SG',case:nom,stem:no]],rc:[],rrc:[],
 choose:[cat:noun,agr:'3SG',poss:'3SG',stem:no]].
[llc:[],lc:[[cat:noun,agr:'3PL',poss:'NONE',case:nom,stem:no]],rc:[],rrc:[],
 choose:[cat:noun,agr:'3SG',poss:'3SG',stem:no]].
[llc:[],lc:[[cat:noun,agr:'3SG',poss:'NONE',case:nom,stem:no]],rc:[],rrc:[],
 choose:[cat:noun,stem:no,agr:'3PL',poss:'3SG']].
[llc:[],lc:[[cat:noun,agr:'3SG',poss:'NONE',case:nom,stem:no]],rc:[],rrc:[],
 choose:[cat:verb,stem:[cat:noun,agr:'3PL',poss:'3SG']]].
[llc:[],lc:[[cat:noun,agr:'3SG',poss:'NONE',case:nom,stem:no]],rc:[],rrc:[],
 choose:[cat:adj,stem:[cat:noun,agr:'3PL',poss:'3SG']]].
[llc:[],lc:[[cat:noun,agr:'3SG',poss:'NONE',case:nom,stem:no]],rc:[],rrc:[],
 choose:[cat:noun,agr:'3SG',poss:'3SG']].

% At the beginning of a sentence, choose adverbial reading
[llc:[],lc:[[cat:'beginning_of_sentence']],rc:[],rrc:[],choose:[cat:adverb]].
```

## D.2 Lexical Delete Rules

```
[delete:[poss:'2SG']].
[delete:[cat:verb,voice:reflex]].
```

```
[delete:[poss:'3PL',agr:'3SG']].
[delete:[tam1:opt,agr:'3SG']].
[delete:[tam1:imp]].
[delete:[case:equ]].
[delete:[poss:'1SG']].
[delete:[cat:verb,stem:[cat:postp]]].
[delete:[cat:verb,stem:[cat:noun],suffix:lan]].
[delete:[cat:verb,stem:[cat:noun],suffix:las]].
[delete:[cat:verb,stem:[cat:adj],suffix:lan]].
[delete:[cat:verb,stem:[cat:adj],suffix:las]].
[delete:[cat:postp,root:sonra]].
[delete:[cat:postp,root:'Once']].
[delete:[cat:postp,root:karSI]].
[delete:[cat:postp,root:birlikte]].
[delete:[cat:postp,root:doGru]].
[delete:[cat:postp,root:baSka]].
[delete:[token:ayrIca,cat:noun]].
[delete:[stem:[stem]]].
[delete:[cat:noun,stem:[cat:adj],suffix:none]].
[delete:[token:'hem',cat:adverb]].
[delete:[token:'bir',type:cardinal]].
[delete:[token:'bir',cat:adverb]].
[delete:[token:'ben',cat:noun]].
[delete:[token:gene,cat:noun]].
[delete:[type:real]].
[delete:[token:'bugUn',cat:noun]].
[delete:[token:'gerCekte',cat:noun]].
[delete:[token:'Cok',cat:adj]].
[delete:[token:'artIk',cat:adj]].
[delete:[token:'sonradan',cat:noun]].
[delete:[token:'geceleri',cat:noun]].
[delete:[token:'toptan',cat:noun]].
[delete:[token:'biz',cat:noun]].
[delete:[token:'ah',cat:noun]].
[delete:[token:'ancak',cat:conn]].
[delete:[token:'Oyle',cat:adj]].
[delete:[token:'akla',case:ins]].
[delete:[token:'asla',case:ins]].
[delete:[token:'oysa',cat:verb]].
[delete:[cat:verb,stem:[cat:noun,poss:'2SG']]].
[delete:[root:'VT!']].
```

# Appendix E

# Learned Rules

## E.1 Learned Choose Rules

Following choose rules are among the choose rules learned from C2000.

```
[llc:[[cat:adj]],lc:[[cat:adj,type:determiner]],rc:[],rrc:[],
 choose:[cat:noun,agr:'3SG',poss:'NONE']].
[llc:[],lc:[[cat:noun,agr:'3SG',case:nom]],rc:[],rrc:[],
 choose:[cat:noun,agr:'3SG',poss:'3SG']].
[llc:[],lc:[[cat:adj,type:determiner]],rc:[],rrc:[],
 choose:[cat:noun,agr:'3SG',poss:'NONE']].
[llc:[],lc:[],rc:[[cat:noun,agr:'3SG',poss:'3SG']],rrc:[],
 choose:[cat:noun,agr:'3SG',case:nom]].
[llc:[[cat:beginning_of_sentence]],lc:[[cat:adj,type:determiner]],rc:[],rrc:[],
 choose:[cat:noun,agr:'3SG',poss:'NONE']].
[llc:[[cat:noun,agr:'3SG',case:nom]],lc:[[cat:conn,root:ve]],rc:[],rrc:[],
 choose:[cat:noun,agr:'3SG',poss:'NONE']].
[llc:[],lc:[],rc:[[cat:conn,root:ve]],rrc:[[cat:noun,agr:'3SG',poss:'NONE']],
 choose:[cat:noun,agr:'3SG',case:nom]].
[llc:[[cat:adj]],lc:[[cat:noun,agr:'3SG',case:nom]],rc:[],rrc:[],
 choose:[cat:noun,agr:'3SG',poss:'3SG']].
[llc:[],lc:[[cat:adj]],rc:[],rrc:[],choose:[cat:noun,agr:'3SG',poss:'NONE']].
[llc:[[cat:adj,type:determiner]],lc:[[cat:noun,agr:'3SG',case:nom]],rc:[],rrc:[],
 choose:[cat:noun,agr:'3SG',poss:'3SG']].
[llc:[[cat:adj,type:determiner]],lc:[[cat:adj]],rc:[],rrc:[],
 choose:[cat:noun,agr:'3SG',poss:'NONE']].
```

```
[llc:[[cat:noun,agr:'3SG',case:nom]],lc:[[cat:noun,agr:'3SG',case:nom]],rc:[],rrc:[],
 choose:[cat:noun,agr:'3SG',poss:'3SG']].
[llc:[],lc:[],rc:[[cat:noun,agr:'3SG',poss:'3SG']],
 rrc:[[cat:noun,agr:'3SG',poss:'NONE']],
 choose:[cat:noun,agr:'3SG',case:nom]].
[llc:[],lc:[],rc:[[cat:noun,agr:'3SG',poss:'3SG']],
 rrc:[[cat:noun,agr:'3SG',poss:'3SG']],
 choose:[cat:noun,agr:'3SG',case:nom]].
[llc:[],lc:[],rc:[[cat:adj]],rrc:[[cat:noun,agr:'3SG',poss:'NONE']],
 choose:[cat:adverb]].
[llc:[],lc:[],rc:[[cat:noun,agr:'3SG',poss:'3SG']],rrc:[[cat:adj]],
 choose:[cat:noun,agr:'3SG',case:nom]].
[llc:[],lc:[[cat:noun,agr:'3SG',case:nom]],rc:[],rrc:[],
 choose:[cat:noun,agr:'3PL',poss:'3SG']].
[llc:[],lc:[],rc:[[cat:noun,agr:'3PL',poss:'3SG']],rrc:[],
 choose:[cat:noun,agr:'3SG',case:nom]].
[llc:[],lc:[],rc:[[cat:noun,agr:'3SG',poss:'3SG']],rrc:[[cat:adj,type:determiner]],
 choose:[cat:noun,agr:'3SG',case:nom]].
[llc:[],lc:[],rc:[[cat:adj,type:determiner]],rrc:[[cat:noun,agr:'3SG',poss:'NONE']],
 choose:[cat:adj,suffix:yan]].
[llc:[[cat:beginning_of_sentence]],lc:[[cat:adj]],rc:[],rrc:[],
 choose:[cat:noun,agr:'3SG',poss:'NONE']].
[llc:[],lc:[[cat:adj]],rc:[],rrc:[],choose:[cat:adj,type:determiner]].
[llc:[],lc:[[cat:adj]],rc:[],rrc:[],choose:[cat:noun,agr:'3PL',poss:'NONE']].
[llc:[[cat:adj,suffix:li]],lc:[[cat:adj,type:determiner]],rc:[],rrc:[],
 choose:[cat:noun,agr:'3SG',poss:'NONE']].
[llc:[[cat:noun,agr:'3SG',case:nom]],lc:[[cat:noun,agr:'3SG',case:nom]],rc:[],rrc:[],
 choose:[cat:noun,agr:'3SG',poss:'NONE',type:rproper]].
[llc:[],lc:[],rc:[[cat:adj,type:determiner]],rrc:[[cat:noun,agr:'3SG',poss:'NONE']],
 choose:[cat:adj,suffix:li]].
[llc:[[cat:noun,agr:'3SG',case:loc]],lc:[[cat:adj,type:determiner]],rc:[],rrc:[],
 choose:[cat:noun,agr:'3SG',poss:'NONE']].
[llc:[],lc:[],rc:[[cat:noun,agr:'3SG',poss:'3SG']],rrc:[],
 choose:[cat:noun,agr:'3SG',case:gen]].
[llc:[],lc:[],rc:[[cat:adj]],rrc:[],choose:[cat:adverb]].
[llc:[[cat:noun,agr:'3SG',case:nom,type:rproper]],lc:[[cat:conn,root:ve]],rc:[],rrc:[],
 choose:[cat:noun,agr:'3SG',poss:'NONE',type:rproper]].
```

## E.2 Learned Delete Rules

Following delete rules are among the delete rules learned from C2000 text.

```
[llc:[],lc:[],rc:[[cat:verb]],rrc:[],
 delete:[cat:verb]].
[llc:[],lc:[],rc:[[cat:verb]],rrc:[],
 delete:[cat:noun,agr:'3PL',poss:'3SG',case:ins]].
[llc:[],lc:[],rc:[[cat:verb]],rrc:[],
 delete:[cat:noun,agr:'3PL',poss:'3PL',case:loc]].
[llc:[],lc:[],rc:[[cat:verb]],rrc:[],
 delete:[cat:noun,agr:'3PL',poss:'3PL',case:acc]].
[llc:[],lc:[],rc:[[cat:verb]],rrc:[],
 delete:[cat:noun,agr:'3PL',poss:'3PL',case:abl]].
[llc:[],lc:[],rc:[[cat:verb]],rrc:[],
 delete:[cat:conn,root:ile]].
[llc:[],lc:[],rc:[[cat:pronoun,agr:'3SG',poss:'3SG',case:nom]],rrc:[],
 delete:[cat:noun,agr:'3PL',poss:'3PL',case:abl]].
[llc:[],lc:[],rc:[[cat:postp,subcat:nom]],rrc:[],
 delete:[cat:noun,agr:'3SG',poss:'NONE',case:acc]].
[llc:[],lc:[],rc:[[cat:postp,subcat:nom]],rrc:[],
 delete:[cat:noun,agr:'3PL',poss:'3PL',case:nom]].
[llc:[],lc:[],rc:[[cat:postp,subcat:ins]],rrc:[],
 delete:[cat:noun,agr:'3PL',poss:'3PL',case:ins]].
[llc:[],lc:[],rc:[[cat:postp,subcat:abl]],rrc:[],
 delete:[cat:noun,agr:'3PL',poss:'3PL',case:abl]].
[llc:[],lc:[],rc:[[cat:noun,agr:'3SG',poss:'NONE',case:nom]],rrc:[],
 delete:[cat:verb]].
[llc:[],lc:[],rc:[[cat:adverb]],rrc:[],
 delete:[cat:pronoun,agr:'3SG',poss:'NONE',case:nom]].
[llc:[],lc:[],rc:[[cat:adverb]],rrc:[],
 delete:[cat:pronoun,agr:'3PL',poss:'3PL',case:nom]].
[llc:[],lc:[],rc:[[cat:adverb]],rrc:[],
 delete:[cat:noun,agr:'3PL',poss:'NONE',case:acc]].
[llc:[],lc:[],rc:[[cat:adverb]],rrc:[],
 delete:[cat:noun,agr:'3PL',poss:'3SG',case:abl]].
[llc:[],lc:[],rc:[[cat:adverb]],rrc:[],
 delete:[cat:noun,agr:'3PL',poss:'3PL',case:nom]].
[llc:[],lc:[],rc:[[cat:adverb]],rrc:[],
 delete:[cat:noun,agr:'3PL',poss:'3PL',case:loc]].
```

```
[llc:[],lc:[],rc:[[cat:adverb]],rrc:[],
 delete:[cat:noun,agr:'3PL',poss:'3PL',case:dat]].
[llc:[],lc:[],rc:[[cat:adverb]],rrc:[],
 delete:[cat:noun,agr:'3PL',poss:'3PL',case:acc]].
[llc:[],lc:[],rc:[[cat:adj]],rrc:[],
 delete:[cat:verb]].
[llc:[],lc:[[cat:postp,subcat:nom]],rc:[],rrc:[],
 delete:[cat:pronoun,agr:'3SG',poss:'NONE',case:nom]].
[llc:[],lc:[[cat:postp,subcat:nom]],rc:[],rrc:[],
 delete:[cat:postp,subcat:abl]].
[llc:[],lc:[[cat:postp,subcat:nom]],rc:[],rrc:[],
 delete:[cat:noun,agr:'3PL',poss:'NONE',case:acc]].
[llc:[],lc:[[cat:postp,subcat:nom]],rc:[],rrc:[],
 delete:[cat:noun,agr:'3PL',poss:'3SG',case:nom]].
[llc:[],lc:[[cat:postp,subcat:nom]],rc:[],rrc:[],
 delete:[cat:noun,agr:'3PL',poss:'3PL',case:nom]].
[llc:[],lc:[[cat:postp,subcat:nom]],rc:[],rrc:[],
 delete:[cat:conn,root:ne]].
[llc:[],lc:[[cat:adverb]],rc:[],rrc:[],
 delete:[cat:pronoun,agr:'3SG',poss:'NONE',case:nom]].
[llc:[],lc:[[cat:adverb]],rc:[],rrc:[],
 delete:[cat:pronoun,agr:'3PL',poss:'3PL',case:acc]].
[llc:[],lc:[[cat:adverb]],rc:[],rrc:[],
 delete:[cat:noun,agr:'3PL',poss:'NONE',case:acc]].
[llc:[],lc:[[cat:adverb]],rc:[],rrc:[],
 delete:[cat:noun,agr:'3PL',poss:'3SG',case:nom]].
[llc:[],lc:[[cat:adverb]],rc:[],rrc:[],
 delete:[cat:noun,agr:'3PL',poss:'3SG',case:dat]].
[llc:[],lc:[[cat:adverb]],rc:[],rrc:[],
 delete:[cat:noun,agr:'3PL',poss:'3PL',case:nom]].
[llc:[],lc:[[cat:adverb]],rc:[],rrc:[],
 delete:[cat:noun,agr:'3PL',poss:'3PL',case:dat]].
[llc:[],lc:[[cat:adj]],rc:[],rrc:[],
 delete:[cat:pronoun,agr:'3SG',poss:'NONE',case:nom]].
[llc:[],lc:[[cat:adj]],rc:[],rrc:[],
 delete:[cat:pronoun,agr:'3SG',poss:'3SG',case:nom]].
[llc:[],lc:[[cat:adj]],rc:[],rrc:[],
 delete:[cat:noun,agr:'3PL',poss:'3SG',case:nom]].
[llc:[],lc:[[cat:adj]],rc:[],rrc:[],
 delete:[cat:noun,agr:'3PL',poss:'3SG',case:ins]].
[llc:[],lc:[[cat:adj]],rc:[],rrc:[],
 delete:[cat:noun,agr:'3PL',poss:'3PL',case:abl]].
```

# Appendix F

# Sample Disambiguated Text

```
[[@,
  [[cat:beginning_of_sentence]]],
 [arkeologlar,
  [[cat:noun,stem:[cat:noun,root:arkeoloji],suffix:og,agr:'3PL',poss:'NONE',case:nom]]],
 [',',
  [[cat:punct,root:',']]],
 [kazI,
  [[cat:noun,root:kazI,agr:'3SG',poss:'NONE',case:nom]]],
 [yapmanIn,
  [[cat:noun,stem:[cat:verb,root:yap,sense:pos],suffix:ma,
    type:infinitive,agr:'3SG',poss:'NONE',case:gen]]],
 ['yanI sIra',
  [[cat:postp,root:'yanI sIra',subcat:gen]]],
 [',',
  [[cat:punct,root:',']]],
 [o,
  [[cat:adj,root:o,type:determiner]]],
 [kazI,
  [[cat:noun,root:kazI,agr:'3SG',poss:'NONE',case:nom]]],
 [yerini,
  [[cat:noun,root:yer,agr:'3SG',poss:'3SG',case:acc]]],
 ['Cevreleyen',
  [[cat:adj,stem:[cat:verb,root:'Cevrele',sense:pos],suffix:yan]]],
 [alanIn,
  [[cat:noun,root:alan,agr:'3SG',poss:'NONE',case:gen]]],
 [eski,
```

```
  [[cat:adj,root:eski]]],
 [biCimini,
  [[cat:noun,root:biCim,agr:'3SG',poss:'3SG',case:acc]]],
 [de,
  [[cat:conn,root:de]]],
 [yeniden,
  [[cat:adverb,root:yeniden]]],
 [kurmaya,
  [[cat:noun,root:kurmay,agr:'3SG',poss:'NONE',case:dat]]],
 ['CalISIrlar',
  [[cat:verb,root:'CalIS',sense:pos,tam1:aorist,agr:'3PL']]],
 ['.',
  [[cat:punct,root:'.']]],
 [#,
  [[cat:end_of_sentence]]]].

[[@,
  [[cat:beginning_of_sentence]]],
 [ilk,
  [[cat:adj,root:ilK]]],
 [evler,
  [[cat:noun,root:ev,agr:'3PL',poss:'NONE',case:nom]]],
 [',',
  [[cat:punct,root:',']]],
 [burada,
  [[cat:noun,root:bura,agr:'3SG',poss:'NONE',case:loc]]],
 [kuzey,
  [[cat:adj,root:kuzey]]],
 ['Irak\'ta',
  [[cat:noun,root:'Irak',type:rproper,agr:'3SG',poss:'NONE',type:proper,case:loc]]],
 ['kermezdere\'deki',
  [[cat:adj,stem:[cat:noun,root:kermezdere,agr:'3SG',poss:'NONE',type:proper,case:loc],
    suffix:rel],
   [cat:adj,stem:[cat:noun,root:kermezDere,agr:'3SG',poss:'NONE',type:proper,case:loc],
    suffix:rel]]],
 [proto,
  [[cat:noun,root:proto,agr:'3SG',poss:'NONE',case:nom]]],
 [neolitik,
  [[cat:adj,root:neolitik]]],
 [dOnem,
  [[cat:noun,root:dOnem,agr:'3SG',poss:'NONE',case:nom]]],
 [evinde,
```

```
  [[cat:noun,root:ev,agr:'3SG',poss:'3SG',case:loc]]],
 [gOrUldUGU,
  [[cat:noun,stem:[cat:verb,root:gOr,voice:pass,sense:pos],suffix:dik,
    agr:'3SG',poss:'3SG',case:nom]]],
 [gibi,
  [[cat:postp,root:gibi,subcat:nom]]],
 [',',
  [[cat:punct,root:',']]],
 [topraGa,
  [[cat:noun,root:toprak,agr:'3SG',poss:'NONE',case:dat]]],
 [gOmUlU,
  [[cat:adj,stem:[cat:noun,root:gOmU],suffix:li]]],
 [yuvarlak,
  [[cat:adj,root:yuvarlak]]],
 [kulUbelerdi,
  [[cat:verb,stem:[cat:noun,root:kulUbe,agr:'3PL',poss:'NONE',case:nom],suffix:none,
    tam2:past,agr:'3SG']]],
 ['.',
  [[cat:punct,root:'.']]],
 [#,
  [[cat:end_of_sentence]]]].

[[@,
  [[cat:beginning_of_sentence]]],
 [taS,
  [[cat:adj,root:taS]]],
 [ve,
  [[cat:conn,root:ve]]],
 [tahta,
  [[cat:adj,root:tahta]]],
 [aletler,
  [[cat:noun,root:alet,agr:'3PL',poss:'NONE',case:nom]]],
 [arasInda,
  [[cat:noun,root:ara,agr:'3SG',poss:'3SG',case:loc]]],
 [',',
  [[cat:punct,root:',']]],
 [iGne,
  [[cat:noun,root:iGne,agr:'3SG',poss:'NONE',case:nom]]],
 [',',
  [[cat:punct,root:',']]],
 [dikiS,
  [[cat:noun,stem:[cat:verb,root:diK,sense:pos],suffix:yis,agr:'3SG',poss:'NONE',case:nom]]],
```

```
 [iGnesi,
  [[cat:noun,root:iGne,agr:'3SG',poss:'3SG',case:nom]]],
 [',',
  [[cat:punct,root:',']]],
 [bIz,
  [[cat:noun,root:bIz,agr:'3SG',poss:'NONE',case:nom]]],
 [',',
  [[cat:punct,root:',']]],
 [ok,
  [[cat:noun,root:oK,agr:'3SG',poss:'NONE',case:nom]]],
 [baSI,
  [[cat:noun,root:baS,agr:'3SG',poss:'3SG',case:nom]]],
 [',',
  [[cat:punct,root:',']]],
 [mIzrak,
  [[cat:noun,root:mIzrak,agr:'3SG',poss:'NONE',case:nom]]],
 ['ya da',
  [[cat:conn,root:'ya da']]],
 [zIpkIn,
  [[cat:noun,root:zIpkIn,agr:'3SG',poss:'NONE',case:nom]]],
 [uClarI,
  [[cat:noun,root:uC,agr:'3PL',poss:'3SG',case:nom]]],
 [',',
  [[cat:punct,root:',']]],
 ['vb.',
  [[cat:noun,root:'vb.E',type:rproper,agr:'3SG',poss:'NONE',case:nom]]],
 [',',
  [[cat:punct,root:',']]],
 [bulunuyordu,
  [[cat:verb,root:bulun,sense:pos,tam1:prog1,tam2:past,agr:'3SG']]],
 ['.',
  [[cat:punct,root:'.']]],
 [#,
  [[cat:end_of_sentence]]]].

[[@,
  [[cat:beginning_of_sentence]]],
 [proto,
  [[cat:noun,root:proto,agr:'3SG',poss:'NONE',case:nom]]],
 [neolitik,
  [[cat:adj,root:neolitik]]],
 ['eriha\'nIn',
```

```
 [[cat:noun,root:eriha,type:rproper,agr:'3SG',poss:'NONE',type:proper,case:gen]]],
[en,
 [[cat:adverb,root:en]]],
['dikkat Cekici',
 [[cat:adj,stem:[cat:verb,root:'dikkat
CeK',sense:pos],suffix:yici]]],
['Ozelliklerinden',
 [[cat:noun,root:'Ozellik',agr:'3PL',poss:'3SG',case:abl]]],
[biri,
 [[cat:pronoun,root:biri,type:quant,agr:'3SG',poss:'3SG',case:nom]]],
[',',
 [[cat:punct,root:',']]],
[surlarIn,
 [[cat:noun,root:sur,agr:'3PL',poss:'NONE',case:gen]]],
[iC,
 [[cat:adj,root:iJ]]],
[tarafIna,
 [[cat:noun,root:taraf,agr:'3SG',poss:'3SG',case:dat]]],
[yapISIk,
 [[cat:adj,root:yapISIk]]],
[taS,
 [[cat:noun,root:taS,agr:'3SG',poss:'NONE',case:nom]]],
[kuleydi,
 [[cat:verb,stem:[cat:noun,root:kule,agr:'3SG',poss:'NONE',case:nom],suffix:none,
   tam2:past,agr:'3SG']]],
['.',
 [[cat:punct,root:'.']]],
[#,
 [[cat:end_of_sentence]]]].

[[@,
 [[cat:beginning_of_sentence]]],
['10',
 [[cat:adj,type:cardinal,root:'10']]],
['m.',
 [[cat:noun,root:'m.E',type:rproper,agr:'3SG',poss:'NONE',case:nom]]],
['CapIndaki',
 [[cat:adj,stem:[cat:noun,root:'Cap',agr:'3SG',poss:'3SG',case:loc],suffix:rel]]],
[kulenin,
 [[cat:noun,root:kule,agr:'3SG',poss:'NONE',case:gen]]],
['8',
 [[cat:adj,type:cardinal,root:'8']]],
```

```
 [metrenin,
  [[cat:noun,root:metre,agr:'3SG',poss:'NONE',case:gen]]],
 ['UstUnde',
  [[cat:noun,root:'Ust',agr:'3SG',poss:'3SG',case:loc]]],
 [bir,
  [[cat:adj,root:bir,type:determiner]]],
 [bOlUmU,
  [[cat:noun,root:bOlUm,agr:'3SG',poss:'NONE',case:acc]]],
 [bugUn,
  [[cat:adverb,root:bugUn]]],
 [de,
  [[cat:conn,root:de]]],
 [ayaktadIr,
  [[cat:verb,stem:[cat:noun,root:ayak,agr:'3SG',poss:'NONE',case:loc],suffix:none,
    tam2:pres,copula:'2',agr:'3SG']]],
 ['.',
  [[cat:punct,root:'.']]],
 [#,
  [[cat:end_of_sentence]]]].

[[@,
  [[cat:beginning_of_sentence]]],
 [doGu,
  [[cat:noun,root:doGu,agr:'3SG',poss:'NONE',case:nom]]],
 [tarafInda,
  [[cat:noun,root:taraf,agr:'3SG',poss:'3SG',case:loc]]],
 ['1.7',
  [[cat:adj,type:real,root:'1.7']]],
 ['m
 ',
  [[cat:noun,root:mE,type:rproper,agr:'3SG',poss:'NONE',case:nom]]],
 [yUksekliGinde,
  [[cat:noun,stem:[cat:adj,root:yUksek],suffix:lik,agr:'3SG',poss:'3SG',case:loc]]],
 [bir,
  [[cat:adj,root:bir,type:determiner]]],
 [kapI,
  [[cat:noun,root:kapI,agr:'3SG',poss:'NONE',case:nom]]],
 [',',
  [[cat:punct,root:',']]],
 [her,
  [[cat:adj,root:her,type:determiner]]],
 [biri,
```

```
  [[cat:pronoun,root:biri,type:quant,agr:'3SG',poss:'3SG',case:nom]]],
 [tek,
  [[cat:adj,root:teK]]],
 [bir,
  [[cat:adj,root:bir,type:determiner]]],
 [taS,
  [[cat:adj,root:taS]]],
 [bloGundan,
  [[cat:noun,root:blok,agr:'3SG',poss:'3SG',case:abl]]],
 [yapIlmIS,
  [[cat:adj,stem:[cat:verb,root:yap,voice:pass,sense:pos,tam1:narr],suffix:none]]],
 ['22',
  [[cat:adj,type:cardinal,root:'22']]],
 [basamaklI,
  [[cat:adj,stem:[cat:noun,root:basamak],suffix:li]]],
 [bir,
  [[cat:adj,root:bir,type:determiner]]],
 [merdivene,
  [[cat:noun,root:merdiven,agr:'3SG',poss:'NONE',case:dat]]],
 [aCIlIr,
  [[cat:verb,root:aJ,voice:pass,sense:pos,tam1:aorist,agr:'3SG']]],
 ['.',
  [[cat:punct,root:'.']]],
 [#,
  [[cat:end_of_sentence]]]].

[[@,
  [[cat:beginning_of_sentence]]],
 [tell,
  [[cat:noun,root:tell,agr:'3SG',poss:'NONE',case:nom]]],
 ['brak\'taki',
  [[cat:adj,stem:[cat:noun,root:brak,agr:'3SG',poss:'NONE',type:proper,case:loc],
    suffix:rel]]],
 [mO,
  [[cat:adj,root:mO]]],
 ['.',
  [[cat:punct,root:'.']]],
 [#,
  [[cat:end_of_sentence]]]].

[[@,
  [[cat:beginning_of_sentence]]],
```

```
['4.',
 [[cat:adj,type:ordinal,root:'4.']]],
[binyIl,
 [[cat:noun,root:binyIl,agr:'3SG',poss:'NONE',case:nom]]],
[tapInaGInda,
 [[cat:noun,root:tapInak,agr:'3SG',poss:'3SG',case:loc]]],
['300\'U',
 [[cat:noun,stem:[cat:adj,type:cardinal,root:'300\''],suffix:none,
   agr:'3SG',poss:'NONE',case:acc]]],
[aSkIn,
 [[cat:postp,root:aSkIn,subcat:acc]]],
['(',
 [[cat:punct,root:'(']]],
[ayrIca,
 [[cat:adverb,stem:[cat:adj,root:ayrI],suffix:ca,type:manner]]],
[parCa,
 [[cat:noun,root:parCa,agr:'3SG',poss:'NONE',case:nom]]],
[halinde,
 [[cat:noun,root:hVl,agr:'3SG',poss:'3SG',case:loc]]],
[binlerce,
 [[cat:adj,root:binlerce,type:cardinal]]],
[')',
 [[cat:punct,root:')']]],
[taS,
 [[cat:noun,root:taS,agr:'3SG',poss:'NONE',case:nom]]],
['ya da',
 [[cat:conn,root:'ya da']]],
[piSmiS,
 [[cat:adj,stem:[cat:verb,root:piS,sense:pos,tam1:narr],suffix:none]]],
[kilden,
 [[cat:noun,root:kil,agr:'3SG',poss:'NONE',case:abl]]],
[yapIlmIS,
 [[cat:verb,root:yap,voice:pass,sense:pos,tam1:narr,agr:'3SG'],
  [cat:adj,stem:[cat:verb,root:yap,voice:pass,sense:pos,tam1:narr],suffix:none]]],
['"',
 [[cat:punct,root:'"']]],
[gOz,
 [[cat:noun,root:gOz,agr:'3SG',poss:'NONE',case:nom]]],
[putu,
 [[cat:noun,root:put,agr:'3SG',poss:'3SG',case:nom]]],
['"',
 [[cat:punct,root:'"']]],
```

```
 [bulunmuStur,
  [[cat:verb,root:bulun,sense:pos,tam1:narr,copula:'2',agr:'3SG']]],
 ['.',
  [[cat:punct,root:'.']]],
 [#,
  [[cat:end_of_sentence]]]].

[[@,
  [[cat:beginning_of_sentence]]],
 [tapInakta,
  [[cat:noun,root:tapInak,agr:'3SG',poss:'NONE',case:loc]]],
 [',',
  [[cat:punct,root:',']]],
 [yUkseklikleri,
  [[cat:noun,stem:[cat:adj,root:yUksek],suffix:lik,agr:'3PL',poss:'NONE',case:acc],
   [cat:noun,stem:[cat:adj,root:yUksek],suffix:lik,agr:'3PL',poss:'3SG',case:nom],
   [cat:noun,stem:[cat:adj,root:yUksek],suffix:lik,agr:'3PL',poss:'3PL',case:nom]]],
 ['2',
  [[cat:adj,type:cardinal,root:'2']]],
 [ile,
  [[cat:conn,root:ile]]],
 ['11',
  [[cat:adj,type:cardinal,root:'11']]],
 [cm,
  [[cat:noun,root:cmE,type:rproper,agr:'3SG',poss:'NONE',case:nom]]],
 [arasInda,
  [[cat:noun,root:ara,agr:'3SG',poss:'3SG',case:loc]]],
 [deGiSen,
  [[cat:adj,stem:[cat:verb,root:deGiS,sense:pos],suffix:yan],
   [cat:adj,stem:[cat:verb,root:deG,voice:recip,sense:pos],suffix:yan]]],
 [bu,
  [[cat:adj,root:bu,type:determiner]]],
 [adak,
  [[cat:noun,root:adak,agr:'3SG',poss:'NONE',case:nom]]],
 [simgelerinden,
  [[cat:noun,root:simge,agr:'3PL',poss:'3SG',case:abl]]],
 ['20,000 - 22,000',
  [[cat:adj,type:range,root:'20,000-22,000']]],
 [kadar,
  [[cat:postp,root:kadar,subcat:dat,type:temp2]]],
 [bulunduGu,
  [[cat:adj,stem:[cat:verb,root:bulun,sense:pos],suffix:dik,poss:'3SG'],
```

[cat:adj,stem:[cat:verb,root:bul,voice:pass,sense:pos],suffix:dik,poss:'3SG']]],
 [hesaplanmIStIr,
  [[cat:verb,root:hesapla,voice:pass,sense:pos,tam1:narr,copula:'2',agr:'3SG']]],
 ['.',
  [[cat:punct,root:'.']]],
 [#,
  [[cat:end_of_sentence]]]].

[[@,
  [[cat:beginning_of_sentence]]],
 [niceliksel,
  [[cat:adj,root:niceliksel]]],
 ['CalISmalar',
  [[cat:noun,stem:[cat:verb,root:'CalIS',sense:pos],suffix:ma,
    type:infinitive,agr:'3PL',poss:'NONE',case:nom]]],
 [yapIlacaGI,
  [[cat:adj,stem:[cat:verb,root:yap,voice:pass,sense:pos],suffix:yacak,poss:'3SG']]],
 [zaman,
  [[cat:noun,root:zaman,type:temp1,agr:'3SG',poss:'NONE',case:nom]]],
 [',',
  [[cat:punct,root:',']]],
 [temsili,
  [[cat:adj,root:temsili]]],
 [nitelikte,
  [[cat:noun,root:nitelik,agr:'3SG',poss:'NONE',case:loc]]],
 ['Ornekler',
  [[cat:adj,stem:[cat:verb,root:'Ornekle',sense:pos,tam1:aorist],suffix:none]]],
 ['elde etmenin',
  [[cat:noun,stem:[cat:verb,root:'elde
 ed',sense:pos],suffix:ma,type:infinitive,agr:'3SG',poss:'NONE',case:gen]]],
 [temel,
  [[cat:adj,root:temel]]],
 [yOntemlerinden,
  [[cat:noun,root:yOntem,agr:'3PL',poss:'3SG',case:abl]]],
 [biri,
  [[cat:pronoun,root:biri,type:quant,agr:'3SG',poss:'3SG',case:nom]]],
 [elemedir,
  [[cat:verb,stem:[cat:noun,stem:[cat:verb,root:ele,sense:pos],suffix:ma,type:infinitive,
    agr:'3SG',poss:'NONE',case:nom],suffix:none,tam2:pres,copula:'2',agr:'3SG']]],
 ['.',
  [[cat:punct,root:'.']]],
 [#,

```
 [[cat:end_of_sentence]]]].

[[@,
 [[cat:beginning_of_sentence]]],
[kumlu,
 [[cat:adj,stem:[cat:noun,root:kum],suffix:li]]],
[toprakta,
 [[cat:noun,root:toprak,agr:'3SG',poss:'NONE',case:loc]]],
[kuru,
 [[cat:adj,root:kuru]]],
[eleme,
 [[cat:noun,root:elem,agr:'3SG',poss:'NONE',case:dat]]],
['zaman zaman',
 [[cat:adverb,root:'zaman zaman']]],
[mUmkUndUr,
 [[cat:verb,stem:[cat:adj,root:mUmkUn],suffix:none,tam2:pres,copula:'2',agr:'3SG']]],
['.',
 [[cat:punct,root:'.']]],
[#,
 [[cat:end_of_sentence]]]].
```